\documentclass[10pt,twocolumn,letterpaper]{article}

\usepackage{titling}
\usepackage{iccv}
\usepackage{times}
\usepackage{epsfig}
\usepackage{graphicx}
\usepackage{amsmath}
\usepackage{amssymb}
\usepackage{multirow}
\usepackage{capt-of}
\usepackage{mathtools}
\usepackage{subcaption}
\usepackage{xcolor}
\usepackage{pdfpages}

\usepackage[ruled]{algorithm2e}
\SetKwInput{KwInput}{Input}
\SetKwInput{KwOutput}{Output}
\SetKwInput{KwRepeat}{Repeat}

\usepackage[breaklinks=true,bookmarks=false]{hyperref}

\iccvfinalcopy %

\def\iccvPaperID{838} %

\ificcvfinal\fi

\newif\ifshowmain
\newif\ifshowsupp
\showmaintrue %
\showsupptrue %

\begin{document}

\title{Unrestricted Facial Geometry Reconstruction Using Image-to-Image Translation}

\author{
 Matan Sela \qquad Elad Richardson \qquad Ron Kimmel \\ Department of Computer Science, Technion - Israel Institute of Technology\\
    {\tt\small \{matansel,eladrich,ron\}@cs.technion.ac.il}
    }

\makeatletter
\let\@oldmaketitle\@maketitle
\renewcommand{\@maketitle}{\@oldmaketitle%
\centering
\mbox{
\includegraphics[height=0.19\linewidth]{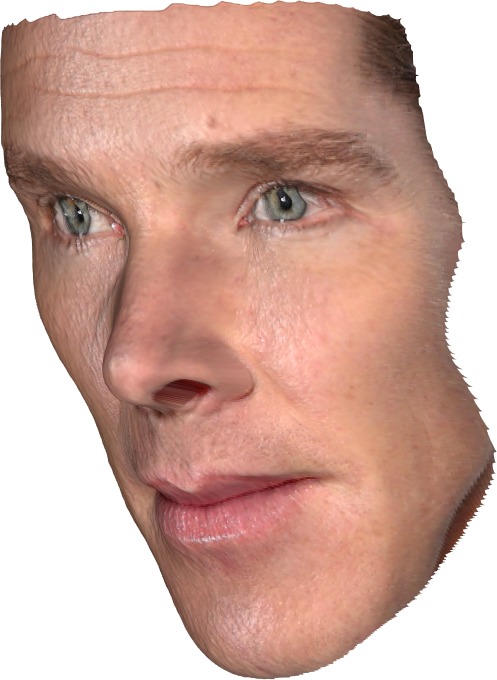}
\includegraphics[height=0.19\linewidth]{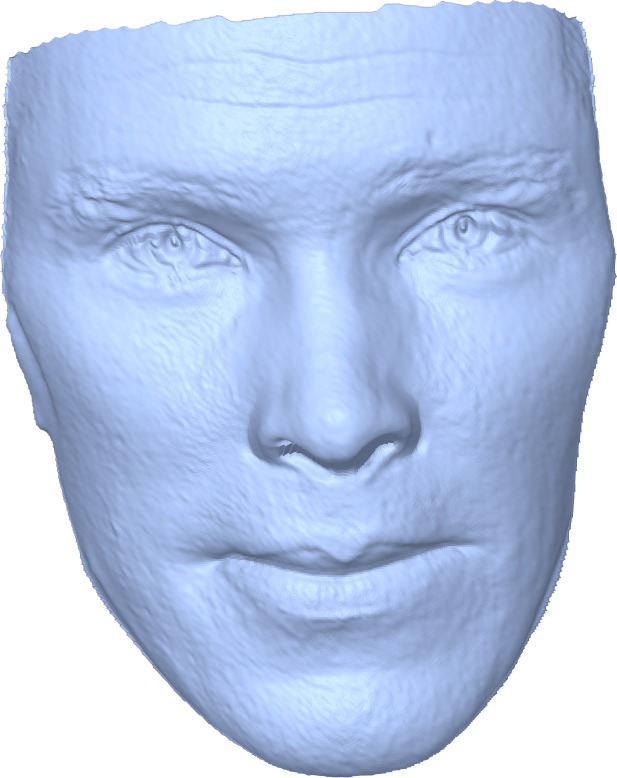}
\includegraphics[height=0.19\linewidth]{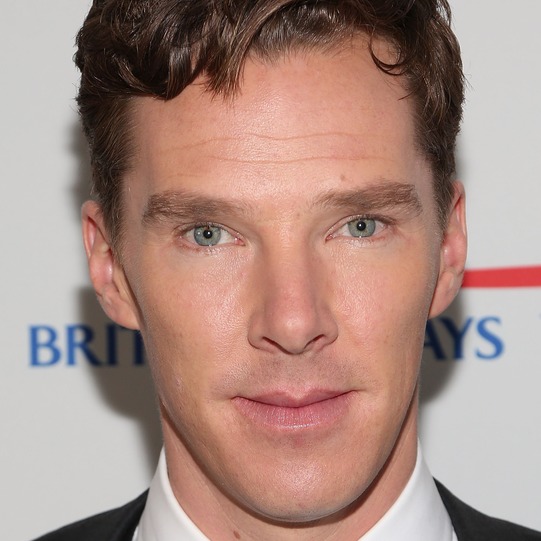}
\hspace{0.2cm}
\includegraphics[height=0.19\linewidth]{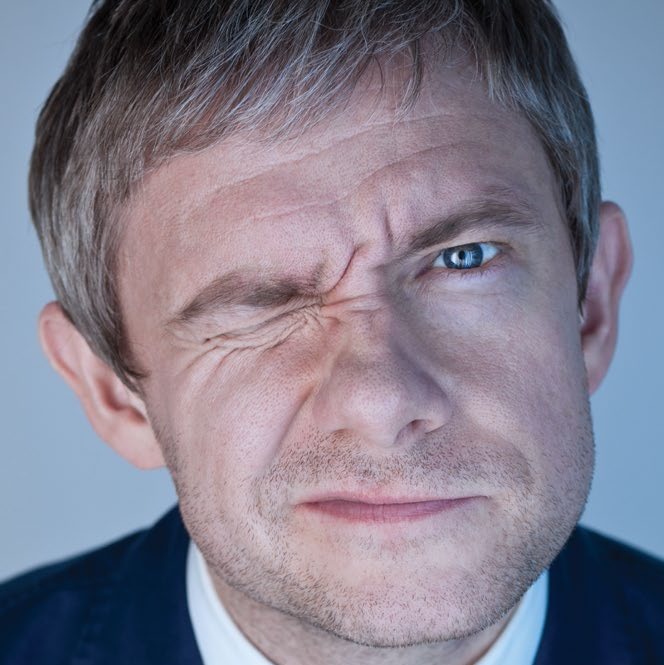}
\includegraphics[height=0.19\linewidth]{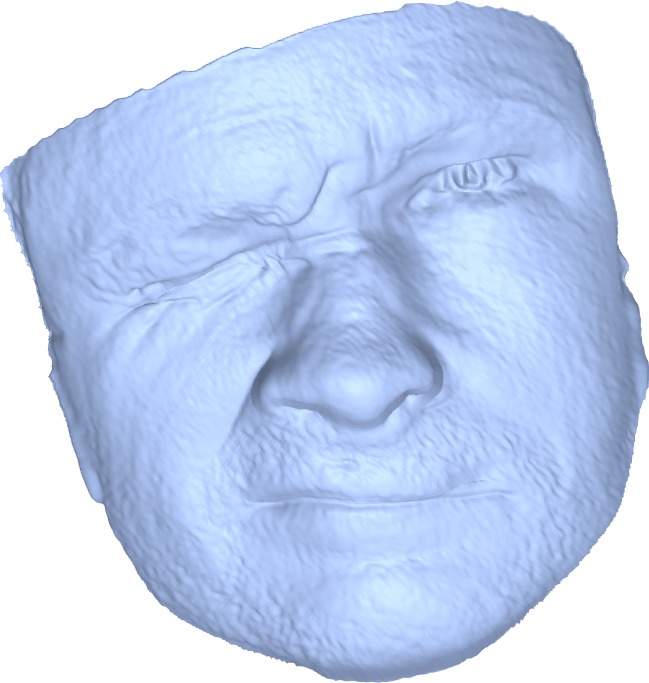}
\includegraphics[height=0.19\linewidth]{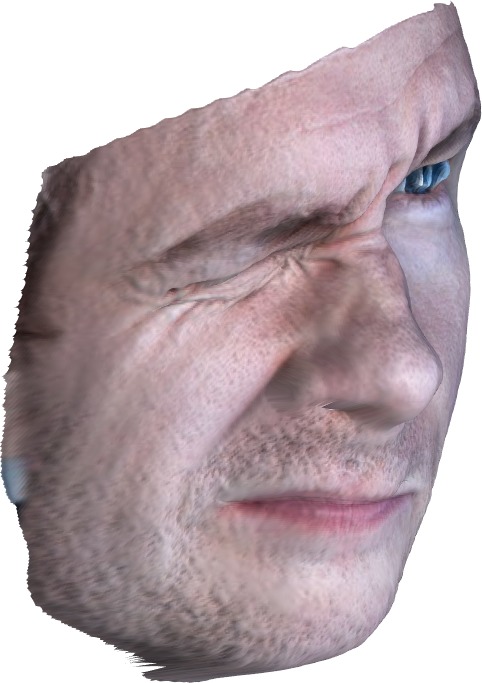}
}
\captionof{figure}{Results of the proposed method.
    Reconstructed geometries are shown next to the corresponding input images.}
    \label{fig:teaser}\bigskip}
\makeatother

\ifshowmain

\maketitle
\thispagestyle{empty}

\begin{abstract}
It has been recently shown that neural networks can recover the geometric structure
 of a face from a single given image.
A common denominator of most existing face geometry reconstruction methods
 is the restriction of the solution space to some low-dimensional subspace.
While such a model significantly simplifies the reconstruction problem,
 it is inherently limited in its expressiveness.
As an alternative, we propose an Image-to-Image translation network that jointly maps the input image to a depth image and a facial correspondence map.
This explicit pixel-based mapping can then be utilized to provide high quality
 reconstructions of diverse faces under extreme expressions, using a purely geometric refinement process.
In the spirit of recent approaches, the network is trained only with synthetic data,
 and is then evaluated on ``in-the-wild'' facial images.
Both qualitative and quantitative analyses demonstrate the accuracy and the robustness of our approach.
\end{abstract}

\section{Introduction}
Recovering the geometric structure of a face is a fundamental task in computer
 vision with numerous applications.
For example, facial characteristics of actors in realistic movies can be manually
 edited with facial rigs that are carefully designed for manipulating the expression~\cite{thies2016face2face}.
While producing animation movies, tracking the geometry of an actor across multiple frames
 allows transferring the expression to an animated
  avatar~\cite{garrido2016reconstruction,cao20133d,cao2015real}.
Image-based face recognition methods deform the recovered geometry for producing a neutralized
 and frontal version of the input face in a given image, reducing the variations between images
 of the same subject~\cite{zhu2015high,hassner2015effective}.
As for medical applications, acquiring the structure of a face allows for fine planning of
 aesthetic operations and plastic surgeries, designing of personalized masks
 \cite{amirav2014design, sela2016customized} and even bio-printing facial organs.

Here, we focus on the recovery of the geometric structure of a face from a single facial image under
 a wide range of expressions and poses.
This problem has been investigated for decades and most existing solutions involve one
 or more of the following components.

\begin{itemize}
\item Facial landmarks~\cite{kazemi2014one,zhang2014facial,ren2014face,zhou2013extensive}
 - a set of automatically detected key points on the face such as the tip of the nose and
  the corners of the eyes, which can guide the reconstruction process
   \cite{zhu2015high,kemelmacher20113d,aldrian2010linear,dou2014robust,liu2015cascaded}.
\item A reference facial model - an average neutral face that is used as an initialization of
 optical flow or shape from shading procedures \cite{hassner2015effective,kemelmacher20113d}.
\item A three-dimensional morphable model - a prior low-dimensional linear subspace of plausible
 facial geometries which allows an efficient, yet rough, recovery of a facial structure
\cite{blanz1999morphable,breuer2008automatic,zhu2015high,saito2016photorealistic,jiang20173d,
 richardson20163d,tran2016regressing},
\end{itemize}
\begin{figure*}
    \centering
      \includegraphics[width=\textwidth]{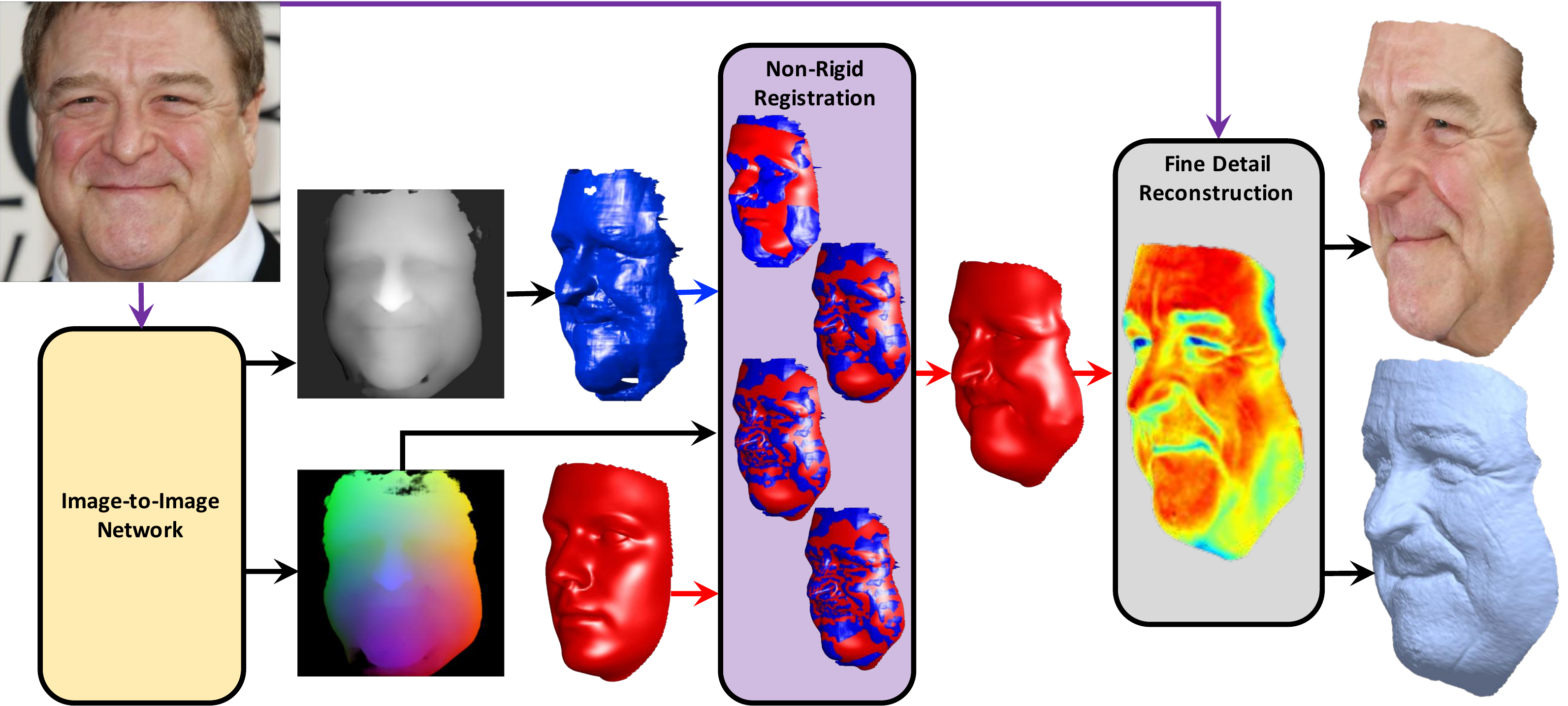}
    \caption{The algorithmic reconstruction pipeline. } %
    \label{fig:pipeline}
\end{figure*}

While using these components can simplify the reconstruction problem, they introduce some inherent limitations.
Methods that rely only on landmarks are limited to a sparse set of constrained points.
Classical techniques that use a reference facial model might fail to recover extreme expressions and non-frontal
 poses, as optical flows restrict the deformation to the image plane.
The morphable model, while providing some robustness, limits the reconstruction as it can express only coarse geometries.
Integrating some of these components together could mitigate the problems, yet, the underlying limitations are still manifested in the final reconstruction.

Alternatively, we propose an unrestricted approach which involves a fully convolutional network that learns to translate an input facial image to a representation containing two maps.
The first map is an estimation of a depth image, while the second is an embedding of a facial
 template mesh in the image domain. This network is trained following the Image-to-Image translation framework of~\cite{isola2016image}, where an additional normal-based loss is introduced to enhance the depth result.
Similar to previous approaches, we use synthetic images for training, where the images are sampled from a wide range of facial identities, poses, expressions, lighting conditions, backgrounds and material parameters.
Surprisingly, even though the network is still trained with faces that are drawn from a limited generative model, it can generalize and produce structures far and beyond the limited scope of that model.
To process the raw network results, an iterative facial deformation procedure is used which combines the representations into a full facial mesh. Finally, a refinement step is applied to produce a detailed reconstruction.
This novel blending of neural networks with purely geometric techniques allows us to reconstruct high-quality meshes with wrinkles and details at a mesoscopic-level from only a single image.

While using a neural network for face reconstruction was proposed in the
 past~\cite{richardson20163d,richardson2016learning,tran2016regressing,zhu2016face,jourabloo2016large},
 previous methods were still limited by the expressiveness of the linear model.
In~\cite{richardson2016learning}, a second network was proposed to refine the coarse facial
 reconstruction, yet, it could not compensate for large geometric variations beyond the given subspace.
For example, the structure of the nose was still limited by the span of a facial morphable model.
By learning the unconstrained geometry directly in the image domain, we overcome this limitation,
 as demonstrated by both quantitative and qualitative experimental results. To further analyze the potential of the proposed representation we devise an application for translating images from one domain to another.
 As a case study, we transform synthetic facial images into realistic ones, enforcing our network as a loss function to preserve the geometry throughout the cross domain mapping.

The main contributions of this paper are:
\begin{itemize}
\item A novel formulation for predicting a geometric representation of a face from a single image, which is not restricted to a linear model.
\item A purely geometric deformation and refinement procedure that utilizes the network representation to produce high quality facial reconstructions.
\item A novel application of the proposed network which allows translating synthetic
 facial images into realistic ones, while keeping the geometric structure intact.
\end{itemize}

\section{Overview}
The algorithmic pipeline is presented in~\autoref{fig:pipeline}.
The input of the network is a facial image, and the network produces two outputs:
The first is an estimated depth map aligned with the input image.
The second output is a dense map from each pixel to a corresponding vertex on a reference facial mesh. To bring the results into full vertex correspondence and complete occluded parts of the face,
 we warp a template mesh in the three-dimensional space by an iterative non-rigid deformation procedure.
Finally, a fine detail reconstruction algorithm guided by the input image recovers the subtle
 geometric structure of the face. Code for evaluation is available at \url{https://github.com/matansel/pix2vertex}.

\section{Learning the Geometric Representation}
There are several design choices to consider when working with neural networks.
First and foremost is the training data, including the input channels, their labels,
 and how to gather the samples.
Second is the choice of the architecture.
A common approach is to start from an existing
 architecture~\cite{krizhevsky2012imagenet,simonyan2014very,szegedy2015going,he2015deep}
 and to adapt it to the problem at hand.
Finally, there is the choice of the training process, including the loss criteria and the optimization technique.
Next, we describe our choices for each of these elements.

\subsection{The Data and its Representation}
The purpose of the suggested network is to regress a geometric representation from a given facial image.
This representation is composed of the following two components:
\paragraph{Depth Image} A depth profile of the facial geometry.
Indeed, for many facial reconstruction tasks providing only the depth profile
 is sufficient~\cite{hassner2013viewing,kemelmacher20113d}.

\begin{figure}[b]
    \centering
     \includegraphics[height=0.15\textwidth]{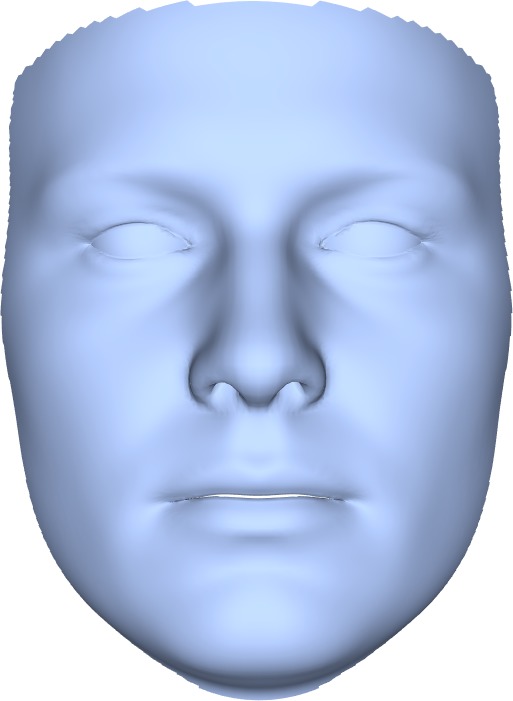}
     \hspace{0.2cm}
      \includegraphics[height=0.15\textwidth]{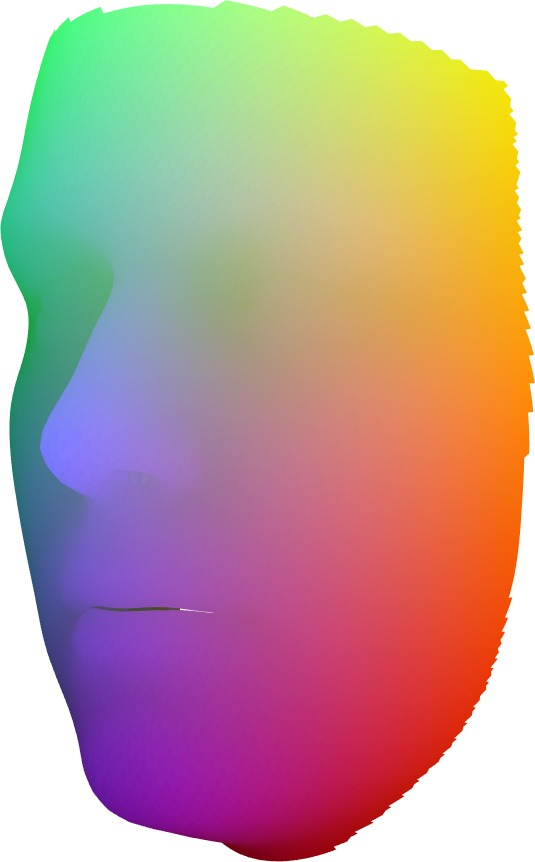}
      \includegraphics[height=0.15\textwidth]{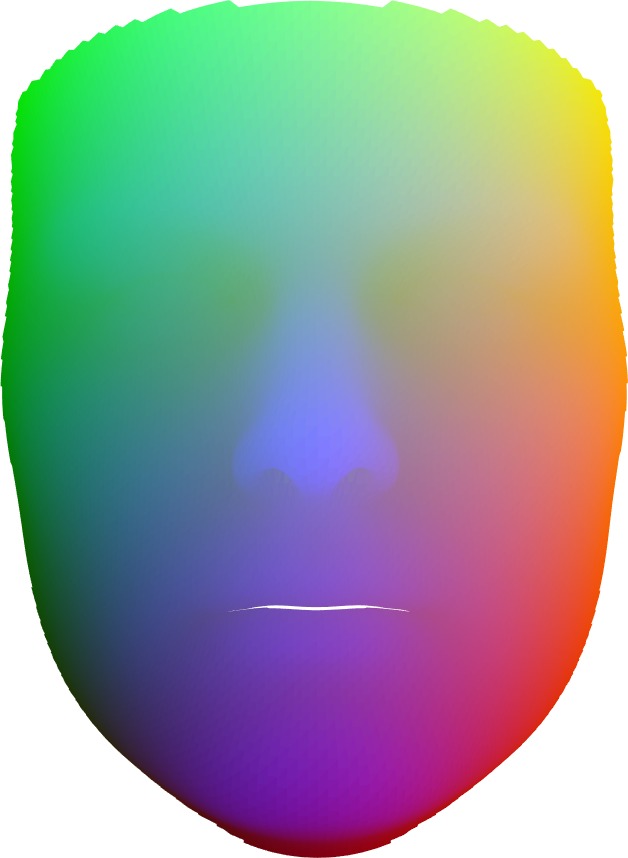}
      \includegraphics[height=0.15\textwidth]{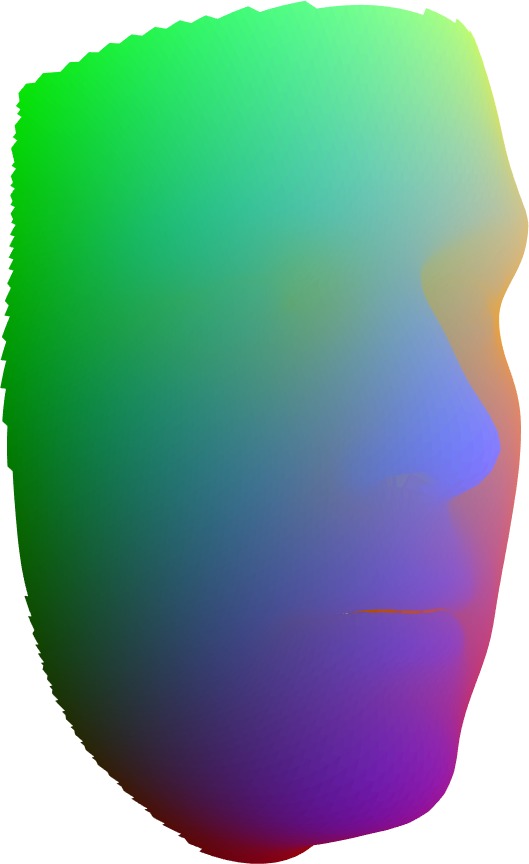}
    \caption{A reference template face presented alongside the dense correspondence signature from different viewpoints.}
    \label{fig:rep}
\end{figure}

\paragraph{Correspondence Map}
An embedding which allows mapping image pixels to points on a template facial model,
 given as a triangulated mesh.
To compute this signature for any facial geometry, we paint each vertex with the $x$, $y$, and $z$ coordinates
 of the corresponding point on a normalized canonical face.
Then, we paint each pixel in the map with the color value of the corresponding projected vertex, see~\autoref{fig:rep}.
This feature map is a deformation agnostic representation, which is useful for applications such
 as facial motion capture~\cite{weise2011realtime}, face normalization~\cite{zhu2015high}
  and texture mapping~\cite{zigelman2002texture}.
While a similar representation was used in~\cite{richardson2016learning,zhu2016face} as feedback channel
 for an iterative network, the facial recovery was still restricted to the span of a facial morphable model.

For training the network, we adopt the same synthetic data generation procedure proposed in~\cite{richardson20163d}.
Each random face is generated by drawing random mesh coordinates $S$ and texture $T$ from a facial morphable model~\cite{blanz1999morphable}.
In practice, we draw a pair of Gaussian random vectors, $\alpha_{g}$ and $\alpha_t$,
 and recover the synthetic face as follows
\begin{align*}
S &= \mu_g + A_g \alpha_g \cr
T &= \mu_t + A_t \alpha_t.
\end{align*}
where $\mu_g$ and $\mu_t$ are the stacked average facial geometry and texture of the model, respectively.
$A_g$ and $A_t$ are matrices whose columns are the bases of low-dimensional linear subspaces spanning
 plausible facial geometries and textures, respectively.
  Notice that geometry basis $A_g$ is composed to both identity and expression basis elements, as proposed in~\cite{chu20143d}.
Next, we render the random textured meshes under various illumination conditions and poses, generating a dataset of synthetic facial images.
As the ground-truth geometry is known for each synthetic image, one readily has the matching depth and correspondence maps to use as labels.
Some examples of input images alongside their desired outputs are shown in~\autoref{fig:examples}.

\begin{figure}[t]
    \centering
      \includegraphics[width=0.48\textwidth]{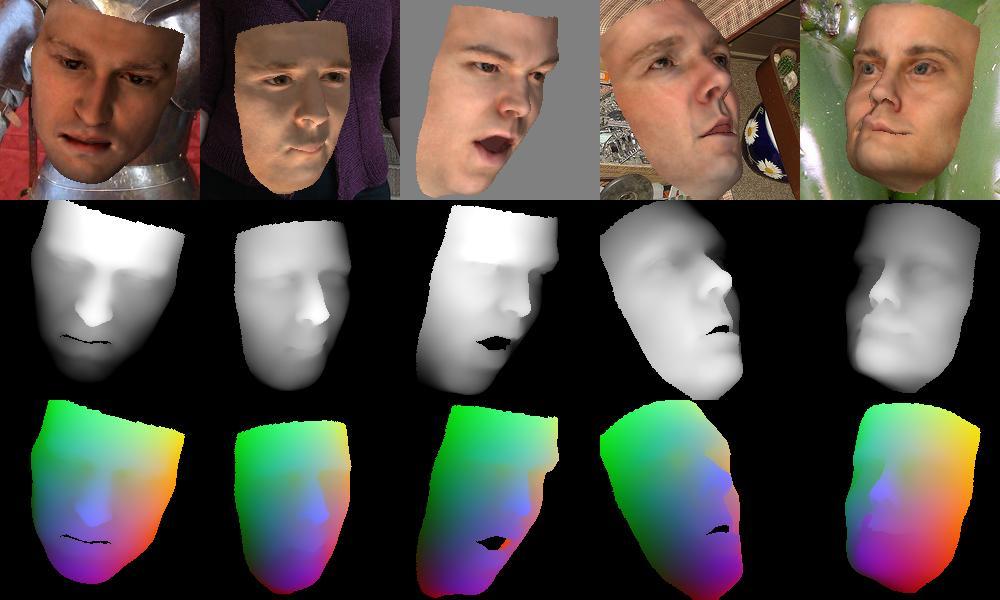}
    \caption{Training data samples alongside their representations.}
    \label{fig:examples}
\end{figure}

Working with synthetic data can still present some gaps when generalizing to ``in-the-wild'' images~\cite{chen2016synthesizing,richardson20163d}, however it provides much-needed flexibility in the generation process and ensures a deterministic connection from an image to its label.
Alternatively, other methods~\cite{guler2016densereg,tran2016regressing} proposed to generate training data by employing existing reconstruction algorithms and regarding their results as ground-truth labels.
For example, G{\"u}ler \emph{et al.}~\cite{guler2016densereg}, used a framework similar to that of~\cite{zhu2016face} to match dense correspondence maps to a dataset of facial images, starting from only a sparse set of landmarks. These correspondence maps were then used as training labels for their method. Notice that such data can also be used for training our network without requiring any other modification.

\subsection{Image to Geometry Translation}
\label{subsec:network1}
Pixel-wise prediction requires a proper network
 architecture~\cite{long2015fully,hariharan2015hypercolumns}.
The proposed structure is inspired by the recent Image-to-Image translation framework proposed in~\cite{isola2016image},
 where a network was trained to map the input image to output images of various types.
The architecture used there is based on the U-net~\cite{ronneberger2015u} layout, where skip connections are used between corresponding layers in the encoder and the decoder. Additional considerations as to the network implementation are given in the supplementary.

While in~\cite{isola2016image} a combination of $L_1$ and adversarial loss functions were used, in the proposed framework, we chose to omit the adversarial loss.
That is because unlike the problems explored in~\cite{isola2016image}, our setup includes less ambiguity in the mapping. Hence, a distributional loss function is less effective, and mainly introduces artifacts.
Still, since the basic $L_1$ loss function favors sparse errors in the depth prediction and does not account for differences between pixel neighborhoods, it is insufficient for producing fine geometric structures, see~\autoref{subfig:normals-no-loss}.
Hence, we propose to augment the loss function with an additional $L_1$ term, which penalizes the
 discrepancy between the normals of the reconstructed depth and ground truth.
\begin{equation}
L_N\left(\hat{z},z\right)=\left\Vert \vec{n}\left(\hat{z}\right)-\vec{n}\left(z\right)\right\Vert _{1},
\end{equation}
where $\hat{z}$ is the recovered depth, and $z$ denotes the ground-truth depth image.
During training we set $\lambda_{{L_1}}=100$ and $\lambda_{{N}}=10$, where $\lambda_{{L_1}}$ and $\lambda_{N}$ are the matching loss weights. Note that for the correspondence image only the $L_1$ loss was applied.
\autoref{fig:add_normals} demonstrates the contribution of the $L_N$ to the quality of the depth reconstruction provided by the network.

\begin{figure}[b] %
    \centering
    \begin{subfigure}[b]{0.125\textwidth}
      \includegraphics[width=\textwidth]{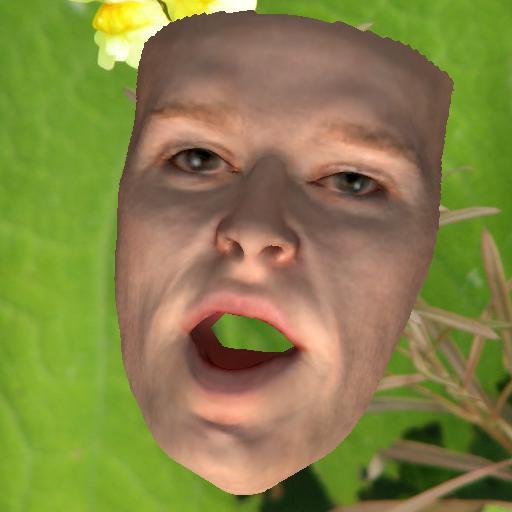}
      \caption{}
      \label{subfig:normals-im}
    \end{subfigure}
    \begin{subfigure}[b]{0.165\textwidth}
      \includegraphics[width=\textwidth,trim={3cm 15cm 2cm 5cm},clip]{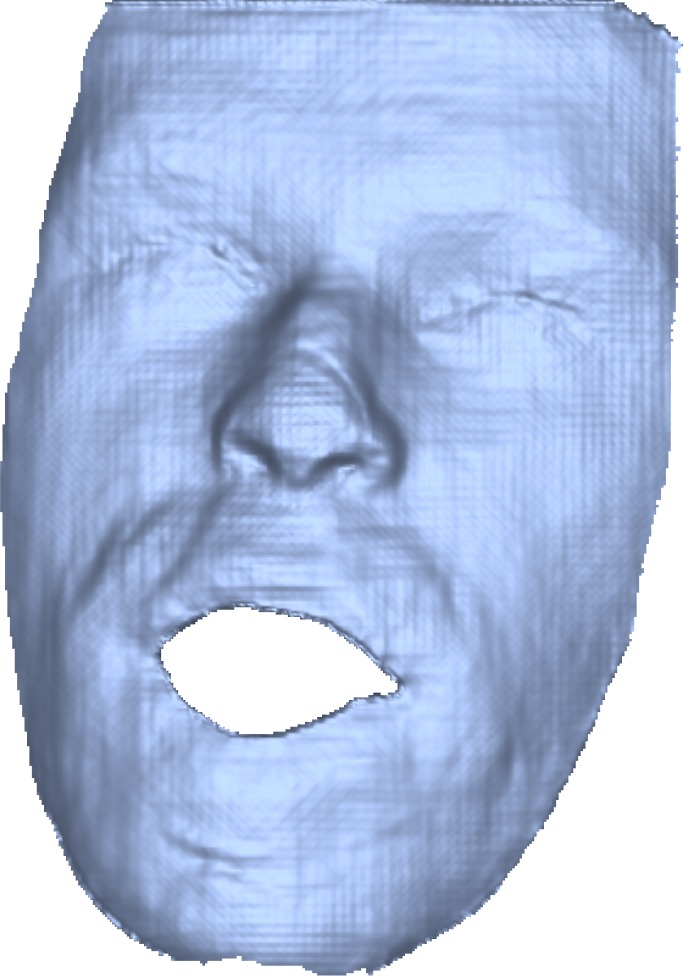}
      \caption{}
      \label{subfig:normals-no-loss}
    \end{subfigure}
    \begin{subfigure}[b]{0.165\textwidth}
      \includegraphics[width=\textwidth,trim={3cm 15cm 2cm 5cm},clip]{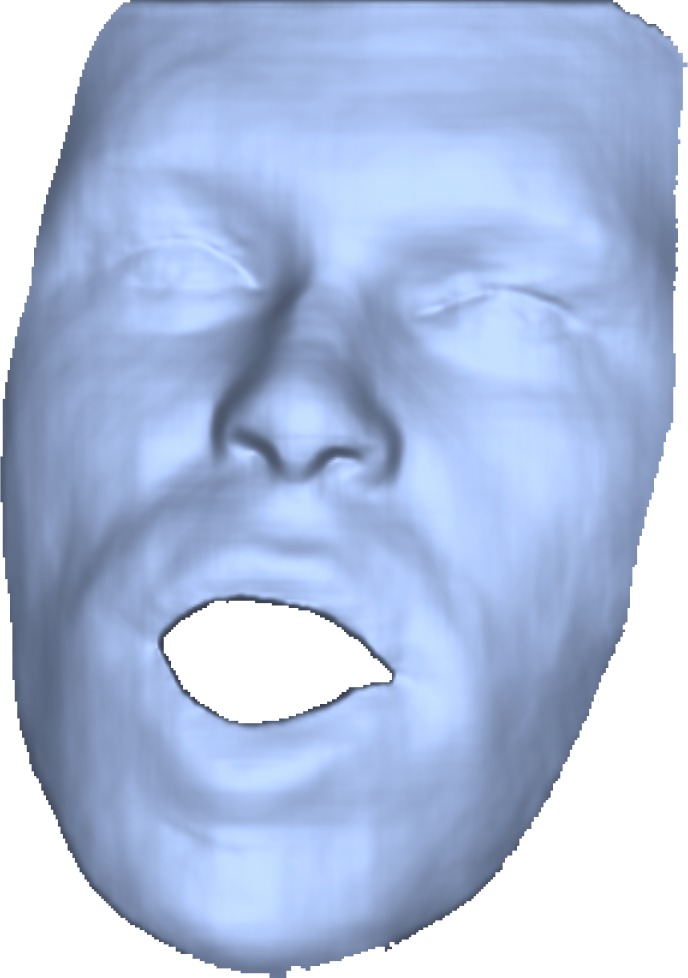}
      \caption{}
      \label{subfig:normals-with-loss}
    \end{subfigure}
    \caption{ (\subref{subfig:normals-im}) the input image, (\subref{subfig:normals-no-loss}) the result with only the $L_1$ loss function and (\subref{subfig:normals-with-loss}) the result with the additional normals loss function. Note the artifacts in (\subref{subfig:normals-no-loss}).}
    \label{fig:add_normals}
\end{figure}

\section{From Representations to a Mesh}
Based on the resulting depth and correspondence we introduce an approach to translate the 2.5D representation to a 3D facial mesh.
The procedure is composed of an iterative elastic deformation algorithm (\ref{subsec:deformation}) followed by a
 fine detail recovery step driven by the input image (\ref{subsec:mesoscopic}).
The resulting output is an accurate reconstructed facial mesh with a full vertex correspondence to a template mesh
 with fixed triangulation.
This type of data is helpful for various dynamic facial processing applications, such as facial rigs, which allows
 creating and editing photo-realistic animations of actors.
As a byproduct, this process also corrects the prediction of the network by completing domains in the face which
 are mistakenly classified as part of the background.
\subsection{Non-Rigid Registration}
\label{subsec:deformation}
Next, we describe the iterative deformation-based registration pipeline.
First, we turn the depth map from the network into a mesh, by connecting neighboring pixels.
Based on the correspondence map from the network, we compute the affine transformation from a
 template face to the mesh.
This operation is done by minimizing the squared Euclidean distances between
 corresponding vertex pairs.
Next, similar to \cite{li2010animation}, an iterative non-rigid registration process deforms the
 transformed template, aligning it with the mesh.
Note that throughout the registration, only the template is warped, while the target mesh remains fixed.
Each iteration involves the following four steps.
\begin{enumerate}
\item Each vertex in the template mesh, $v_i \in \mathcal{V}$, is associated with a vertex, $c_i$,
 on the target mesh, by evaluating the nearest neighbor in the correspondence embedding space.
This step is different from the method described in \cite{li2010animation}, which computes the nearest
 neighbor in the Euclidean space.
As a result, the proposed step allows registering a single template face to different facial
 identities with arbitrary expressions.

\item Pairs, $(v_i,c_i)$, which are physically distant and those whose normal directions disagree are
 detected and ignored in the next step.
\item The template mesh is deformed by minimizing the following energy
\begin{eqnarray}
E(V,C) &=& \alpha_{p2point} \sum_{(v_i,c_i)\in\mathcal{J}} \|v_i - c_i\|^2_2  \cr
 & & + \alpha_{p2plane}\sum_{(v_i,c_i)\in\mathcal{J}}\left| \vec{n}(c_i)(v_i-c_i) \right|^2  \cr
 & & + \alpha_{memb} \sum_{i \in \mathcal{V}}\sum_{v_j\in
             \mathcal{N}(v_i)} w_{i,j} \|v_i-v_j\|^2_2,\cr && \,\,\,
\end{eqnarray}
where, $w_{i,j}$ is the weight corresponding to the biharmonic Laplacian operator (see \cite{helenbrook2003mesh,botsch08variational}), $\vec{n}(c_i)$ is the normal of the corresponding vertex at the target mesh $c_i$, $\mathcal{J}$ is the set of the remaining associated vertex pairs $(v_i,c_i)$, and $\mathcal{N}(v_i)$ is the set 1-ring neighboring vertices about the vertex $v_i$.
Notice that the first term above is the sum of squared Euclidean distances between matches.
The second term is the distance from the point $v_i$ to the tangent plane at the corresponding point of the target mesh.
The third term quantifies the stiffness of the mesh.
\item If the motion of the template mesh between the current iteration and the previous one is below a fixed threshold, we divide the weight $\alpha_{memb}$ by two.
This relaxes the stiffness term and allows a greater deformation in the next iteration.
\end{enumerate}
This iterative process terminates when the stiffness weight is below a given threshold.
Further implementation information and parameters of the registration process are provided in the supplementary material.
The resulting output of this phase is a deformed template with fixed triangulation, which contains the overall facial structure recovered by the network, yet, is smoother and complete, see the third column of~\autoref{fig:stages}.
\subsection{Fine Detail Reconstruction}\label{subsec:mesoscopic}
Although the network already recovers some fine geometric details, such as wrinkles and moles, across parts of the face, a geometric approach can reconstruct details at a finer level, on the entire face, independently of the resolution.
Here, we propose an approach motivated by the passive-stereo facial reconstruction method suggested in \cite{beeler2010singleshot}.
The underlying assumption here is that subtle geometric structures can be explained by local variations in the image domain.
For some skin tissues, such as nevi, this assumption is inaccurate as the intensity variation results from the albedo.
In such cases, the geometric structure would be wrongly modified.
Still, for most parts of the face, the reconstructed details are consistent with the actual variations in depth.

The method begins from an interpolated version of the deformed template.
Each vertex $v\in\mathcal{V}_{\mathcal{D}}$ is painted with the intensity value of the nearest pixel in the image plane.
Since we are interested in recovering small details, only the high spatial frequencies, $\mu(v)$, of the texture, $\tau(v)$, are taken into consideration in this phase.
For computing this frequency band, we subtract the synthesized low frequencies from the original intensity values.
This low-pass filtered part can be computed by convolving the texture with a spatially varying Gaussian kernel in the image domain, as originally proposed.
In contrast, since this convolution is equivalent to computing the heat distribution upon the shape after time $dt$, where the initial heat profile is the original texture, we propose to compute $\mu(v)$ as
\begin{align}
 \mu(v) = \tau(v) - (I - dt\cdot \Delta_g)^{-1}\tau(v),
\end{align}
where $I$ is the identity matrix, $\Delta_g$ is the cotangent weight discrete Laplacian operator for triangulated meshes \cite{meyer2002discrete}, and $dt$ is a scalar proportional to the cut-off frequency of the filter.

Next, we displace each vertex along its normal direction such that $v' = v + \delta(v) \vec{n}(v)$.
The step size of the displacement, $\delta(v)$, is a combination of a data-driven term, $\delta_{\mu}(v)$, and a regularization one, $\delta_{s}(v)$.
The data-driven term is guided by the high-pass filtered part of the texture, $\mu(v)$.
In practice, we require the local differences in the geometry to be proportional to the local variation in the high frequency band of the texture. For each vertex $v$, with a normal $\vec{n}(v)$, and a neighboring vertex $v_i$, the data-driven term is given by
\begin{align}
\delta_{\mu}(v)=\frac{\underset{v_{i}\in\mathcal{N}(v)}{\sum}\alpha_{(v,v_{i})}\left(\mu(v)-\mu(v_{i})\right)\left(1-\frac{\left|\langle v-v_{i},\vec{n}(v)\rangle\right|}{\|v-v_{i}\|}\right)}{\underset{v_{i}\in\mathcal{N}(v)}{\sum}\alpha_{(v,v_{i})}},
 \label{eqn:mesoscopic}
\end{align}
where $\alpha_{(v,v_i)} = \exp{\left(-\|v-v_i\|\right)}$.
For further explanation of ~\autoref{eqn:mesoscopic}, we refer the reader to the supplementary material of this paper or the implementation details of \cite{beeler2010singleshot}.

Since we move each vertex along the normal direction, triangles could intersect each other, particularly in domains of high curvature.
To reduce the probability of such collisions,
a regularizing displacement field, $\delta_s(v)$, is added.
This term is proportional to the mean curvature of the original surface, and is equivalent to a single explicit mesh fairing step \cite{desbrun1999meshfairing}.
The final surface modification is given by
\begin{align}
v' = v + (\eta \delta_{\mu}(v) + (1-\eta) \delta_s(v))\cdot \vec{n}(v),
\end{align}
for some constant $\eta \in \left[0,1\right]$.
A demonstration of the results before and after this step is presented in ~\autoref{fig:mesoscopic}
\begin{figure} %
    \centering
     \includegraphics[height=0.098\textwidth]{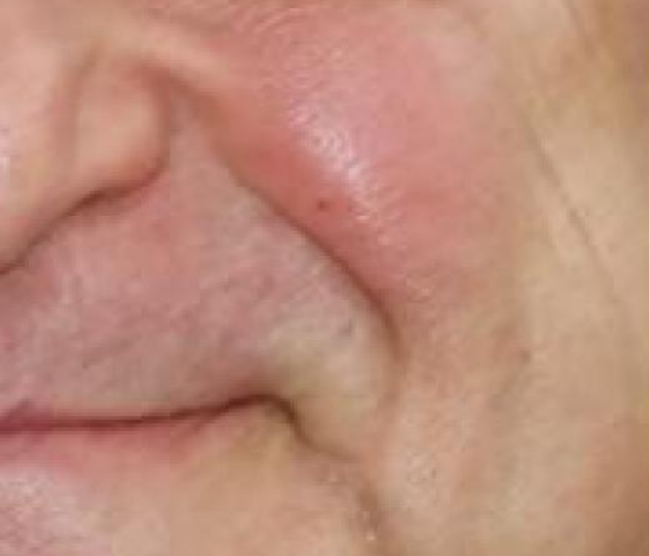}
      \includegraphics[height=0.098\textwidth]{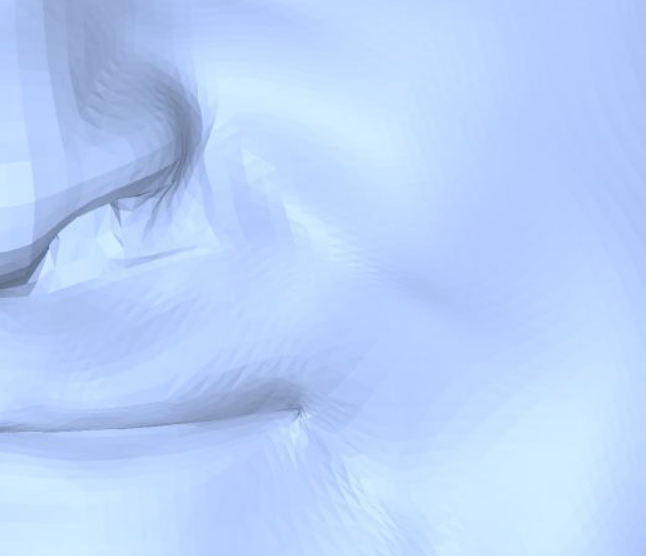}
      \includegraphics[height=0.098\textwidth]{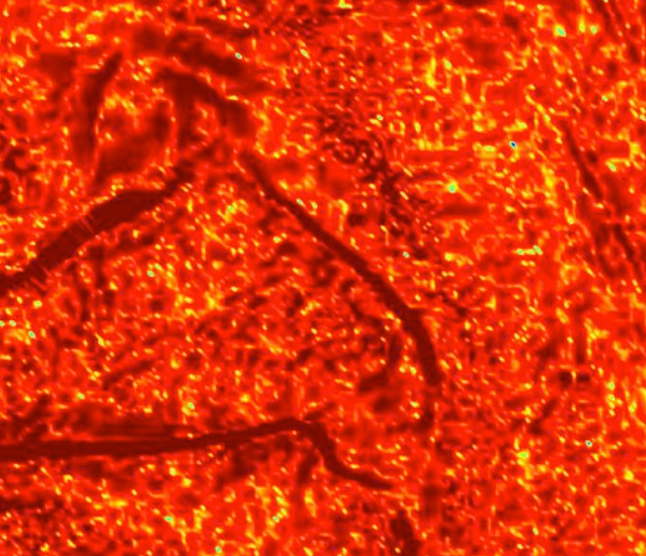}
      \includegraphics[height=0.098\textwidth]{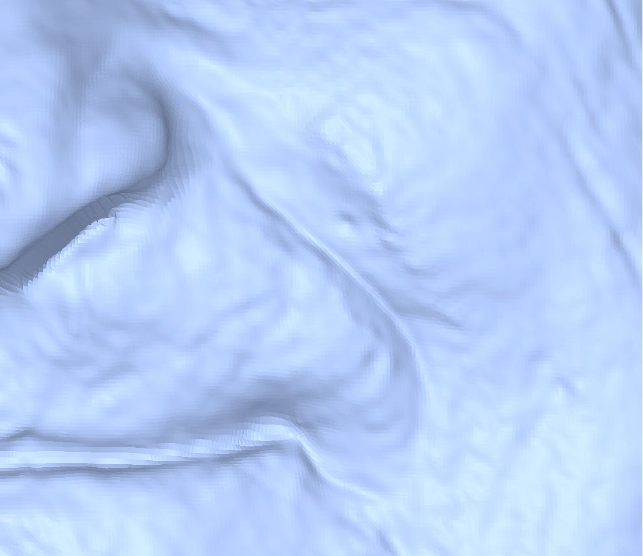}
    \caption{Mesoscopic displacement. From left to right: an input image, the shape after the iterative registration, the high-frequency part of the texture - $\mu(v)$, and the final shape. }
    \label{fig:mesoscopic}
\end{figure}

\section{Experiments}
Next, we present evaluations on both the proposed network and the pipeline as a whole, and comparison to different prominent methods of single image based facial reconstruction~\cite{kemelmacher20113d,zhu2015high,richardson2016learning}.

\subsection{Qualitative Evaluation}
\begin{figure}
    \centering
     \includegraphics[width=0.091\textwidth]{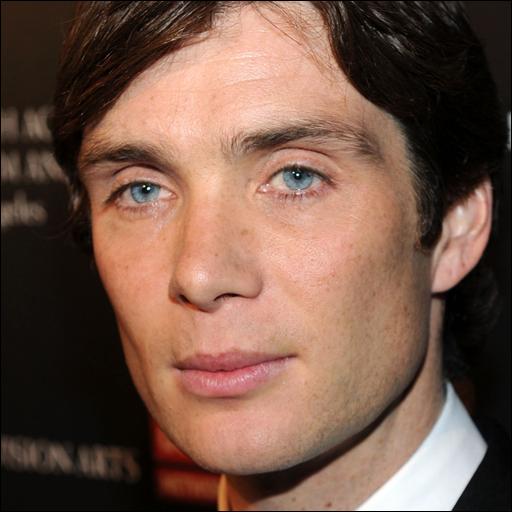}
      \includegraphics[width=0.091\textwidth]{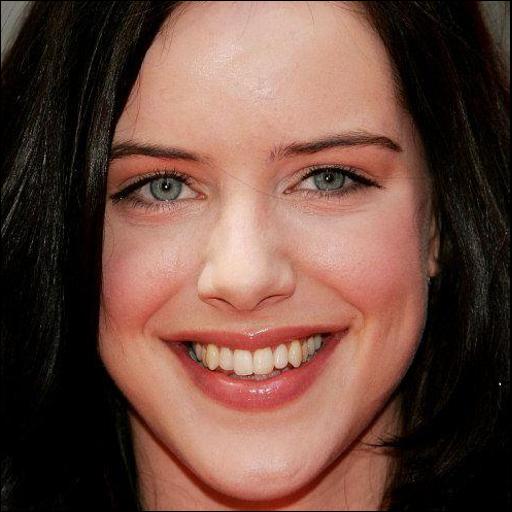}
      \includegraphics[width=0.091\textwidth]{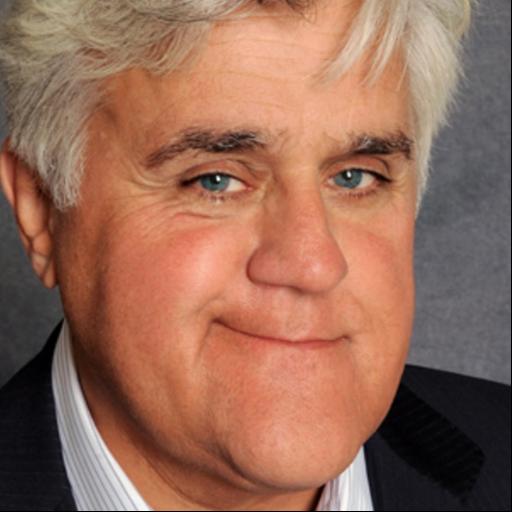}
      \includegraphics[width=0.091\textwidth]{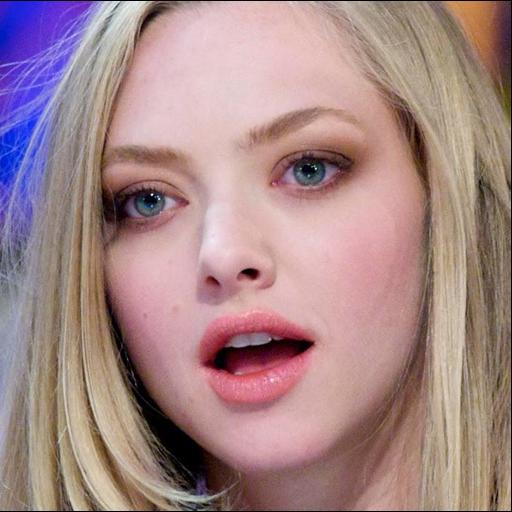}
      \includegraphics[width=0.091\textwidth]{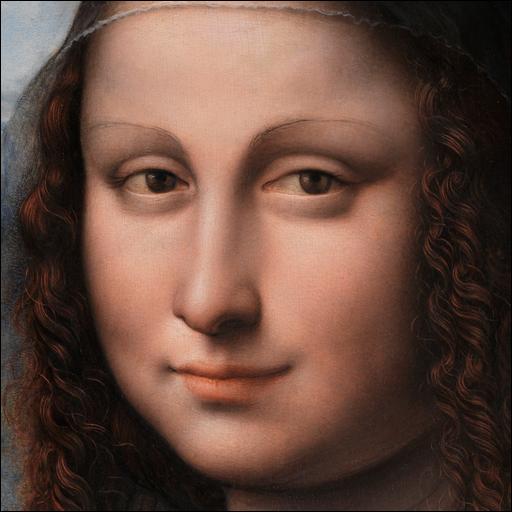}

      \includegraphics[width=0.091\textwidth]{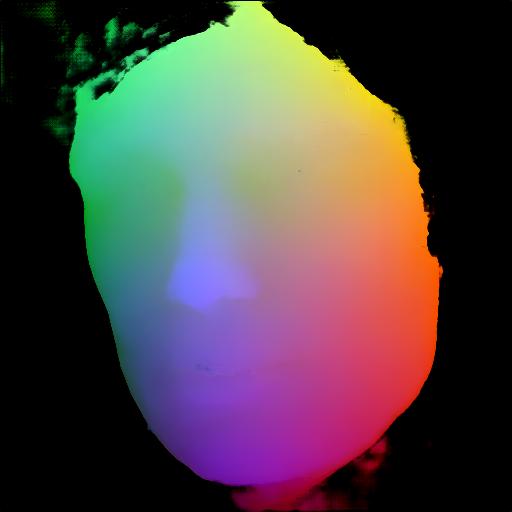}
      \includegraphics[width=0.091\textwidth]{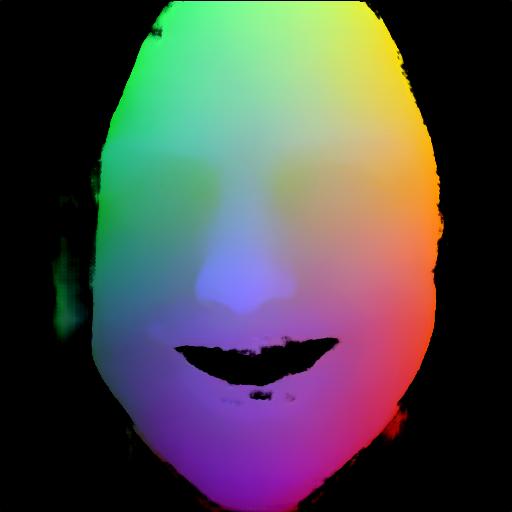}
      \includegraphics[width=0.091\textwidth]{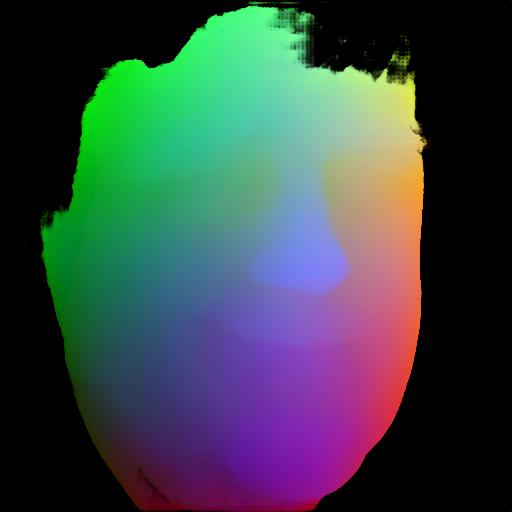}
      \includegraphics[width=0.091\textwidth]{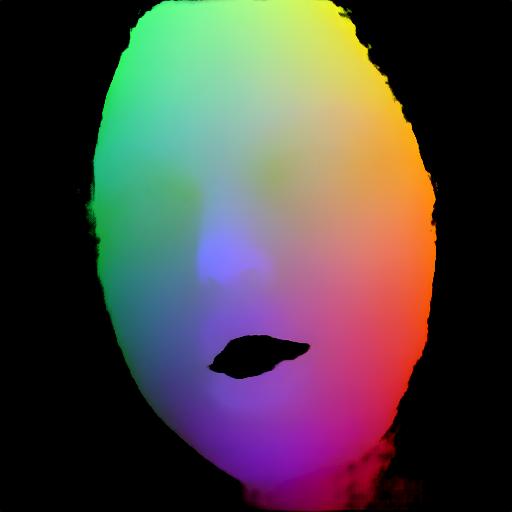}
      \includegraphics[width=0.091\textwidth]{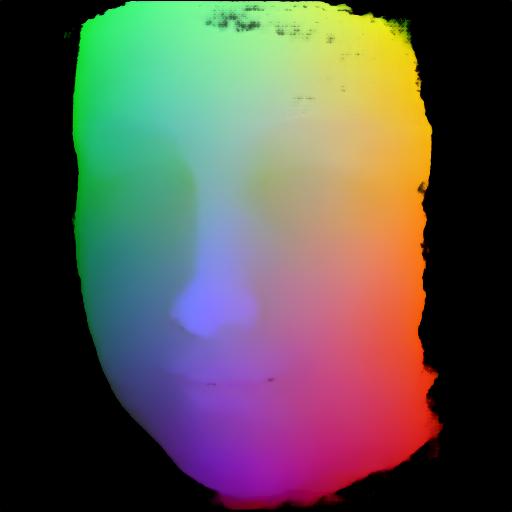}

      \includegraphics[width=0.091\textwidth]{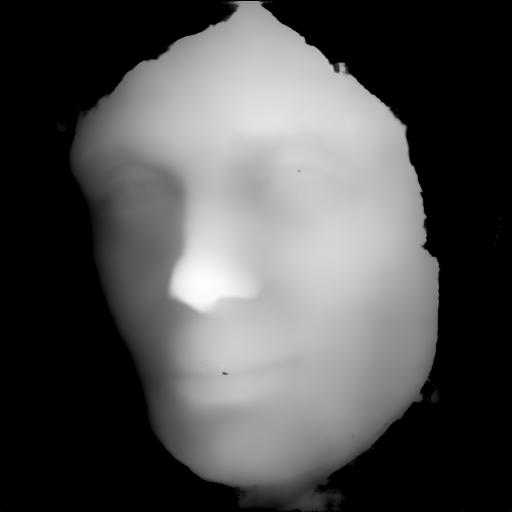}
      \includegraphics[width=0.091\textwidth]{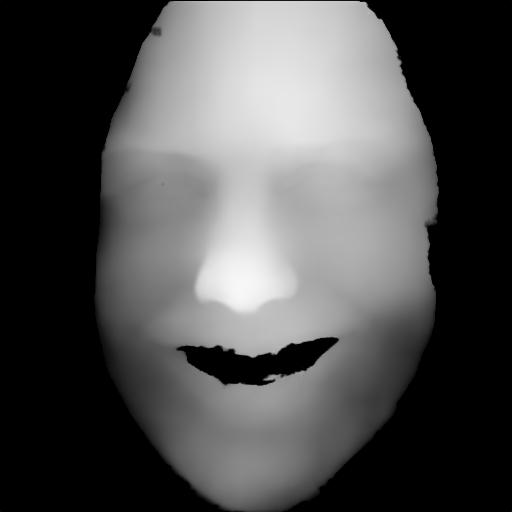}
      \includegraphics[width=0.091\textwidth]{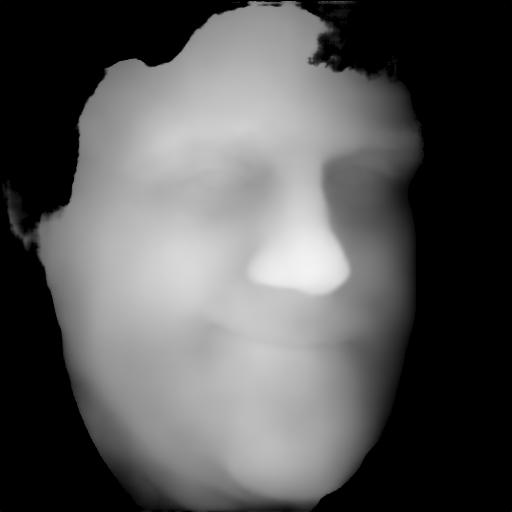}
      \includegraphics[width=0.091\textwidth]{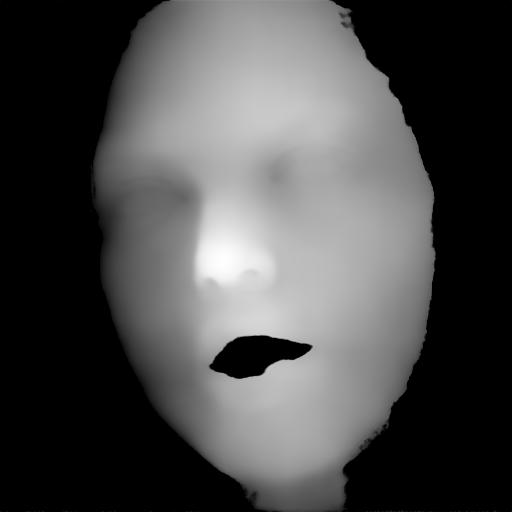}
      \includegraphics[width=0.091\textwidth]{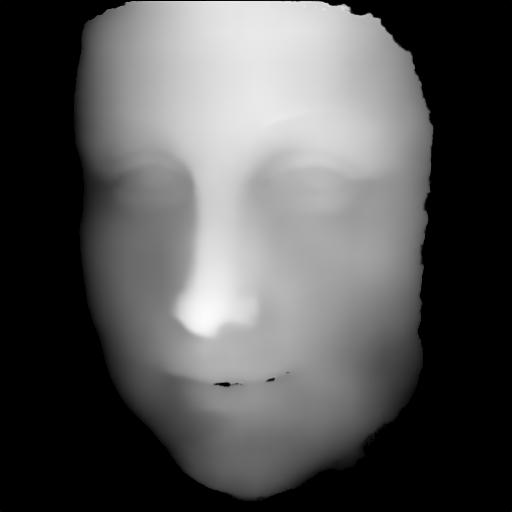}

    \caption{Network Output.}
    \label{fig:net_out}
    \vspace{-0.1cm}
\end{figure}

The first component of our algorithm is an Image-to-Image network.
In~\autoref{fig:net_out}, we show samples of output maps produced by the proposed network.
Although the network was trained with synthetic data, with simple random backgrounds (see~\autoref{fig:examples}), it successfully separates the hair and background from the face itself and learns the corresponding representations.
To qualitatively assess the accuracy of the correspondence, we present a visualization where an average facial texture is mapped to the image plane via the predicted embedding, see~\autoref{fig:corr_vis}, this shows how the network successfully learns to represent the facial structure.
Next, in~\autoref{fig:stages} we show the reconstruction of the network, alongside the registered template and the final shape.
Notice how the structural information retrieved by the network is preserved through the geometric stages. \autoref{fig:qual_results} shows a qualitative comparison between the proposed method and others.
One can see that our method better matches the global structure, as well as the facial details. To better perceive these differences, see~\autoref{fig:qual_results_zoom}.
Finally, to demonstrate the limited expressiveness of the 3DMM space compared to our method, \autoref{fig:3dmm_compare} presents our registered template next to its projection onto the 3DMM space. This clearly shows that our network is able to learn structures which are not spanned by the 3DMM model.

\begin{figure}
    \centering
     \includegraphics[width=0.115\textwidth]{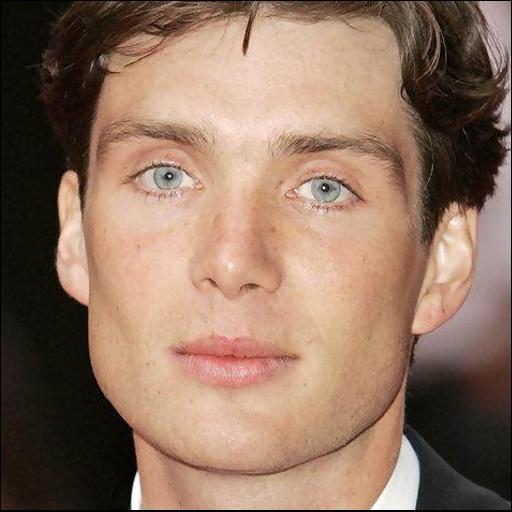}
      \includegraphics[width=0.115\textwidth]{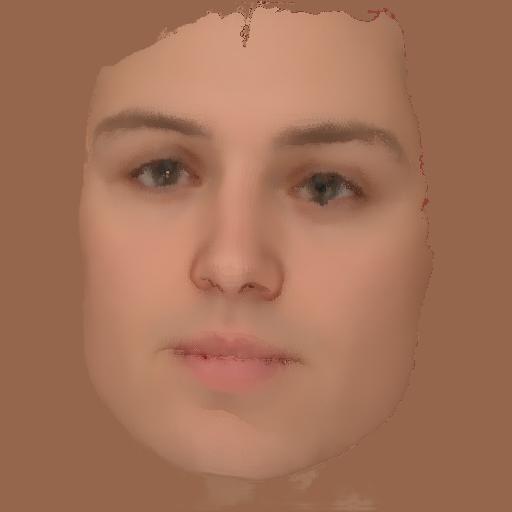}
      \includegraphics[width=0.115\textwidth]{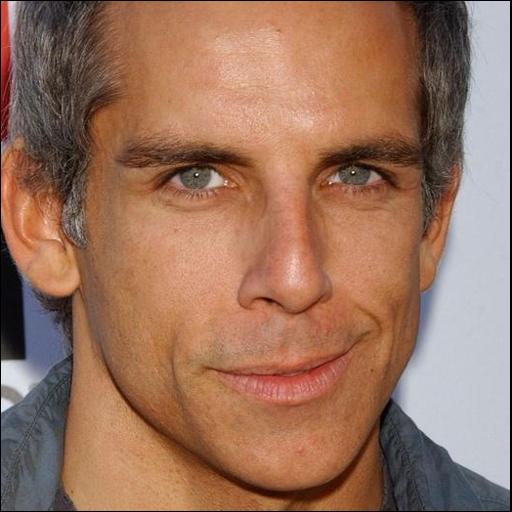}
      \includegraphics[width=0.115\textwidth]{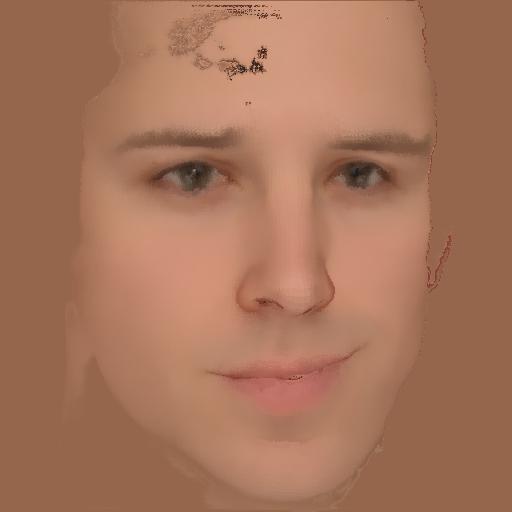}

      \includegraphics[width=0.115\textwidth]{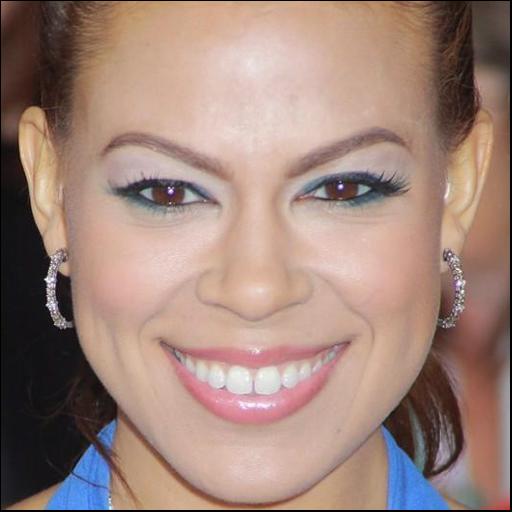}
      \includegraphics[width=0.115\textwidth]{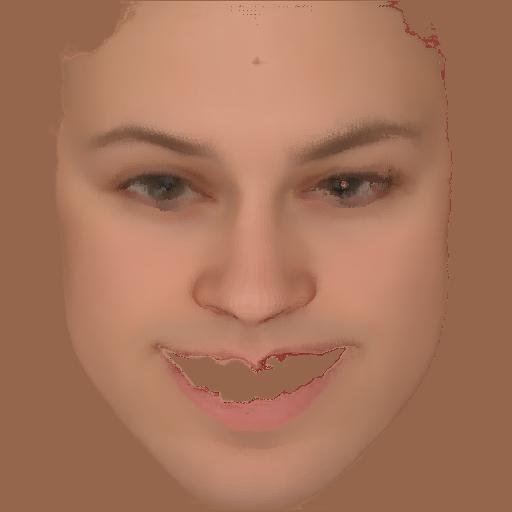}
       \includegraphics[width=0.115\textwidth]{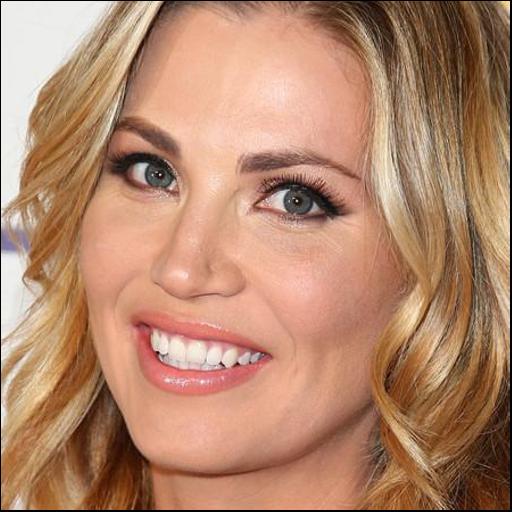}
      \includegraphics[width=0.115\textwidth]{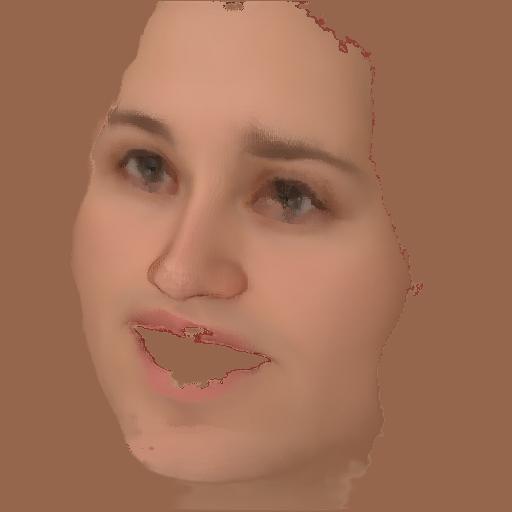}

    \caption{Texture mapping via the embedding.}
    \label{fig:corr_vis}
    \vspace{-0.1cm}
\end{figure}

\begin{figure}[b]
	\setlength{\tabcolsep}{0.5pt}
    \centering
    \begin{tabular}{cccc}
      \includegraphics[height=0.15\textwidth]{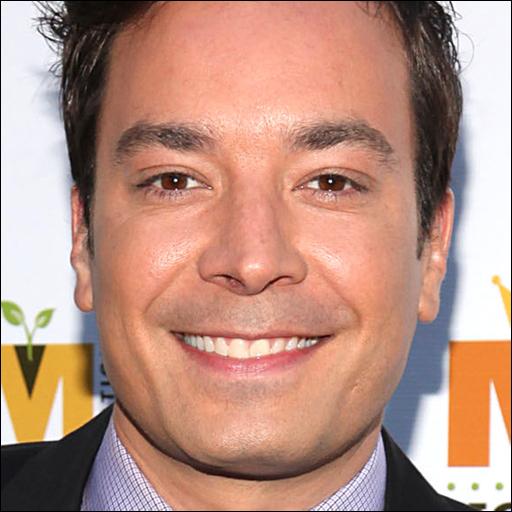}&
      \includegraphics[height=0.15\textwidth]{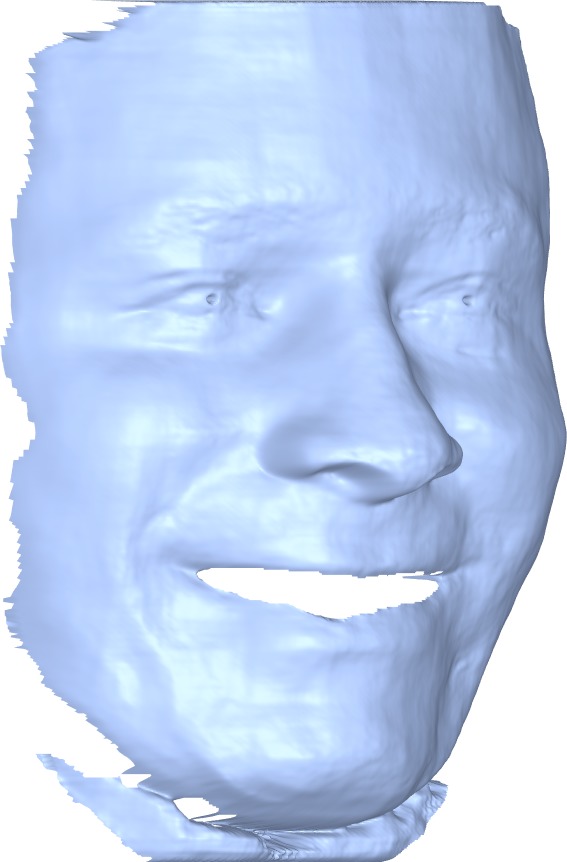}&
      \includegraphics[height=0.15\textwidth]{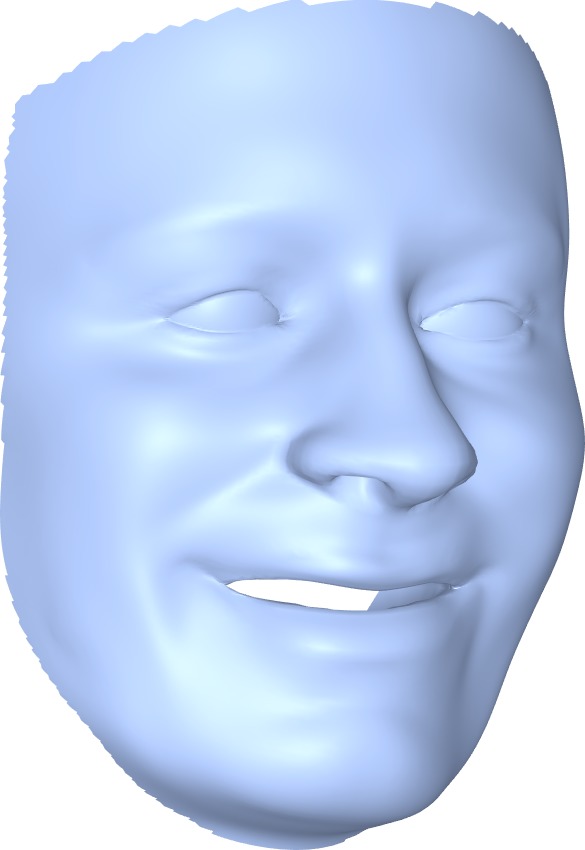}&
      \includegraphics[height=0.15\textwidth]{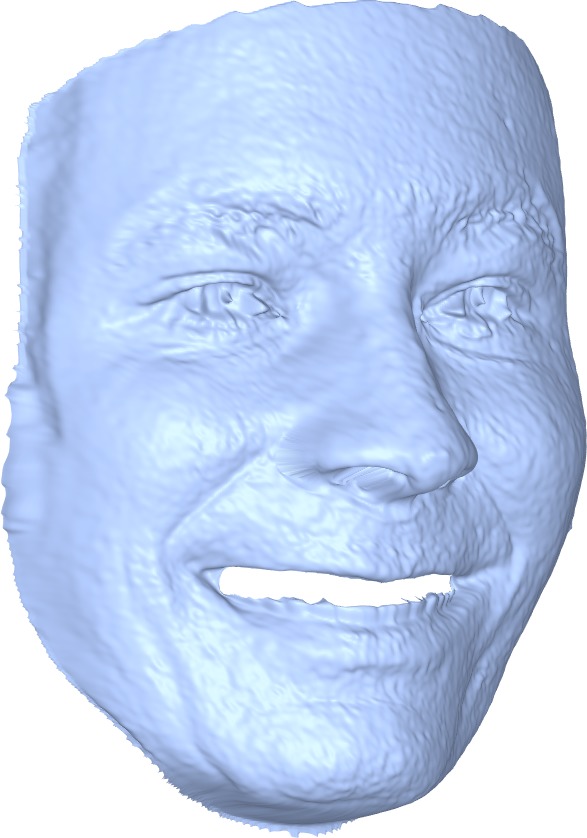}\tabularnewline
      \includegraphics[height=0.15\textwidth]{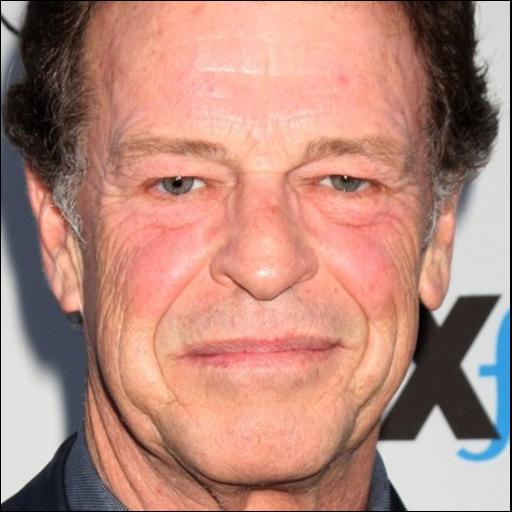}&
      \includegraphics[height=0.15\textwidth]{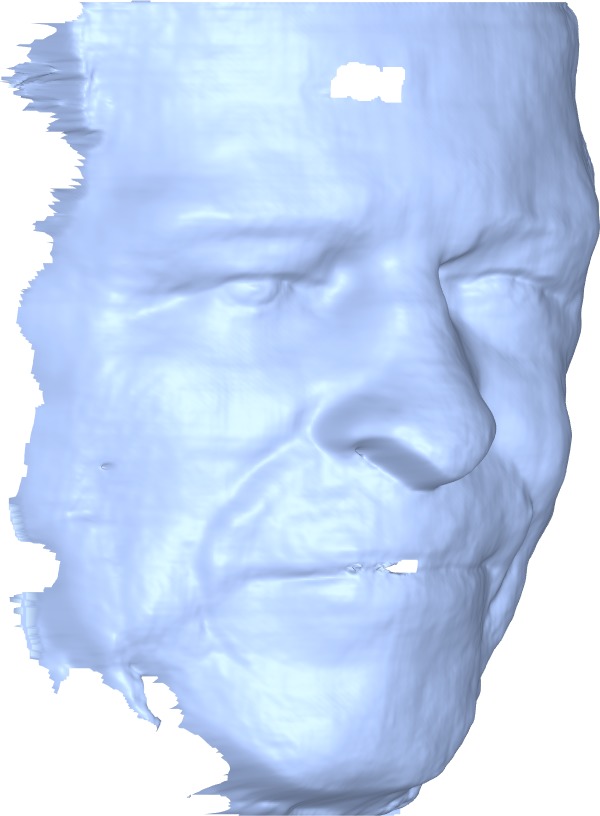}&
      \includegraphics[height=0.15\textwidth]{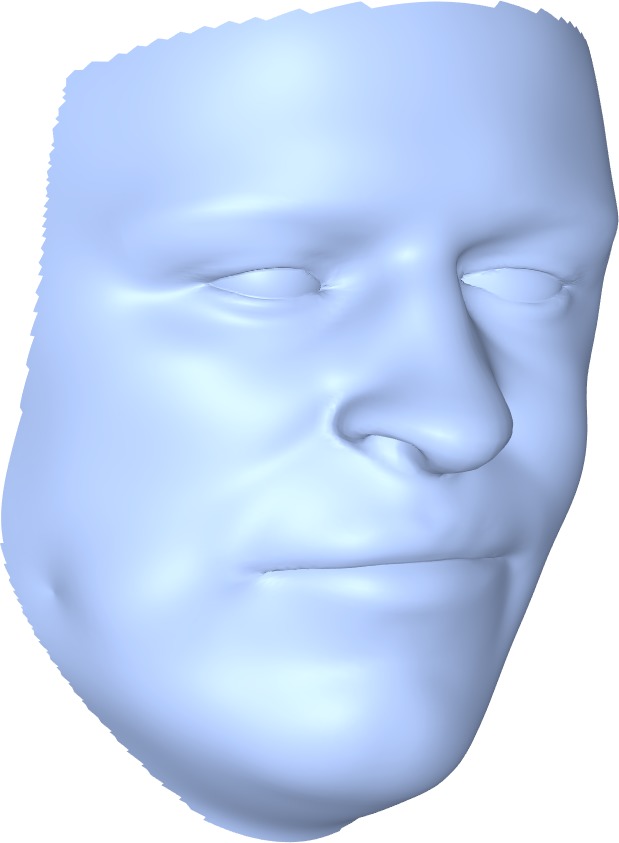}&
      \includegraphics[height=0.15\textwidth]{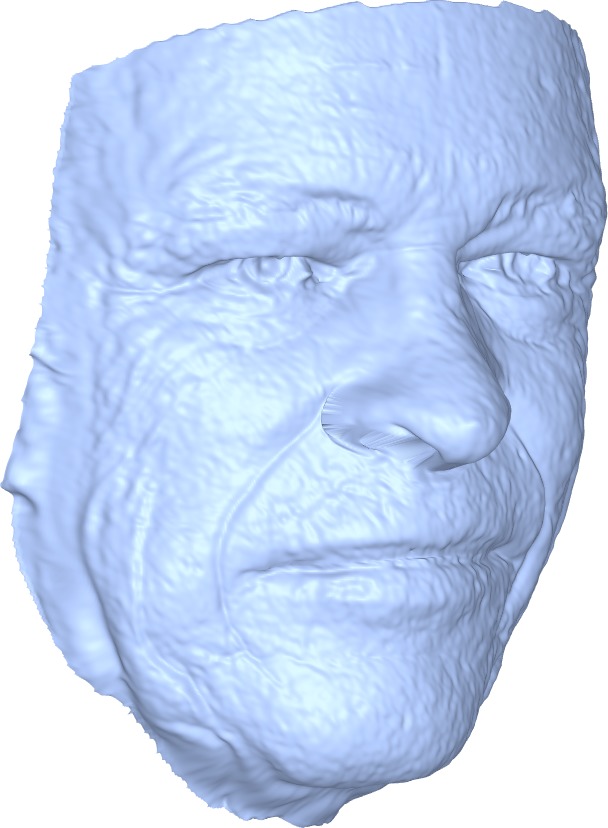}\tabularnewline
      \includegraphics[height=0.15\textwidth]{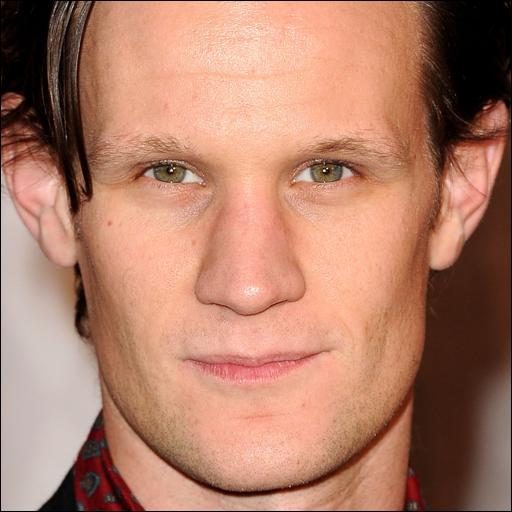}&
      \includegraphics[height=0.15\textwidth]{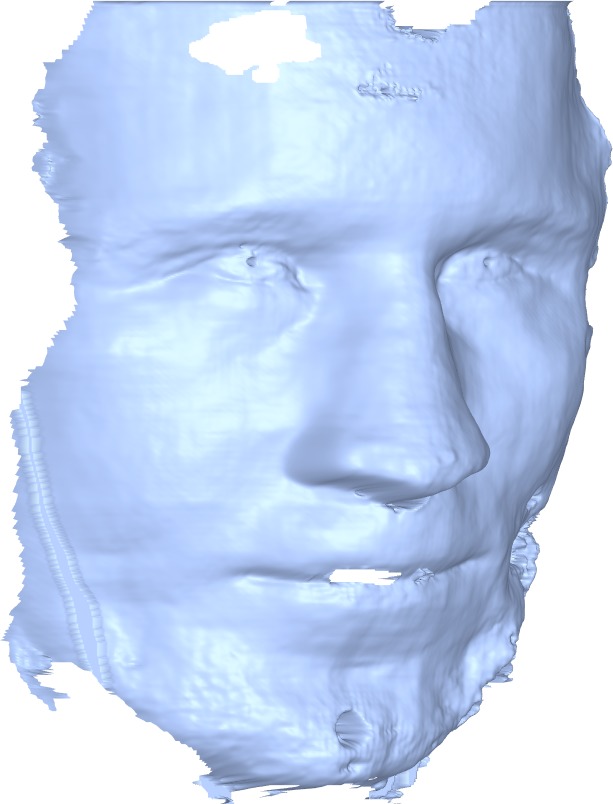}&
      \includegraphics[height=0.15\textwidth]{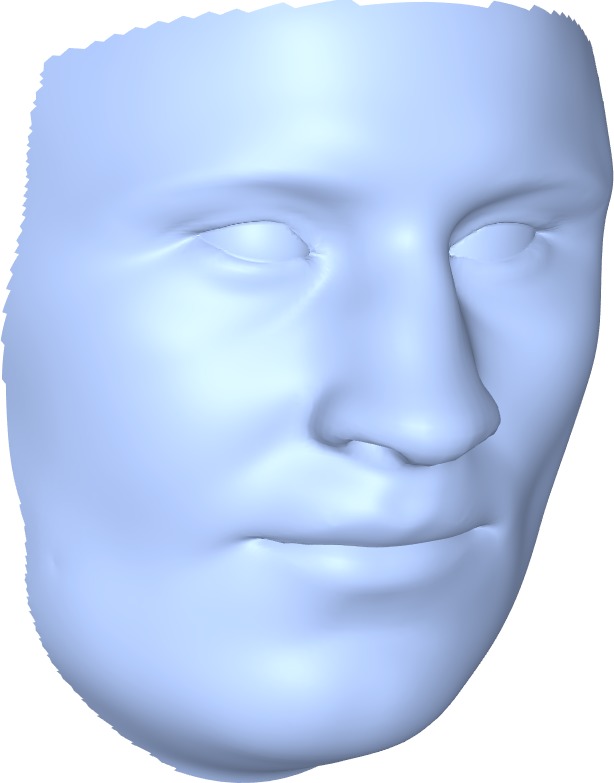}&
      \includegraphics[height=0.15\textwidth]{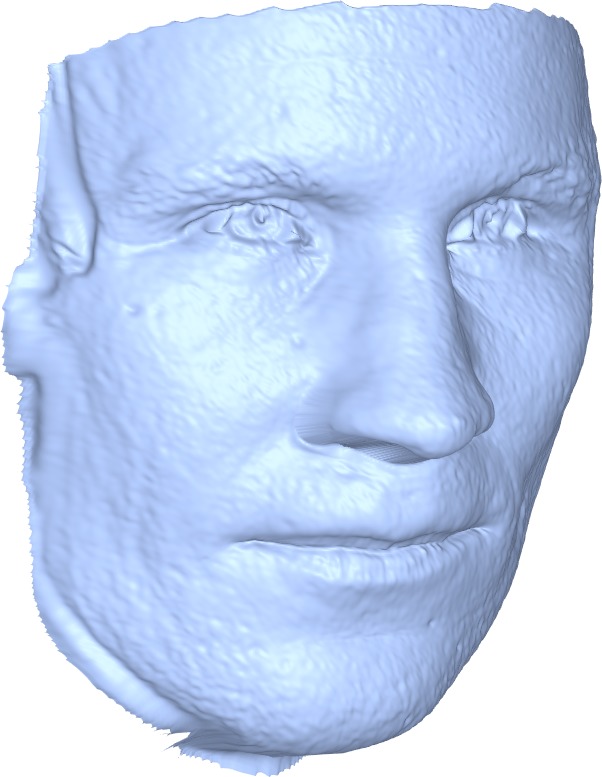}\tabularnewline
      \includegraphics[height=0.15\textwidth]{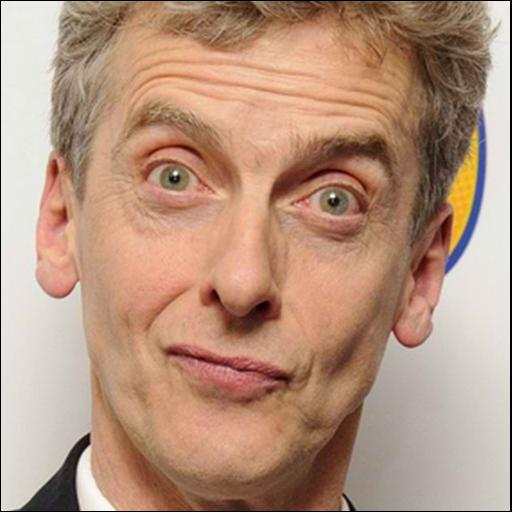}&
      \includegraphics[height=0.15\textwidth]{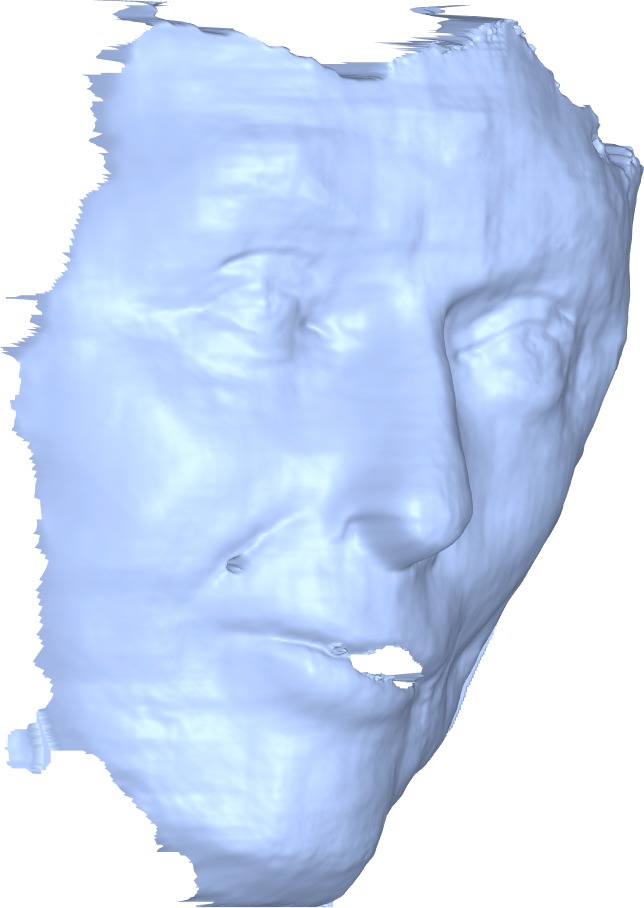}&
      \includegraphics[height=0.15\textwidth]{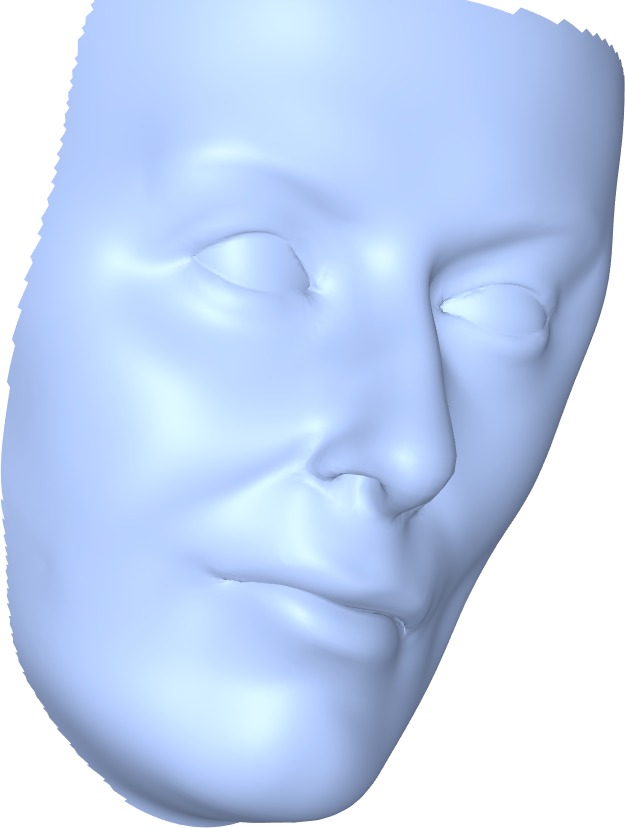}&
      \includegraphics[height=0.15\textwidth]{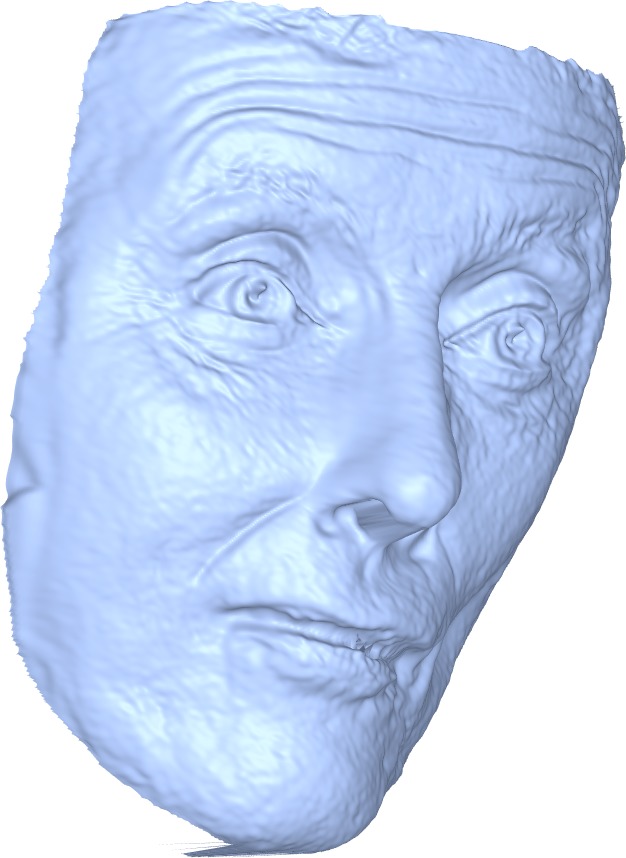}\tabularnewline
      \end{tabular}
    \caption{The reconstruction stages. From left to right: the input image, the reconstruction of the network, the registered template and the final shape.}
    \label{fig:stages}
\end{figure}

\begin{figure*} %
	\setlength{\tabcolsep}{2pt}
    \centering
    \begin{tabular}{ccccccccc}
 \includegraphics[height=0.13\textwidth]{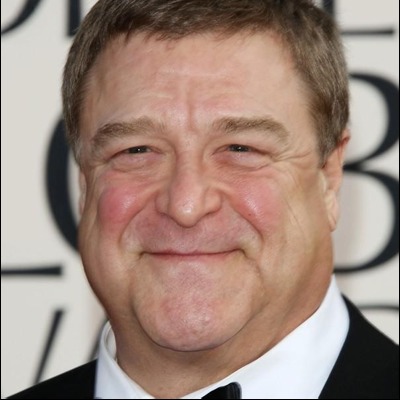}&
 \includegraphics[height=0.13\textwidth]{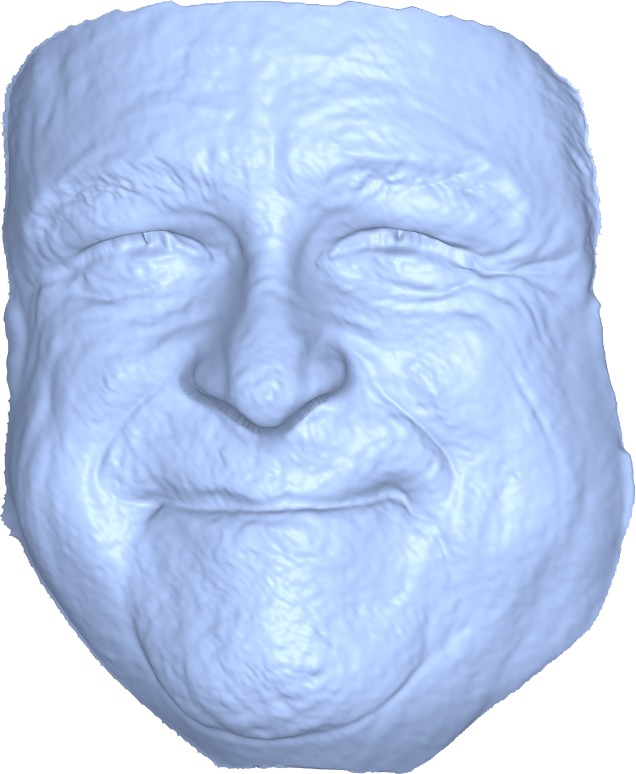}&
 \includegraphics[height=0.13\textwidth]{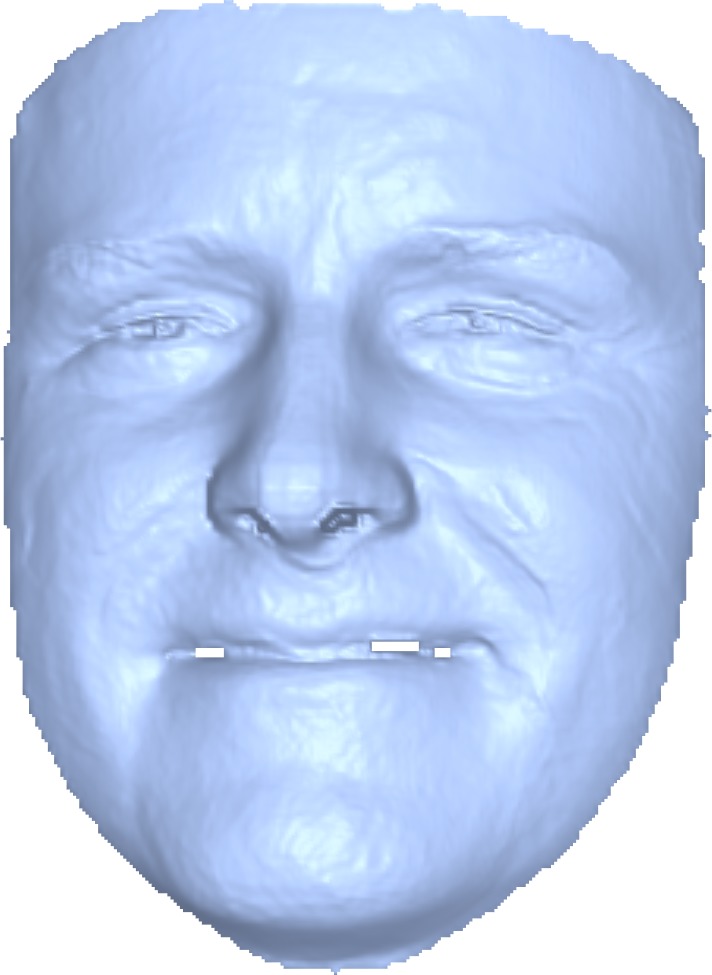}&
 \includegraphics[height=0.13\textwidth]{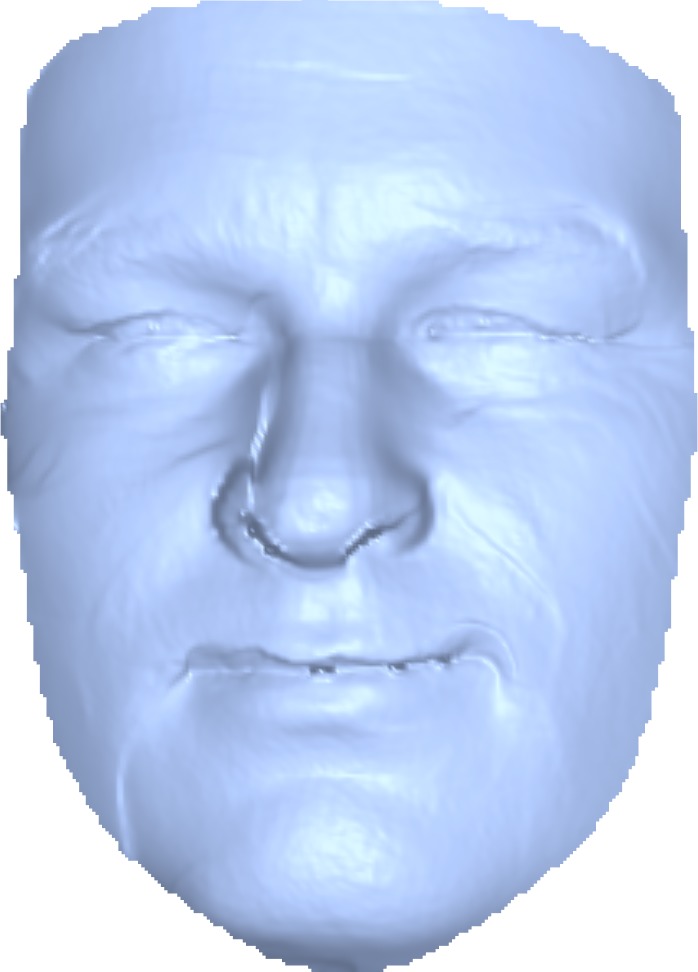}&
 \includegraphics[height=0.13\textwidth]{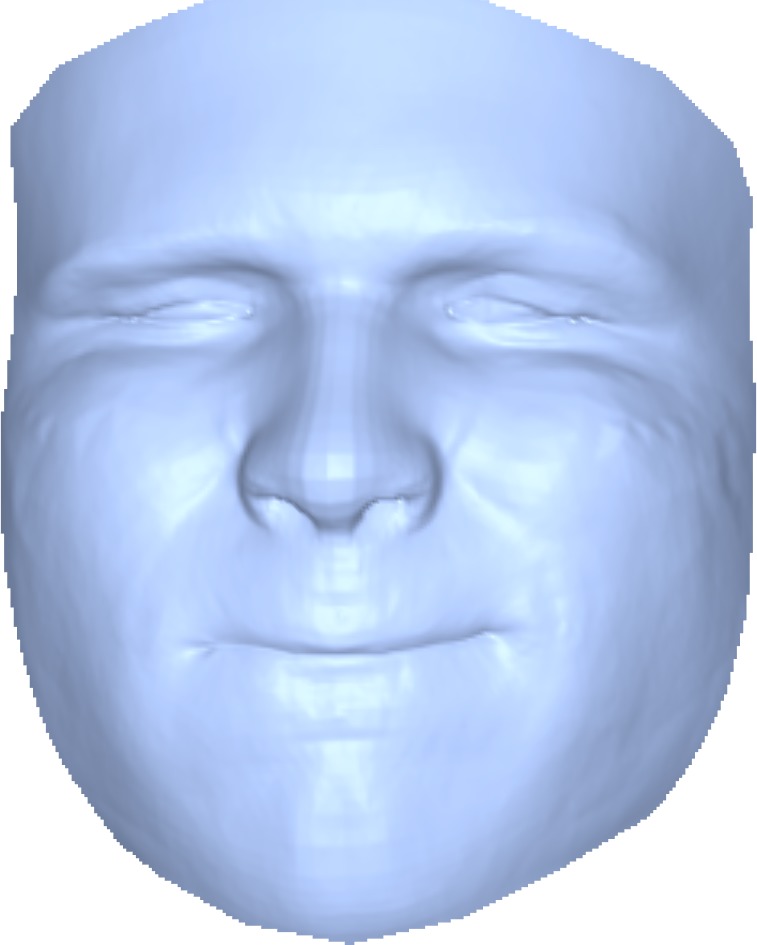}&
 \includegraphics[height=0.13\textwidth]{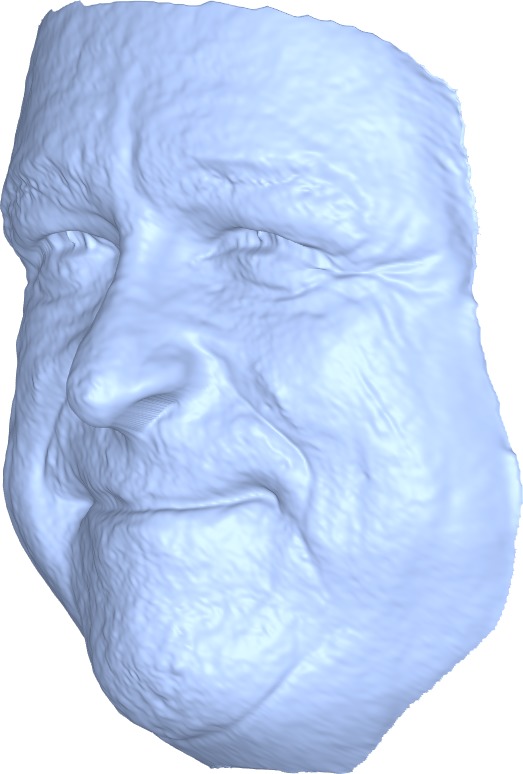}&
 \includegraphics[height=0.13\textwidth]{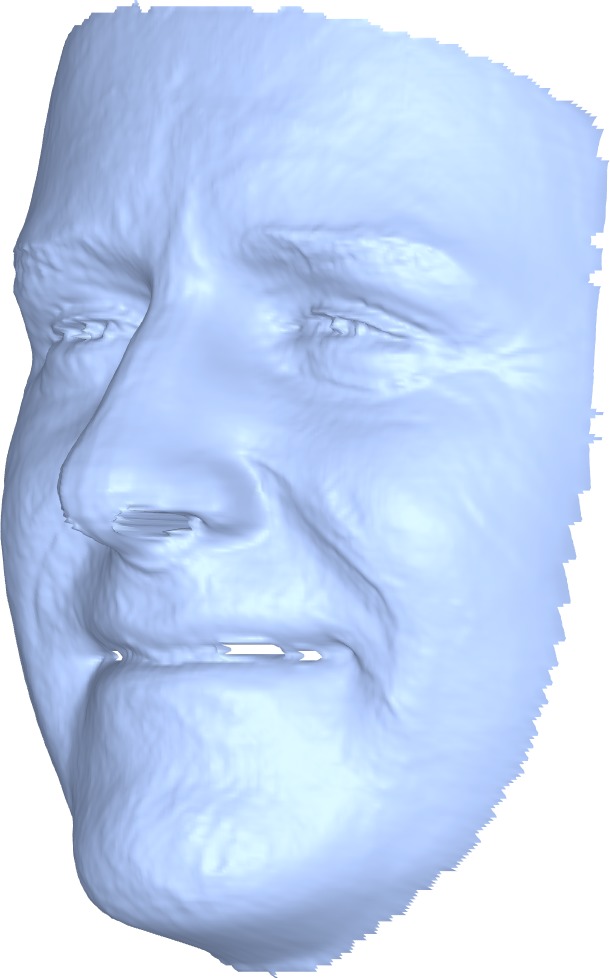}&
 \includegraphics[height=0.13\textwidth]{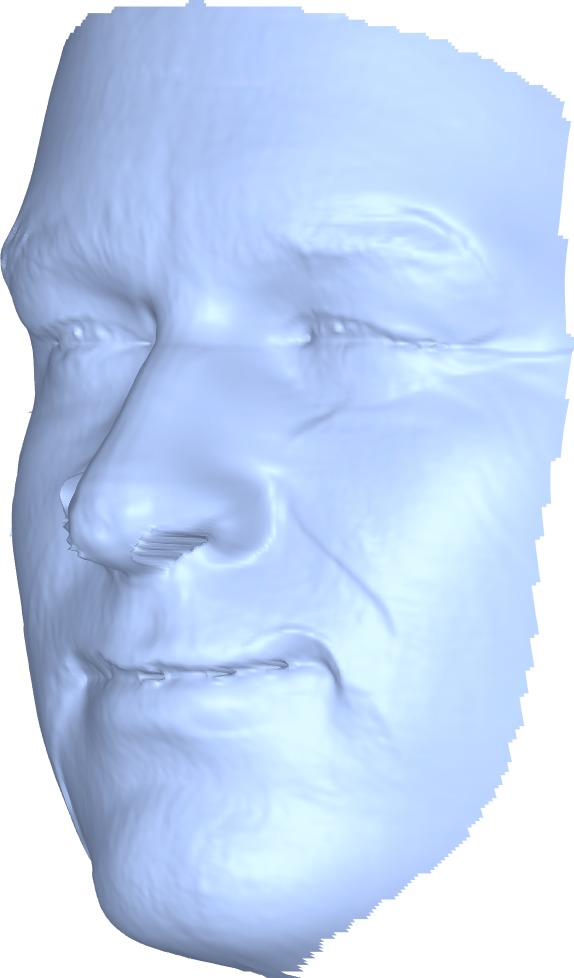}&
 \includegraphics[height=0.13\textwidth]{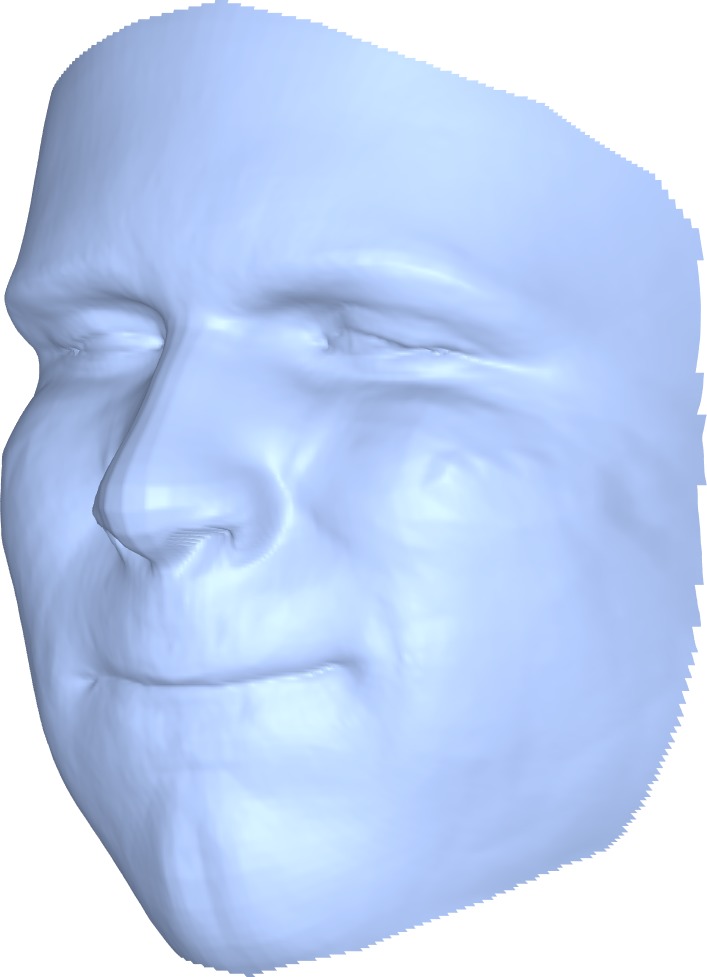}\tabularnewline
 \includegraphics[height=0.13\textwidth]{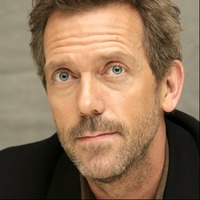}&
 \includegraphics[height=0.13\textwidth]{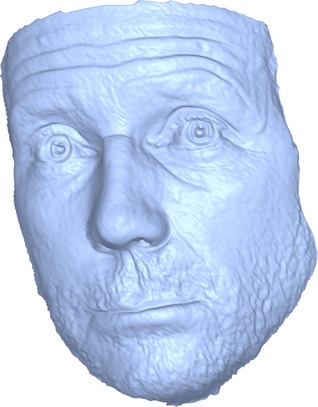}&
 \includegraphics[height=0.13\textwidth]{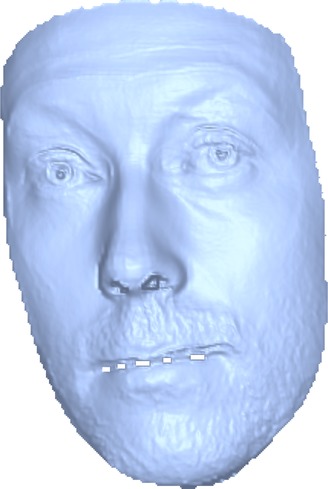}&
 \includegraphics[height=0.13\textwidth]{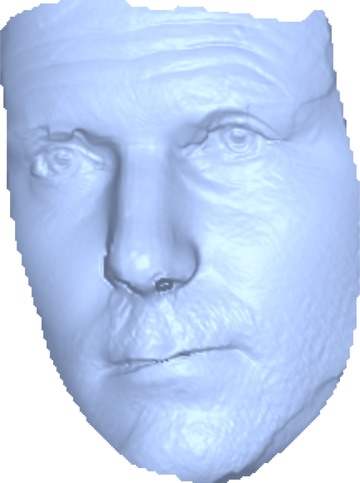}&
 \includegraphics[height=0.13\textwidth]{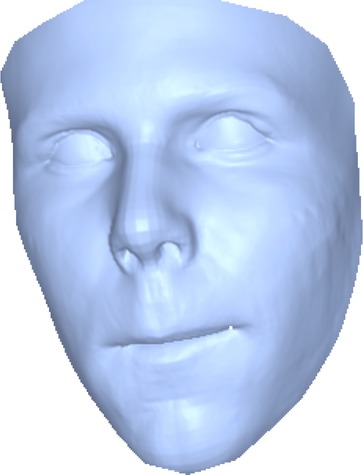}&
 \includegraphics[height=0.13\textwidth]{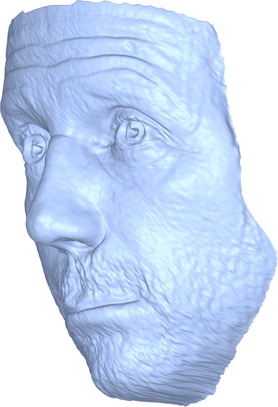}&
 \includegraphics[height=0.13\textwidth]{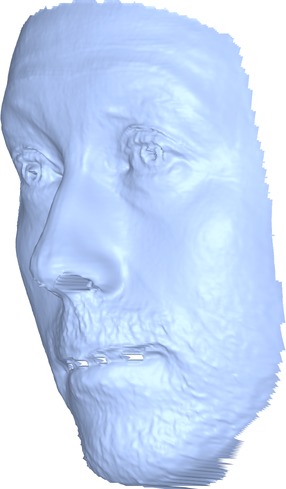}&
 \includegraphics[height=0.13\textwidth]{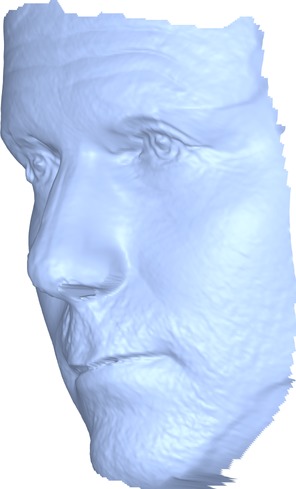}&
 \includegraphics[height=0.13\textwidth]{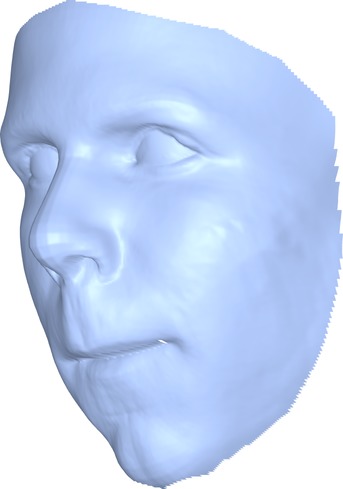}\tabularnewline
 \includegraphics[height=0.13\textwidth]{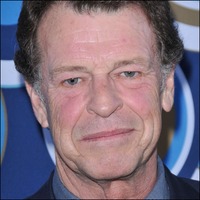}&
 \includegraphics[height=0.13\textwidth]{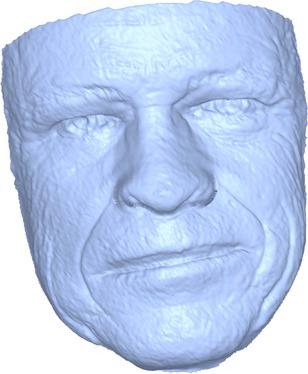}&
 \includegraphics[height=0.13\textwidth]{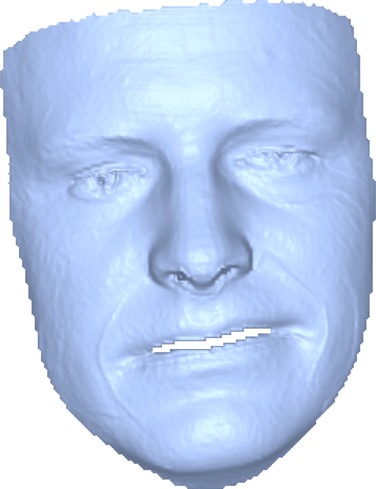}&
 \includegraphics[height=0.13\textwidth]{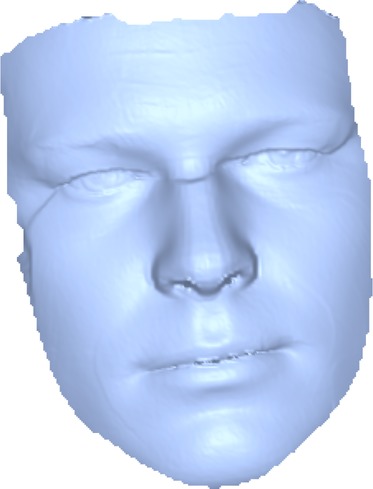}&
 \includegraphics[height=0.13\textwidth]{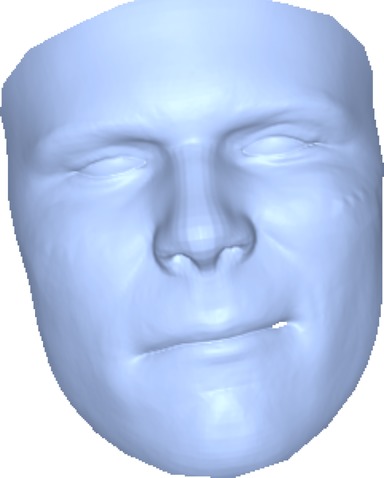}&
 \includegraphics[height=0.13\textwidth]{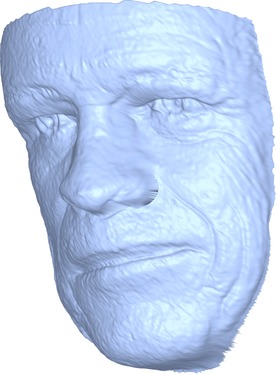}&
 \includegraphics[height=0.13\textwidth]{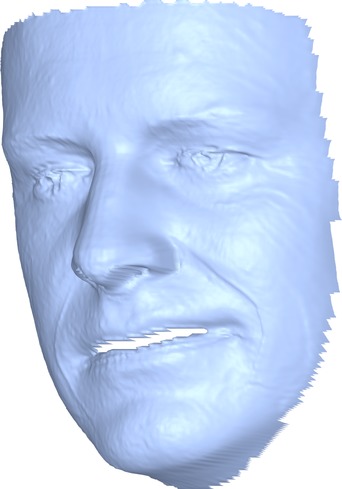}&
 \includegraphics[height=0.13\textwidth]{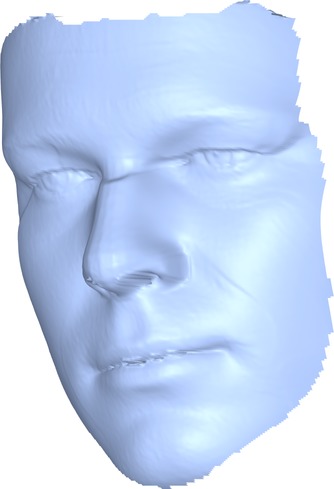}&
 \includegraphics[height=0.13\textwidth]{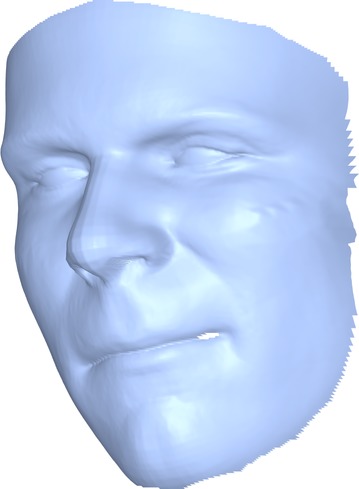}\tabularnewline
  \includegraphics[height=0.13\textwidth]{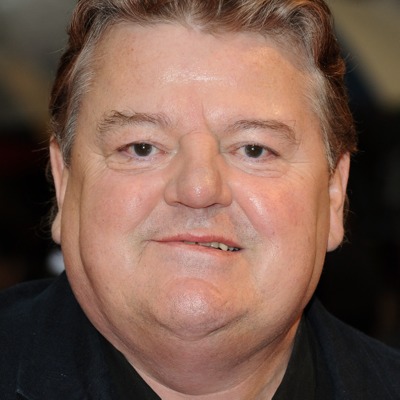}&
 \includegraphics[height=0.13\textwidth]{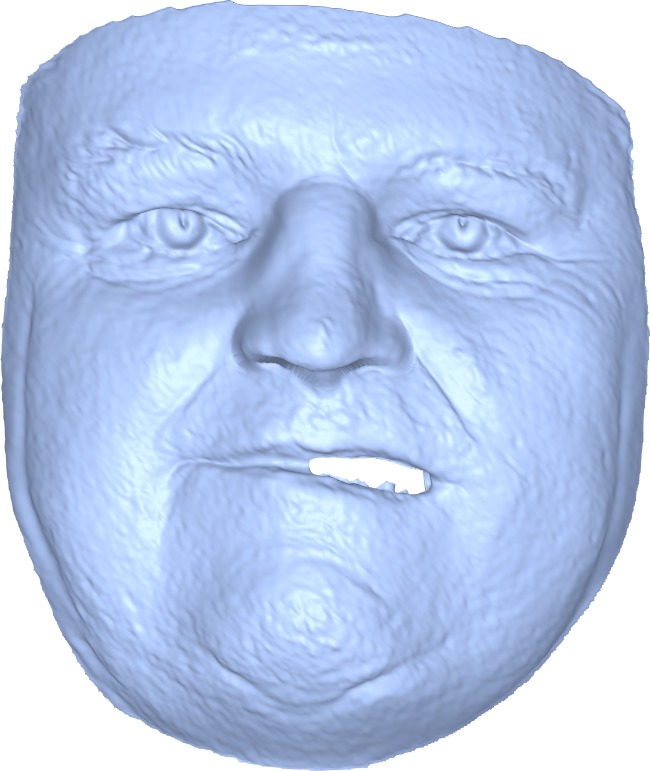}&
 \includegraphics[height=0.13\textwidth]{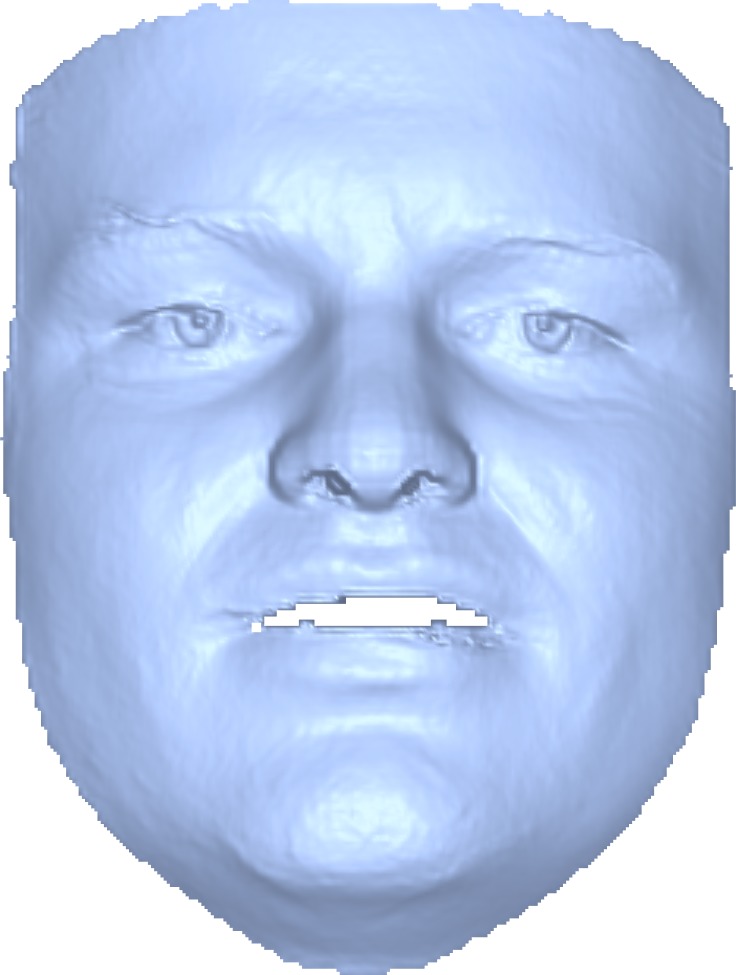}&
 \includegraphics[height=0.13\textwidth]{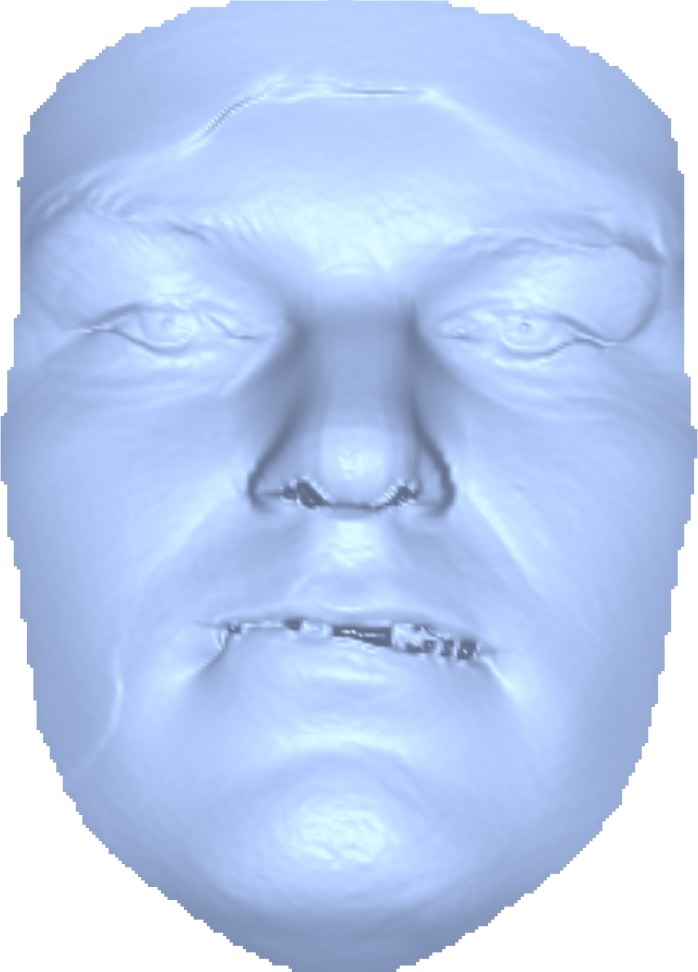}&
 \includegraphics[height=0.13\textwidth]{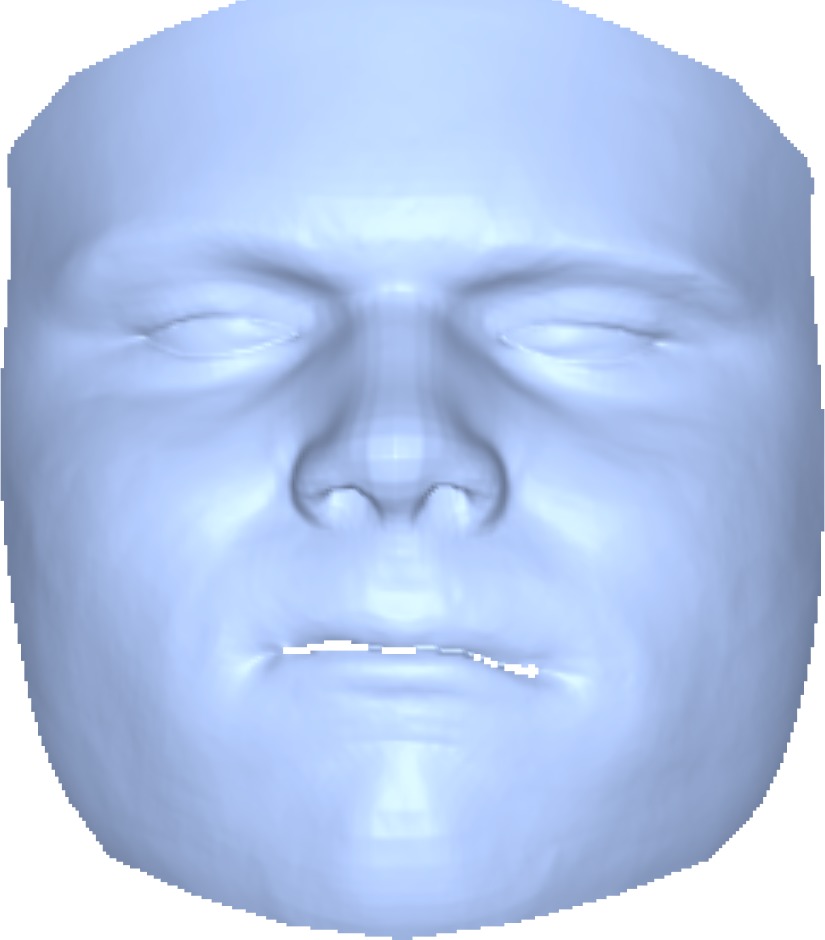}&
 \includegraphics[height=0.13\textwidth]{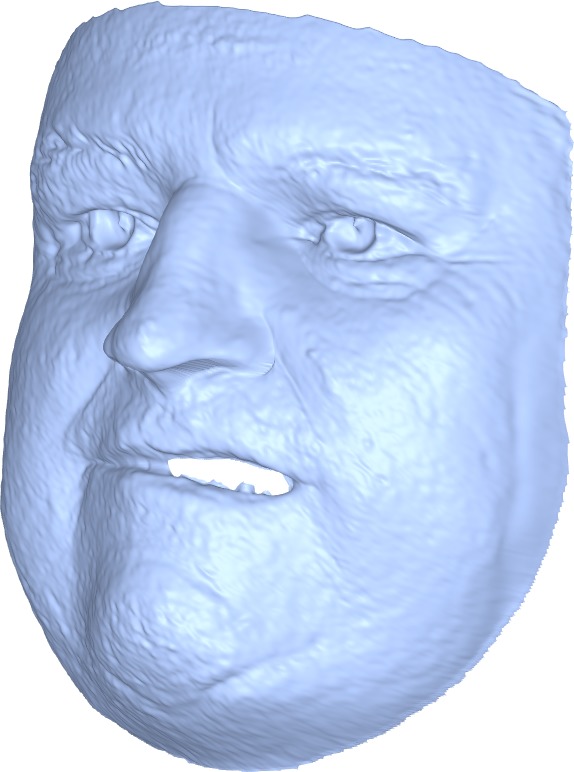}&
 \includegraphics[height=0.13\textwidth]{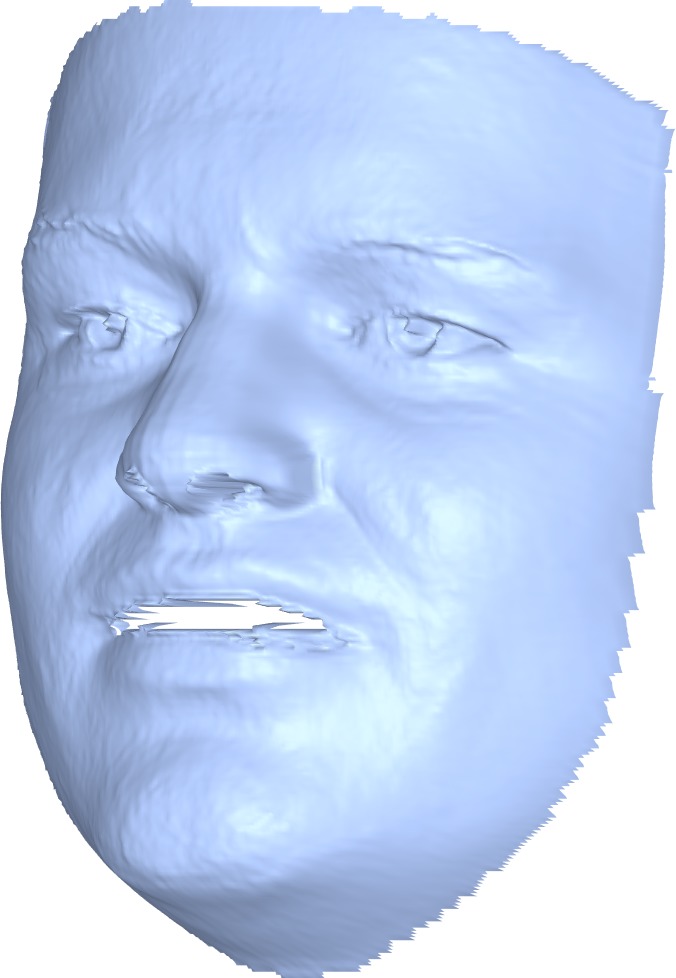}&
 \includegraphics[height=0.13\textwidth]{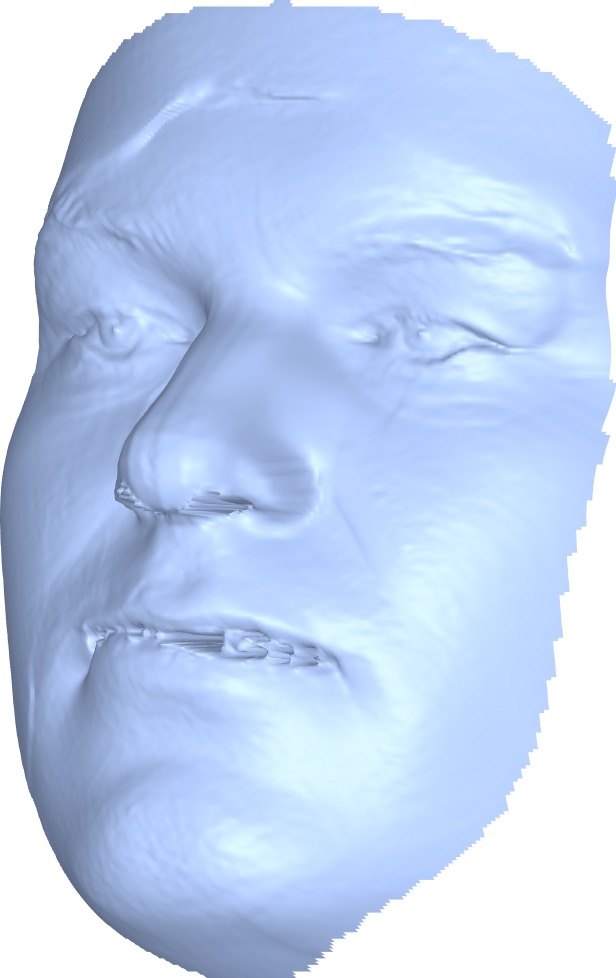}&
 \includegraphics[height=0.13\textwidth]{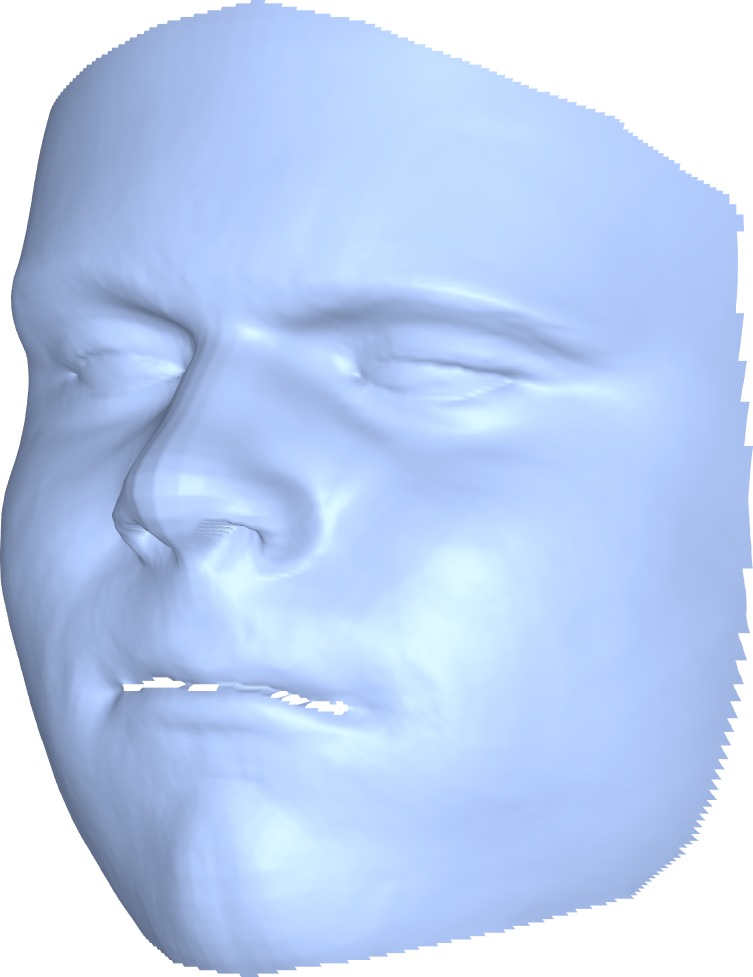}\tabularnewline
 Input & Proposed & \cite{richardson2016learning} & \cite{kemelmacher20113d} & \cite{zhu2015high} & Proposed & \cite{richardson2016learning} & \cite{kemelmacher20113d} & \cite{zhu2015high}
\end{tabular}
    \caption{Qualitative comparison. Input images are presented alongside the reconstructions of the different methods.}
    \label{fig:qual_results}
\end{figure*}

\subsection{Quantitative Evaluation}
For a quantitative comparison, we used the first 200 subjects from the BU-3DFE dataset~\cite{yin20063d}, which contains facial images aligned with ground truth depth images.
Each method provides its own estimation for the depth image alongside a binary mask, representing the valid pixels to be taken into account in the evaluation.
Obviously, since the problem of reconstructing depth from a single image is ill-posed, the estimation needs to be judged up to global scaling and transition along the depth axis.
Thus, we compute these paramters using the Random Sample Concensus (RANSAC) approach~\cite{fischler1981random}, for normalizing the estimation according to the ground truth depth.
This significantly reduces the absolute error of each method as the global parameter estimation is robust to outliers.
Note that the parameters of the RANSAC were identical for all the methods and samples.
The results of this comparison are given in~\autoref{table:quan_basic}, where the units are given in terms of the percentile of the ground-truth depth range.
As a further analysis of the reconstruction accuracy, we computed the mean absolute error of each method based on expressions, see~\autoref{table:quan_by_expr}.

\begin{figure}
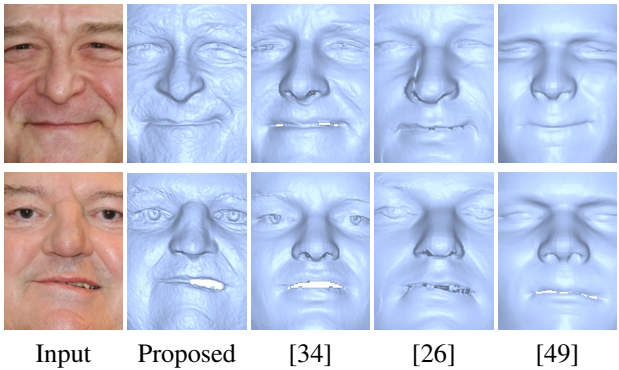

\setlength{\tabcolsep}{1pt}
\centering
\begin{tabular}{ccccc}
\includegraphics[width=0.09\textwidth,trim={4cm 3.3cm 4.5cm 3.3cm},clip]{images/qual_compare/2327-3_input_image}&
\includegraphics[width=0.09\textwidth,trim={3.1cm 4.7cm 5.0cm 3.4cm},clip]{images/qual_compare/2327-3_input_image-depth-out-front-iccv_tight}&
\includegraphics[width=0.09\textwidth,trim={3.2cm 5.8cm 4.5cm 5.2cm},clip]{images/qual_compare/2327-3_input_image-depth-out-front-cvpr_tight}&
\includegraphics[width=0.09\textwidth,trim={3.2cm 5.8cm 4.5cm 5.8cm},clip]{images/qual_compare/2327-3_input_image-depth-out-front-ira_tight}&
\includegraphics[width=0.09\textwidth,trim={3.2cm 4.7cm 4.9cm 3.7cm},clip]{images/qual_compare/2327-3_input_image-depth-out-front-3dmm_tight}\tabularnewline

\includegraphics[width=0.09\textwidth,trim={4cm 3.3cm 4.5cm 3.35cm},clip]{images/qual_compare/RobbieColtrane_input_image}&
\includegraphics[width=0.09\textwidth,trim={3.1cm 4.7cm 5.0cm 2.8cm},clip]{images/qual_compare/RobbieColtrane_input_image-depth-out-front-iccv_tight}&
\includegraphics[width=0.09\textwidth,trim={3.2cm 5.8cm 4.5cm 4.3cm},clip]{images/qual_compare/RobbieColtrane_input_image-depth-out-front-cvpr_tight}&
\includegraphics[width=0.09\textwidth,trim={3.2cm 5.8cm 4.5cm 6.0cm},clip]{images/qual_compare/RobbieColtrane_input_image-depth-out-front-ira_tight}&
\includegraphics[width=0.09\textwidth,trim={4.4cm 4.7cm 7.2cm 5.1cm},clip]{images/qual_compare/RobbieColtrane_input_image-depth-out-front-3dmm_tight}\tabularnewline
Input & Proposed & \cite{richardson2016learning} & \cite{kemelmacher20113d} & \cite{zhu2015high}
\end{tabular}
    \caption{Zoomed qualitative result of first and fourth subjects from~\autoref{fig:qual_results}.}
    \label{fig:qual_results_zoom}
\end{figure}

\begin{table}
\centering
\begin{tabular}{|c|c|c|c|c|}
\hline
 & Mean Err. & Std Err. & Median Err. & 90\% Err. \\
\hline
\cite{kemelmacher20113d} & 3.89 & 4.14 & 2.94 & 7.34 \\
\hline
\cite{zhu2015high} & 3.85 & 3.23 & 2.93 & 7.91 \\
\hline
\cite{richardson2016learning} & 3.61 & 2.99 & 2.72 & 6.82 \\
\hline
Ours & \textbf{3.51} & \textbf{2.69} & \textbf{2.65} & \textbf{6.59} \\
\hline
\end{tabular}
\caption{Quantitative evaluation on the BU-3DFE Dataset. From left to right: the absolute depth errors evaluated by mean, standard deviation, median and the average ninety percent largest error.}
\label{table:quan_basic}
\end{table}

\begin{table}
\centering
\begin{tabular}{|c|c|c|c|c|c|c|c|}
\hline
 & AN & DI & FE & HA & NE & SA & SU \\
\hline
\cite{kemelmacher20113d} & 3.47 & 4.03 & 3.94 & 4.30 & 3.43 & 3.52 & 4.19 \\
\hline
\cite{zhu2015high} & 4.00 & 3.93 & 3.91 & 3.70 & 3.76 & 3.61 & \textbf{3.96} \\
\hline
\cite{richardson2016learning} & \textbf{3.42} & 3.46 & 3.64 & 3.41 & 4.22 & 3.59 & 4.00 \\
\hline
Ours & 3.67 & \textbf{3.34} & \textbf{3.36} & \textbf{3.01} & \textbf{3.17} & \textbf{3.37} & 4.41 \\
\hline
\end{tabular}
\caption{The mean error by expression. From left to right: Anger, Disgust, Fear, Happy, Neutral, Sad, Surprise.}
\label{table:quan_by_expr}
\end{table}

\begin{figure}
	\setlength{\tabcolsep}{1pt}
    \centering
    \begin{tabular}{ccccc}
      \includegraphics[height=0.13\textwidth]{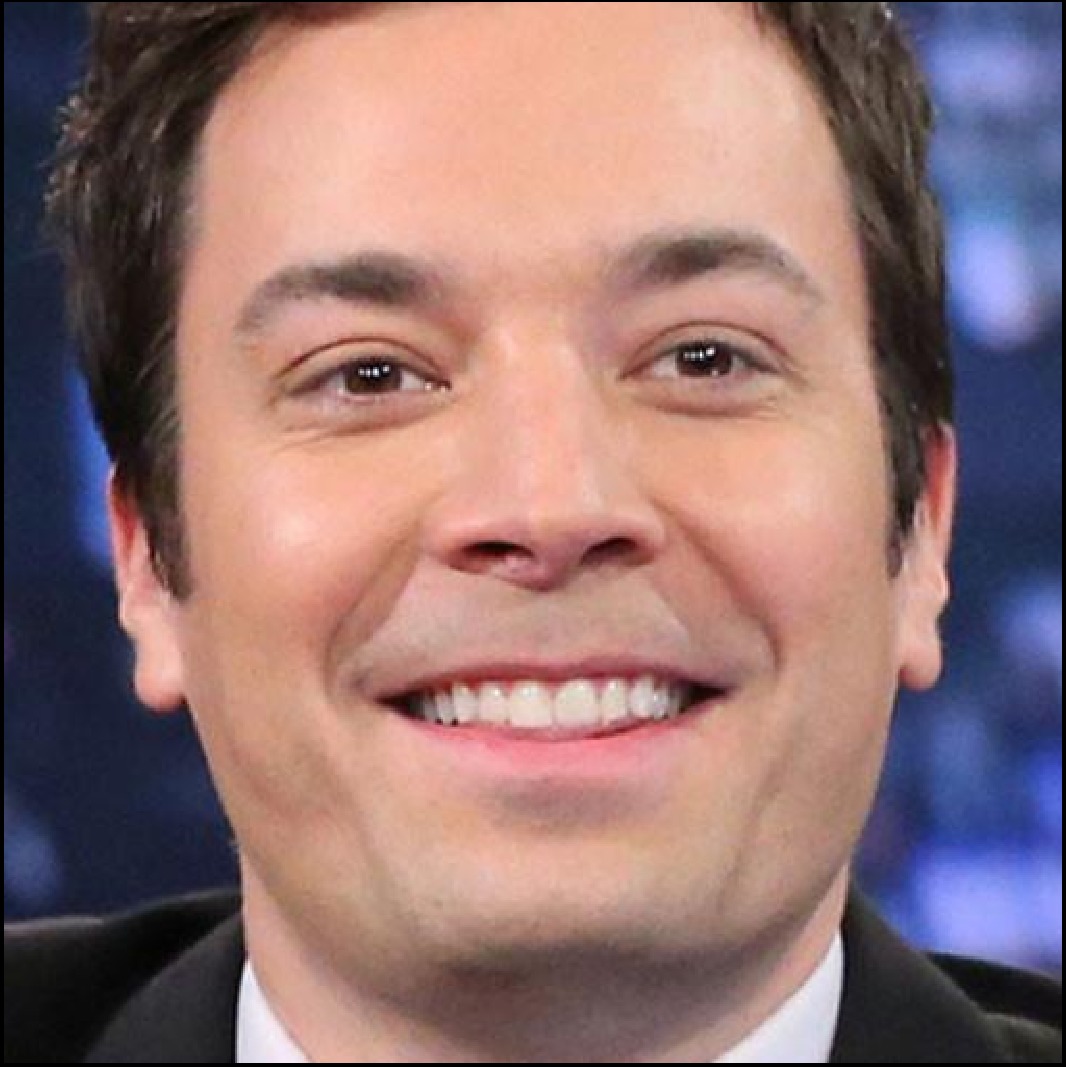}&
      \includegraphics[height=0.13\textwidth]{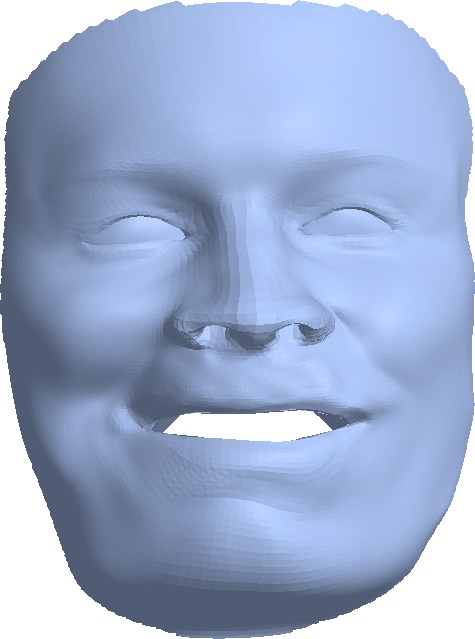}&
      \includegraphics[height=0.13\textwidth]{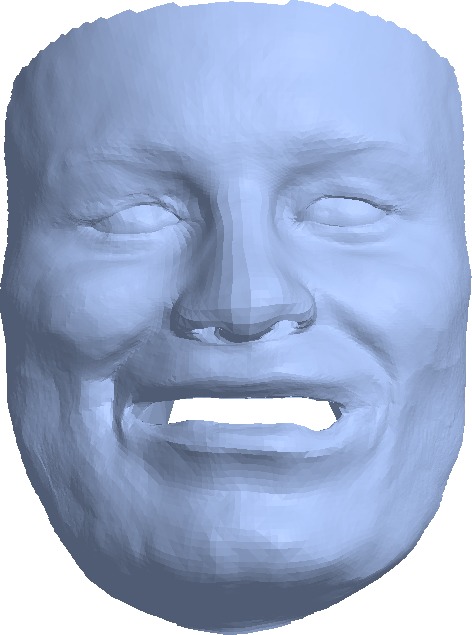}&
      \includegraphics[height=0.13\textwidth,trim={6cm 10.7cm 10.2cm 10.1cm},clip]{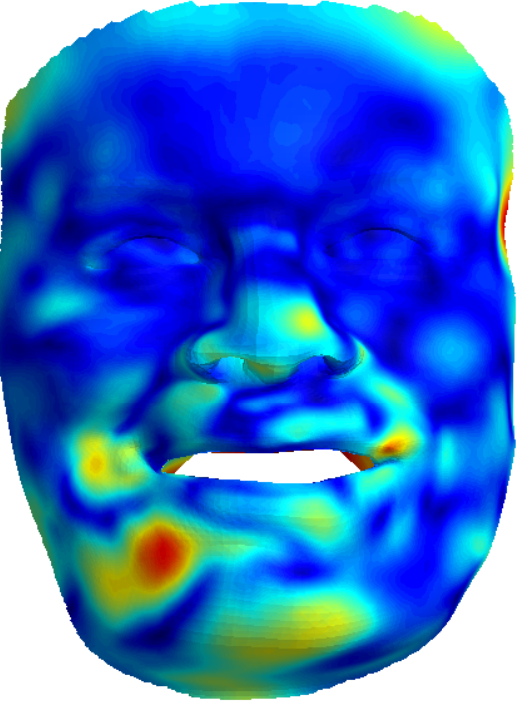}&
      \includegraphics[height=0.13\textwidth,trim={13cm 10.6cm 7cm 10cm},clip]{images/qual_3dmm/1809-15-err}\tabularnewline
      \includegraphics[height=0.13\textwidth]{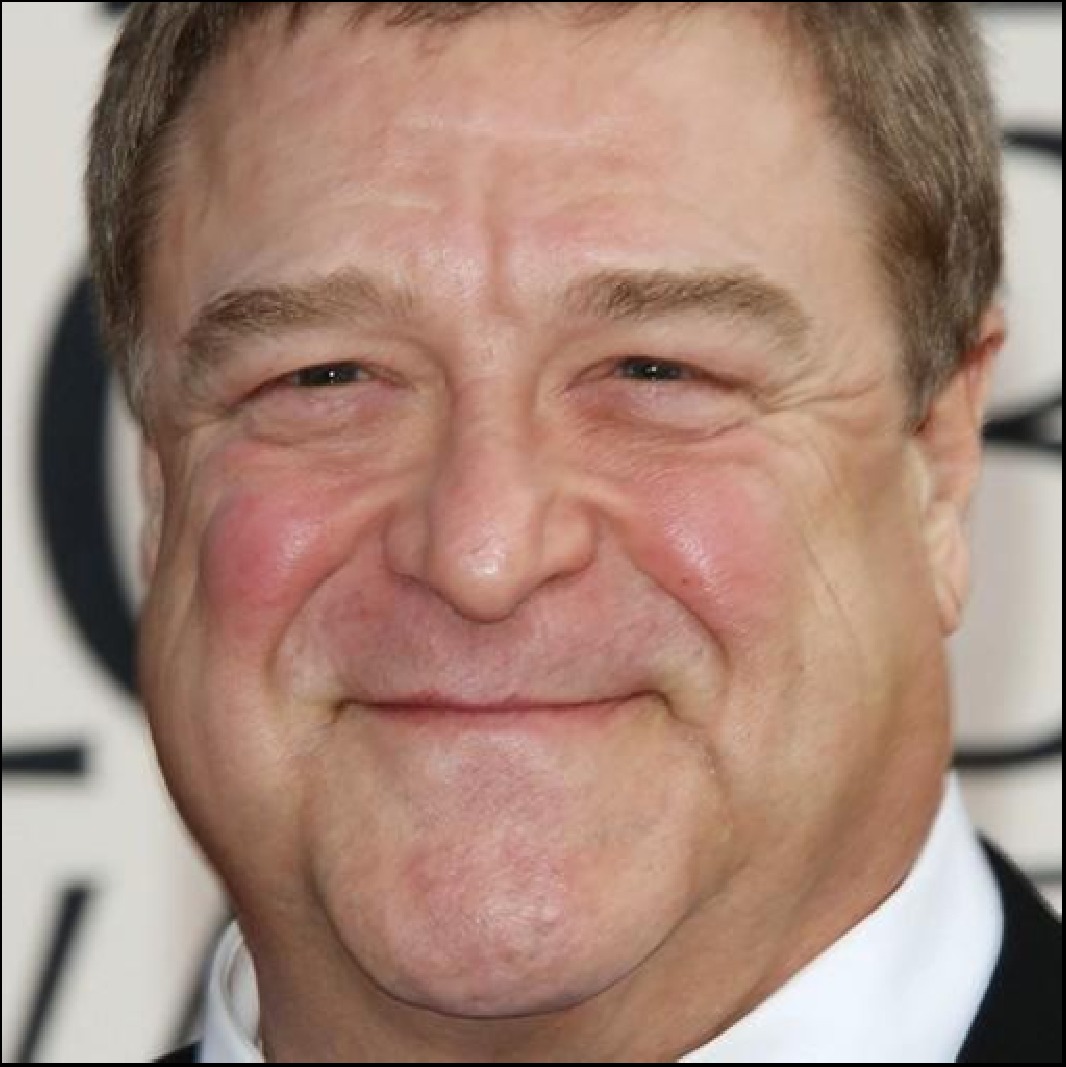}&
      \includegraphics[height=0.13\textwidth]{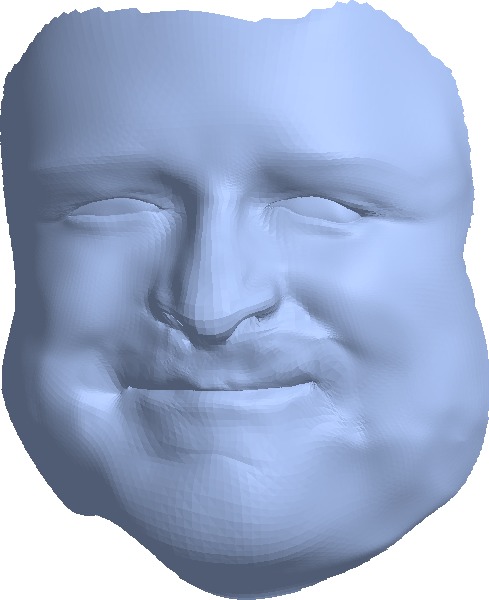}&
      \includegraphics[height=0.13\textwidth]{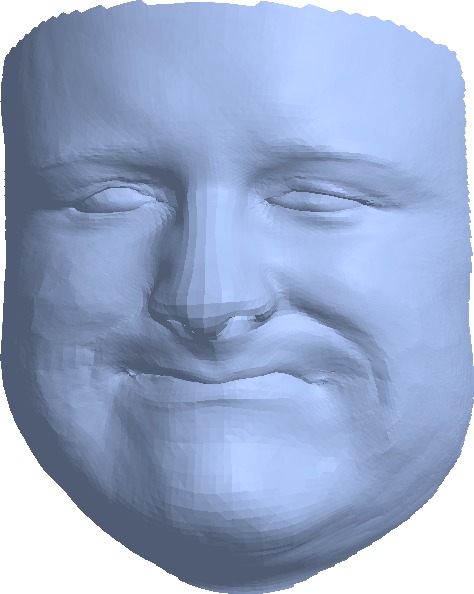}&
      \includegraphics[height=0.13\textwidth,trim={6cm 10.7cm 9.7cm 10.1cm},clip]{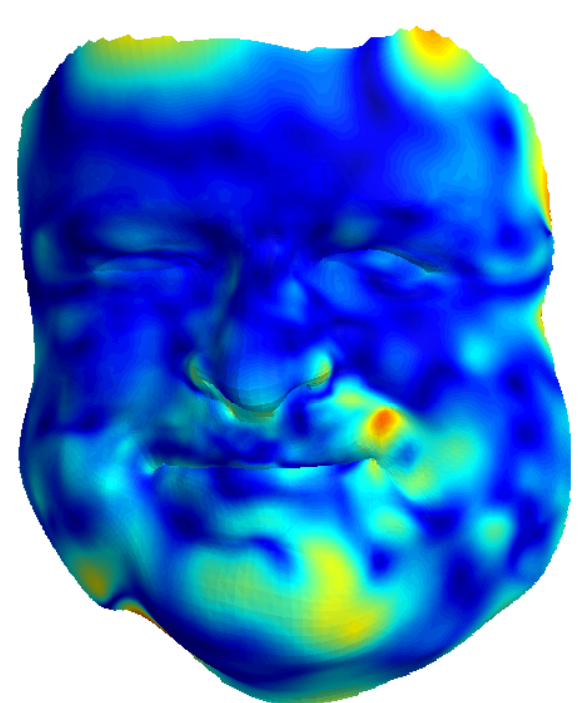}&
      \includegraphics[height=0.13\textwidth,trim={13cm 10.6cm 7cm 10cm},clip]{images/qual_3dmm/1809-15-err}\tabularnewline
      \end{tabular}
    \caption{3DMM Projection. From left to right: the input image, the registered template, the projected mesh and the projection error.}
    \label{fig:3dmm_compare}
    \vspace{-0.4cm}
\end{figure}

\subsection{The Network as a Geometric Constraint}
As demonstrated by the results, the proposed network successfully learns both the depth and the embedding representations for a variety of images. This representation is the key part behind the reconstruction pipeline. However, it can also be helpful for other face-related tasks. As an example, we show that the network can be used as a geometric constraint for facial image manipulations, such as transforming synthetic images into realistic ones.
This idea is based on recent advances in applying Generative Adversarial Networks (GAN)~\cite{goodfellow2014generative} for domain adaption tasks~\cite{taigman2016unsupervised}.

In the basic GAN framework, a Generator Network (G) learns to map from the source domain, $\mathcal{D}_S$,
 to the target domain $\mathcal{D}_T$, where a Discriminator Network (D) tries to distinguish between generated images and samples from the target domain, by optimizing the following objective
\vspace{-0.1cm}
\begin{align}
\underset{G}{\min}\underset{D}{\max}V\left(D,G\right)
=& \quad \:
\mathbb{E}_{y\sim\mathcal{D}_{T}}\left[\log D\left(y\right)\right]  \\
 &  +  \mathbb{E}_{x\sim\mathcal{D}_{S}}
        \left[\log\left(1-D\left(G\left(x\right)\right)\right)\right]. \nonumber
\end{align}

Theoretically, this framework could also translate images from the synthetic domain into the realistic one. However, it does not guarantee that the underlying geometry of the synthetic data is preserved throughout that transformation.
That is, the generated image might look realistic, but have a completely different facial structure from the synthetic input.
To solve that potential inconsistency, we suggest to involve the proposed network as an additional loss function on the output of the generator.
\vspace{-0.1cm}
\begin{align}
L_{Geom}\left(x\right)=\left\Vert Net\left(x\right)
    -Net\left(G\left(x\right)\right)\right\Vert _{1},
\end{align}
where $Net(\cdot)$ represents the operation of the introduced network. Note that this is feasible, thanks to the fact that the proposed network is fully differentiable.
The additional geometric fidelity term forces the generator to learn a mapping that makes a synthetic image more realistic while keeping the underlying geometry intact. This translation process could potentially be useful for data generation procedures, similarly to~\cite{shrivastava2016learning}.
Some successful translations are visualized in~\autoref{fig:adapt_results}.
Notice that the network implicitly learns to add facial hair and teeth, and modify the texture the and shading, without changing the facial structure.
As demonstrated by this analysis, the proposed network learns a strong representation that has merit not only for reconstruction, but for other tasks as well.

\begin{figure}
    \centering
     \includegraphics[width=0.091\textwidth]{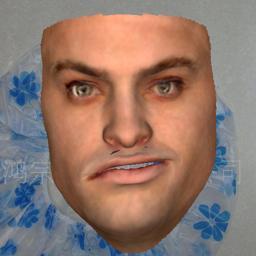}
      \includegraphics[width=0.091\textwidth]{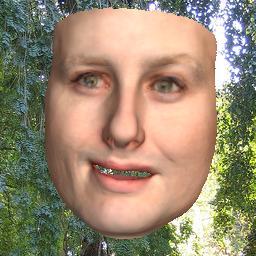}
      \includegraphics[width=0.091\textwidth]{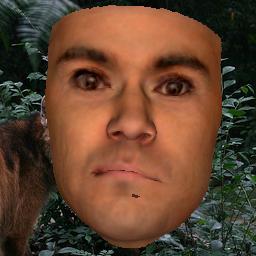}
      \includegraphics[width=0.091\textwidth]{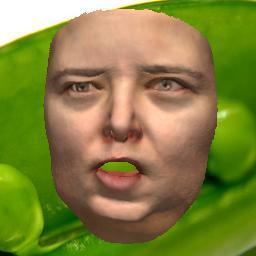}
      \includegraphics[width=0.091\textwidth]{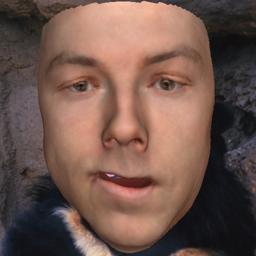}

     \includegraphics[width=0.091\textwidth]{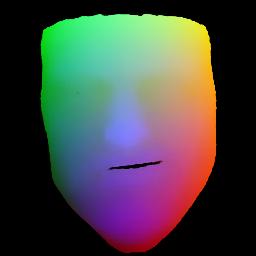}
      \includegraphics[width=0.091\textwidth]{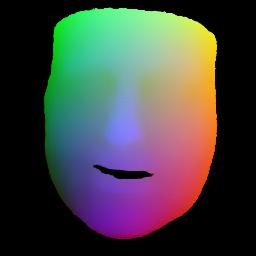}
      \includegraphics[width=0.091\textwidth]{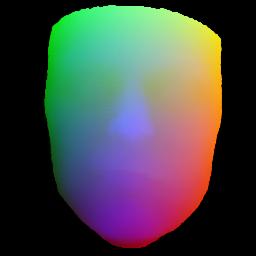}
      \includegraphics[width=0.091\textwidth]{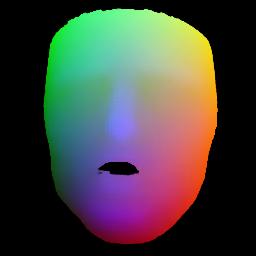}
      \includegraphics[width=0.091\textwidth]{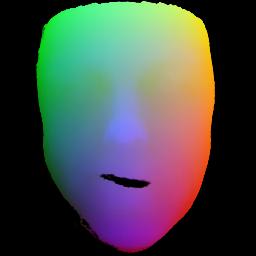}

     \includegraphics[width=0.091\textwidth]{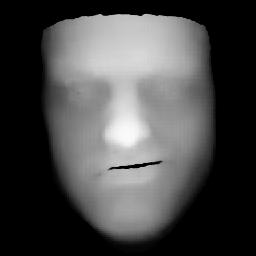}
      \includegraphics[width=0.091\textwidth]{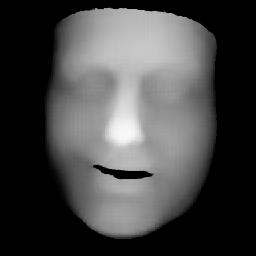}
      \includegraphics[width=0.091\textwidth]{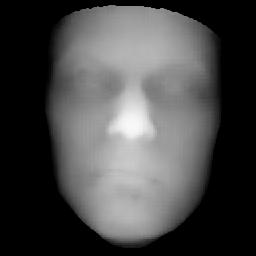}
      \includegraphics[width=0.091\textwidth]{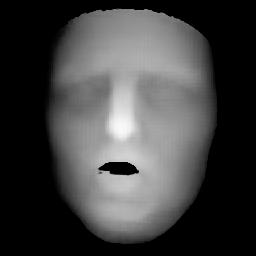}
      \includegraphics[width=0.091\textwidth]{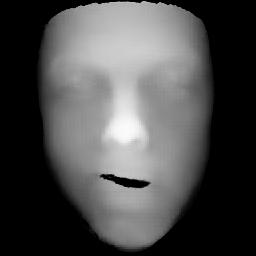}

      \includegraphics[width=0.091\textwidth]{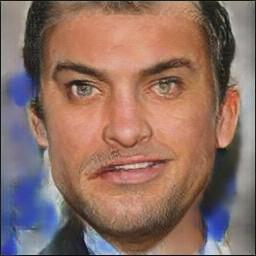}
      \includegraphics[width=0.091\textwidth]{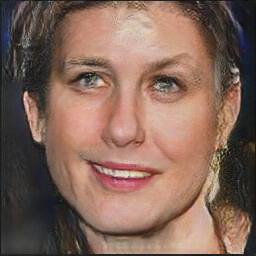}
      \includegraphics[width=0.091\textwidth]{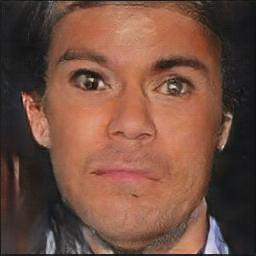}
      \includegraphics[width=0.091\textwidth]{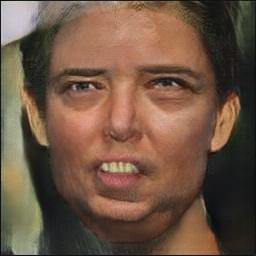}
      \includegraphics[width=0.091\textwidth]{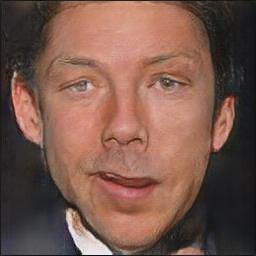}

    \caption{Translation results.
        From top to bottom: synthetic input images, the correspondence and the depth maps recovered by the network, and the transformed result.
         }
    \label{fig:adapt_results}
        \vspace{-0.4cm}
\end{figure}

\section{Limitations}
One of the core ideas of this work was a model-free approach, where the solution space is not restricted by a low dimensional subspace. Instead, the Image-to-Image network represents the solution in the extremely high-dimensional image domain.
This structure is learned from synthetic examples, and shown to successfully generalize to ``in-the-wild'' images. Still, facial images that significantly deviate from our training domain are challenging, resulting in missing areas and errors inside the representation maps.
More specifically, our network has difficulty handling extreme occlusions such as sunglasses, hands or beards, as these were not seen in the training data. Similarly to other methods, reconstructions under strong rotations are also not well handled.
 Reconstructions under such scenarios are shown in the supplementary material.
 Another limiting factor of our pipeline is speed. While the suggested network by itself can be applied efficiently, our template registration step is currently not optimized for speed and can take a few minutes to converge.
\section{Conclusion}
We presented an unrestricted approach for recovering the geometric structure of a face from a single image.
Our algorithm employs an Image-to-Image network which maps the input image to a pixel-based geometric representation, followed by geometric deformation and refinement steps.
The network is trained only by synthetic facial images, yet, is capable of reconstructing real faces.
Using the network as a loss function, we propose a framework for translating synthetic facial images into realistic ones while preserving the geometric structure.

\vspace{-0.3cm}
\subsubsection*{Acknowledgments}
\vspace{-0.1cm}
We would like to thank Roy Or-El for the helpful discussions and comments.
{\small
\bibliographystyle{ieee}
\bibliography{egbib}
}
\fi

\ifshowsupp
\ifshowmain
\newpage
\title{Supplementary Material}
\else
\title{Supplementary Material:\\ Unrestricted Facial Geometry Reconstruction Using Image-to-Image Translation}
\fi

\onecolumn
\appendix
\date{}
\author{}
\makeatletter
\renewcommand{\@maketitle}{\@oldmaketitle}
\maketitle
\section{Additional Network Details}
Here, we summarize additional considerations concerning the network and its training procedure.
\begin{itemize}
\item  The proposed architecture is based on the one introduced in~\cite{isola2016image}.
For allowing further refinement of the results, three additional convolution layers with a kernel of size $1\times 1$ were concatenated at the end.
Following the notations of~\cite{isola2016image}, the encoder architecture is given as $$C64-C128-C256-C512-C512-C512-C512-C512,$$ while the decoder is given by
 $$CD512-CD512-CD512-C512-C512-C256-C128-C64-C^*64-C^*32-C^*4,$$ where $C^*$ represents a $1\times 1$ convolution with stride $1$.
\item The resolution of the input and output training images was $512\times 512$ pixels.
While this is a relatively large input size for training, the Image-to-Image architecture was able to process it successfully, and provided accurate results.
Although, one could train a network on smaller resolutions and then evaluate it on larger images, as shown in~\cite{isola2016image}, we found that our network did not successfully scale up for unseen resolutions.
\item While a single network was successfully trained to retrieve both depth and correspondence representations, our experiments show that training separated networks to recover the representations is preferable.
Note that the architectures of both networks were identical.
This can be justified by the observation that during training, a network allocates its resources for a specific translation task and the representation maps we used have different characteristics.
\item A necessary parameter for the registration step is the scale of the face with respect to the image dimensions.
While this can be estimated based on global features, such as the distance between the eyes, we opted to retrieve it directly by training the network to predict the $x$ and $y$ coordinates of each pixel in the image alongside the $z$ coordinate.
\end{itemize}
\section{Additional Registration and Refinement Details}
Next, we provide a detailed version of the iterative deformation-based registration phase, including implementation details of the fine detail reconstruction.
\subsection{Non-Rigid Registration}
First, we turn the $x$,$y$ and $z$ maps from the network into a mesh, by connecting four neighboring pixels, for which the coordinates are known, with a couple of triangles.
This step yields a target mesh that might have holes but has dense map to our template model.
Based on the correspondence given by the network, we compute the affine transformation from a template face to the mesh.
This operation is done by minimizing the squared Euclidean distances between corresponding vertex pairs.
To handle outliers, a RANSAC approach is used \cite{fischler1981random} with $1,000$ iterations and a threshold of $3$ millimeters for detecting inliers.
Next, similar to \cite{li2010animation}, an iterative non-rigid registration process deforms the transformed  template, aligning it with the mesh.
Note, that throughout the registration, only the template is warped, while the target mesh remains fixed.
Each iteration involves the following four steps.
\begin{enumerate}
\item Each vertex in the template mesh, $v_i \in \mathcal{V}$, is associated with a vertex,
 $c_i$, on the target mesh, by evaluating the nearest neighbor in the embedding space.
This step is different from the method described in \cite{li2010animation}, which computes the nearest neighbor in the Euclidean space.
As a result, the proposed step allows registering a single template face to different facial identities with arbitrary expressions.

\item Pairs, $(v_i,c_i)$, which are physically distant by more than $1$ millimeter and those with normal direction disagreement of more than $5$ degrees are detected and ignored in the next step.
\item The template mesh is deformed by minimizing the following energy
\begin{eqnarray}
E(V,C) &=& \alpha_{p2point} \sum_{(v_i,c_i)\in\mathcal{J}} \|v_i - c_i\|^2_2  \cr
 & & + \alpha_{p2plane}\sum_{(v_i,c_i)\in\mathcal{J}}\left| \vec{n}(c_i)(v_i-c_i) \right|^2  \cr
 & & + \alpha_{memb} \sum_{i \in \mathcal{V}}\sum_{v_j\in
             \mathcal{N}(v_i)} w_{i,j} \|v_i-v_j\|^2_2,\cr && \,\,\,
\label{eqn:def_energy}
\end{eqnarray}
where, $w_{i,j}$ is the weight corresponding to the biharmonic Laplacian operator (see \cite{helenbrook2003mesh,botsch08variational}), $\vec{n}(c_i)$ is the normal of the corresponding vertex at the target mesh $c_i$, $\mathcal{J}$ is the set of the remaining associated vertex pairs $(v_i,c_i)$, and $\mathcal{N}(v_i)$ is the set 1-ring neighboring vertices about the vertex $v_i$.
Notice that the first term above is the sum of squared Euclidean distances between matches and its weight $\alpha_{p2point}$ is set to $0.1$.
The second term is the distance from the point $v_i$ to the tangent plane at the corresponding point on the target mesh, and its weight $\alpha_{p2plane}$ is set to $1$.
The third term quantifies the stiffness of the mesh and its weight $\alpha_{memb}$ is initialized to $10^8$.
In practice, the energy term given in~\autoref{eqn:def_energy} is minimized iteratively by an inner loop which contains a linear system of equations.
We run this loop until the norm of the difference between the vertex positions of the current iteration and the previous one is below $0.01$.
\item If the motion of the template mesh between the current outer iteration and the previous one is below $0.1$, we divide the weight $\alpha_{memb}$ by two.
This relaxes the stiffness term and allows a greater deformation in the next outer iteration.
In addition, we evaluate the difference between the number of remaining pairwise matches in the current iteration versus the previous one.
If the difference is below 500, we modify the vertex association step to estimate the physical nearest neighbor vertex, instead of the the nearest neighbor in the space of the embedding given by the network.
\end{enumerate}
This iterative process terminates when the stiffness weight $\alpha_{memb}$ is below $10^6$.
The resulting output of this phase is a deformed template with fixed triangulation, which contains the overall facial structure recovered by the network, yet, is smoother and complete.
\subsection{Fine Detail Reconstruction}
Although the network already recovers fine geometric details, such as wrinkles and moles, across parts of the face, a geometric approach can reconstruct details at a finer level, on the entire face, independently of the resolution.
Here, we propose an approach motivated by the passive-stereo facial reconstruction method suggested in \cite{beeler2010singleshot}.
The underlying assumption here is that subtle geometric structures can be explained by local variations in the image domain.
For some skin tissues, such as nevi, this assumption is inaccurate as the intensity variation results from the albedo.
In such cases, the geometric structure would be wrongly modified.
Still, for most parts of the face, the reconstructed details are consistent with the actual variations in depth.

The method begins from an interpolated version of the deformed template, provided by a surface subdivision technique.
Each vertex $v\in\mathcal{V}_{\mathcal{D}}$ is painted with the intensity value of the nearest pixel in the image plane.
Since we are interested in recovering small details, only the high spatial frequencies, $\mu(v)$, of the texture, $\tau(v)$, are taken into consideration in this phase.
For computing this frequency band, we subtract the synthesized low frequencies from the original intensity values.
This low-pass filtered part can be computed by convolving the texture with a spatially varying Gaussian kernel in the image domain, as originally proposed.
In contrast, since this convolution is equivalent to computing the heat distribution upon the shape after time $dt$, where the initial heat profile is the original texture, we propose to compute $\mu(v)$ as
\begin{align}
 \mu(v) = \tau(v) - (I - dt\cdot \Delta_g)^{-1}\tau(v),
\end{align}
where $I$ is the identity matrix, $\Delta_g$ is the cotangent weight discrete Laplacian operator for triangulated meshes \cite{meyer2002discrete}, and $dt = 0.2$ is a scalar proportional to the cut-off frequency of the filter.

Next, we displace each vertex along its normal direction such that $v' = v + \delta(v) \vec{n}(v)$.
The step size of the displacement, $\delta(v)$, is a combination of a data-driven term, $\delta_{\mu}(v)$, and a regularization one, $\delta_{s}(v)$.
The data-driven term is guided by the high-pass filtered part of the texture, $\mu(v)$.
In practice, we require the local differences in the geometry to be proportional to the local variation in the high frequency band of the texture. That is for each vertex $v$, with a normal $\vec{n}(v)$, and a neighboring vertex $v_i$, the data-driven term is given by
\begin{align}
(\mu(v) - \mu(v_i)) =  \langle v +\delta_{\mu}(v) \vec{n}(v) -v_i,\vec{n}(v) \rangle.
\end{align}
Thus, the step size assuming a single neighboring vertex can be calculated by
\begin{align}
\delta_{\mu}(v) = \gamma(\mu(v) - \mu(v_i) ) -\langle v  -v_i,\vec{n}(v) \rangle.
\end{align}
In the presence of any number of neighboring vertices of $v$, we compute the weighted average of its 1-ring neighborhood
\begin{align}
\delta_{\mu}(v) = \frac{\sum_{v_i \in \mathcal{N}(v)} \alpha(v,v_i)
 \gamma \left[ \left(\mu(v)-\mu(v_i)\right) -\langle v  -v_i,\vec{n}(v) \rangle \right] }{\sum_{v_i \in \mathcal{N}(v)} \alpha(v,v_i)},
\end{align}
An alternative term can spatially attenuate the contribution of the data-driven term in curved regions for regularizing the reconstruction by
\begin{align}
\delta_{\mu}(v)=\frac{\underset{v_{i}\in\mathcal{N}(v)}{\sum}\alpha_{(v,v_{i})}\left(\mu(v)-\mu(v_{i})\right)\left(1-\frac{\left|\langle v-v_{i},\vec{n}(v)\rangle\right|}{\|v-v_{i}\|}\right)}{\underset{v_{i}\in\mathcal{N}(v)}{\sum}\alpha_{(v,v_{i})}},
\end{align}
where $\alpha_{(v,v_i)} = \exp{\left(-\|v-v_i\|\right)}$.
 where $\mathcal{N}(v)$ is the set 1-ring neighboring vertices about the vertex $v$, and $\vec{n}(v)$ is the unit normal at the vertex $v$.

Since we move each vertex along the normal direction, triangles could intersect each other, particularly in regions with high curvature.
To reduce the probability of such collisions,
a regularizing displacement field, $\delta_s(v)$, is added.
This term is proportional to the mean curvature of the original surface, and is equivalent to a single explicit mesh fairing step \cite{desbrun1999meshfairing}.
The final surface modification is given by
\begin{align}
v' = v + (\eta \delta_{\mu}(v) + (1-\eta) \delta_s(v))\cdot \vec{n}(v),
\end{align}
for a constant $\eta = 0.2$.
\newpage
\section{Additional Experimental Results}
We present additional qualitative results of our method. \autoref{fig:net_out_supp} shows the output representations of the proposed network for a variety of different faces, notice the failure cases presented in the last two rows. One can see that the network generalizes well, but is still limited by the synthetic data. Specifically, the network might fail in presence of occlusions, facial hair or extreme poses. This is also visualized in~\autoref{fig:lims_result} where the correspondence error is visualized using the texture mapping.
Additional reconstruction results of our method are presented in \autoref{fig:more_qual}.
For analyzing the distribution of the error along the face, we present an additional comparison in~\autoref{fig:heat_maps}, where the absolute error, given in percents of the ground truth depth, is shown for several facial images.

\begin{figure}[h]
    \centering
\includegraphics[width=0.16\textwidth]{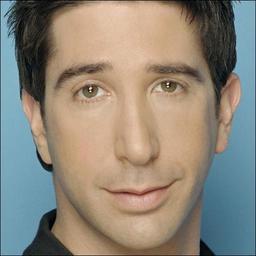}
 \includegraphics[width=0.16\textwidth]{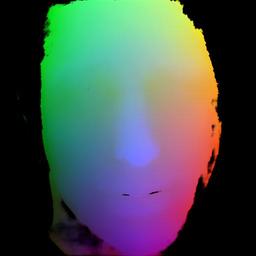}
  \includegraphics[width=0.16\textwidth]{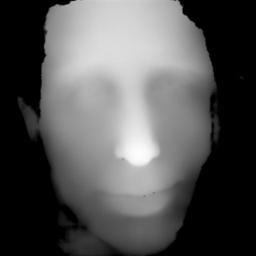}
\includegraphics[width=0.16\textwidth]{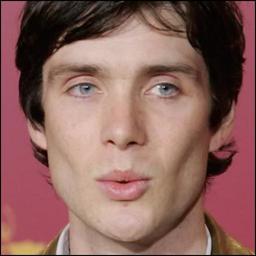}
 \includegraphics[width=0.16\textwidth]{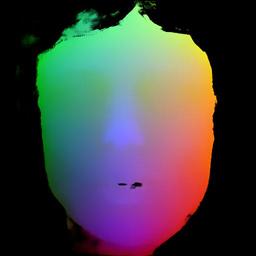}
  \includegraphics[width=0.16\textwidth]{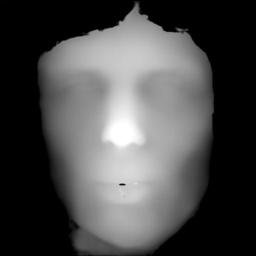}

  \includegraphics[width=0.16\textwidth]{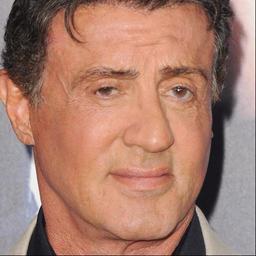}
 \includegraphics[width=0.16\textwidth]{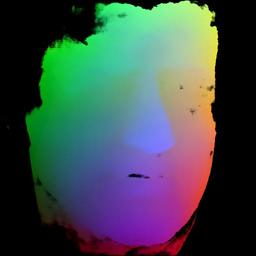}
  \includegraphics[width=0.16\textwidth]{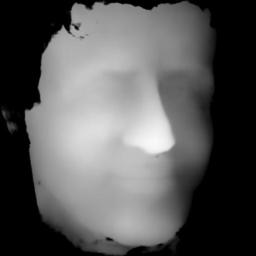}
\includegraphics[width=0.16\textwidth]{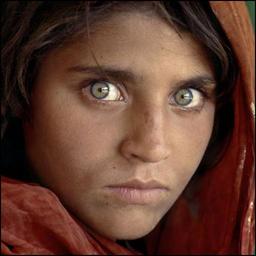}
 \includegraphics[width=0.16\textwidth]{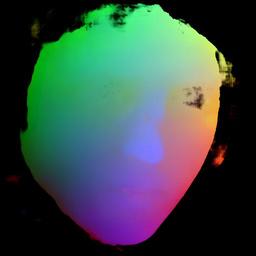}
  \includegraphics[width=0.16\textwidth]{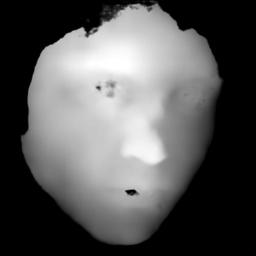}

  \includegraphics[width=0.16\textwidth]{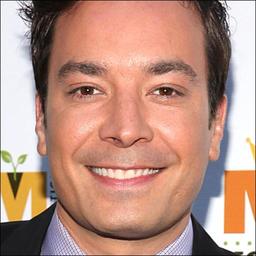}
 \includegraphics[width=0.16\textwidth]{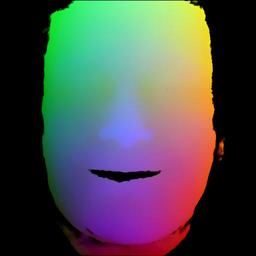}
  \includegraphics[width=0.16\textwidth]{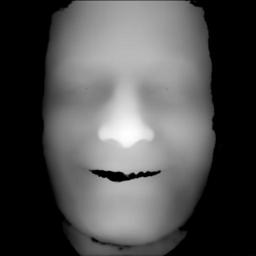}
\includegraphics[width=0.16\textwidth]{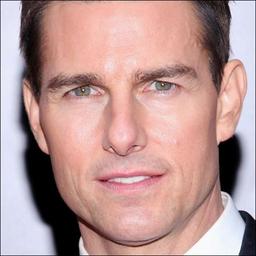}
 \includegraphics[width=0.16\textwidth]{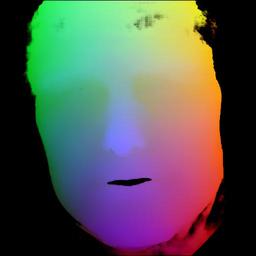}
  \includegraphics[width=0.16\textwidth]{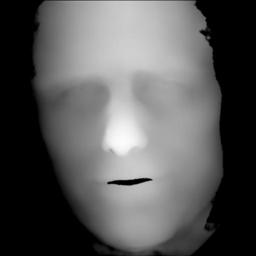}

\includegraphics[width=0.16\textwidth]{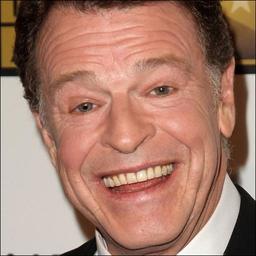}
 \includegraphics[width=0.16\textwidth]{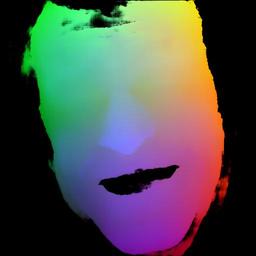}
  \includegraphics[width=0.16\textwidth]{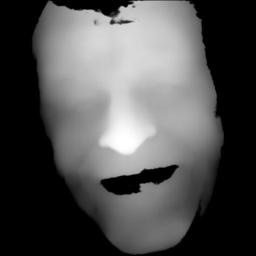}
\includegraphics[width=0.16\textwidth]{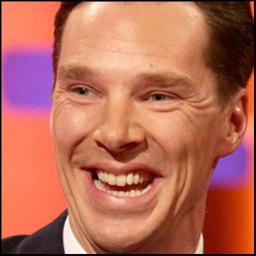}
 \includegraphics[width=0.16\textwidth]{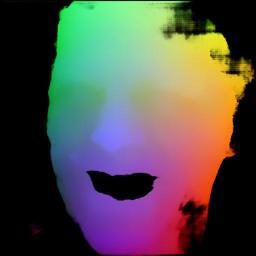}
  \includegraphics[width=0.16\textwidth]{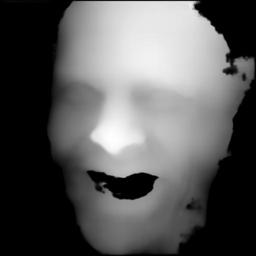}

  \quad

  \includegraphics[width=0.16\textwidth]{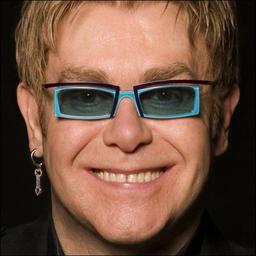}
 \includegraphics[width=0.16\textwidth]{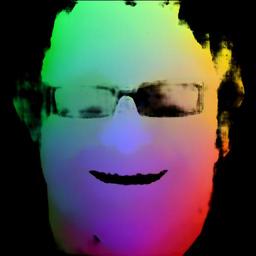}
  \includegraphics[width=0.16\textwidth]{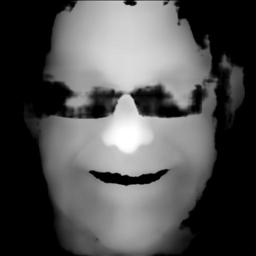}
\includegraphics[width=0.16\textwidth]{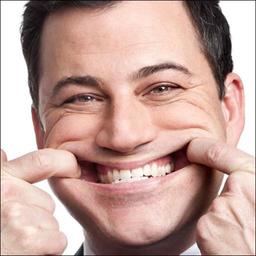}
 \includegraphics[width=0.16\textwidth]{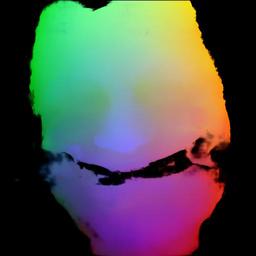}
  \includegraphics[width=0.16\textwidth]{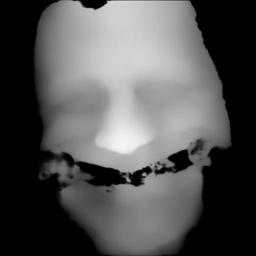}

  \includegraphics[width=0.16\textwidth]{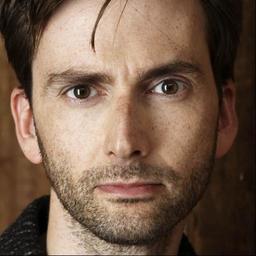}
 \includegraphics[width=0.16\textwidth]{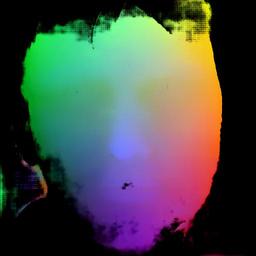}
  \includegraphics[width=0.16\textwidth]{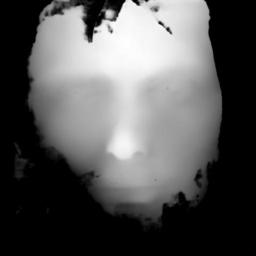}
  \includegraphics[width=0.16\textwidth]{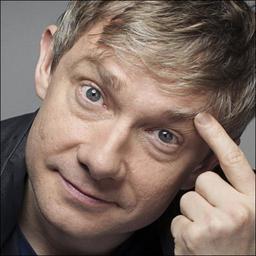}
 \includegraphics[width=0.16\textwidth]{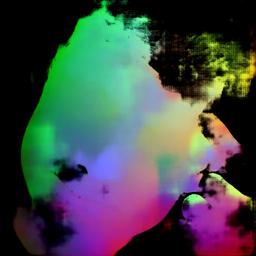}
  \includegraphics[width=0.16\textwidth]{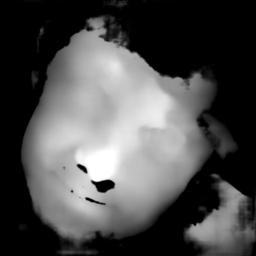}

    \caption{Network Output.}
    \label{fig:net_out_supp}
\end{figure}

\begin{figure}[b]
    \centering

    \includegraphics[width=0.16\textwidth]{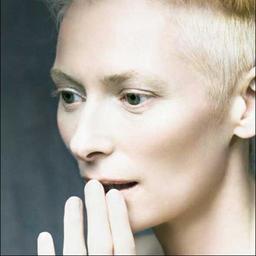}
    \includegraphics[width=0.16\textwidth]{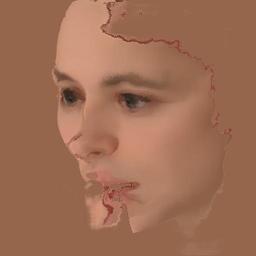}
    \includegraphics[width=0.16\textwidth]{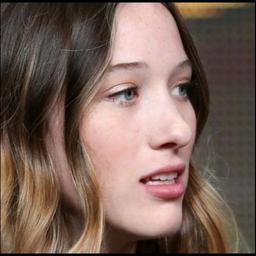}
    \includegraphics[width=0.16\textwidth]{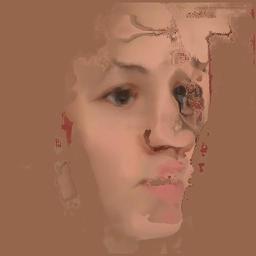}
    \includegraphics[width=0.16\textwidth]{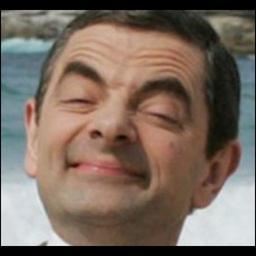}
    \includegraphics[width=0.16\textwidth]{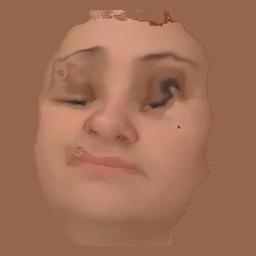}

    \includegraphics[width=0.16\textwidth]{images/all_maps/Elton_john-in}
    \includegraphics[width=0.16\textwidth]{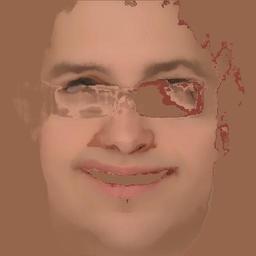}
    \includegraphics[width=0.16\textwidth]{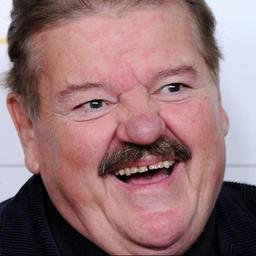}
    \includegraphics[width=0.16\textwidth]{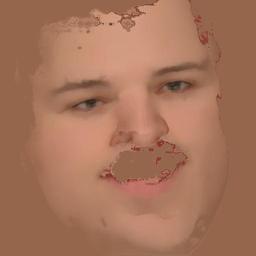}
    \includegraphics[width=0.16\textwidth]{images/all_maps/Jimmy-Kimmel-Header-in}
    \includegraphics[width=0.16\textwidth]{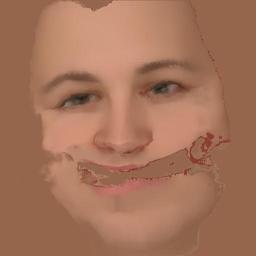}

    \caption{Results under occlusions and rotations.
        Input images are shown next to the matching correspondence result, visualized using the texture mapping to better show the errors.
         }
    \label{fig:lims_result}
\end{figure}

\begin{figure}
	\setlength{\tabcolsep}{12pt}
    \centering
    \begin{tabular}{ccc}
      \includegraphics[height=0.35\textwidth]{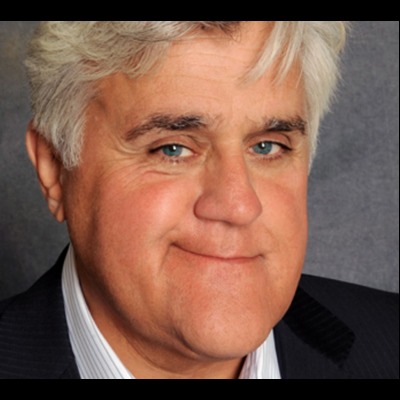}&
      \includegraphics[height=0.35\textwidth]{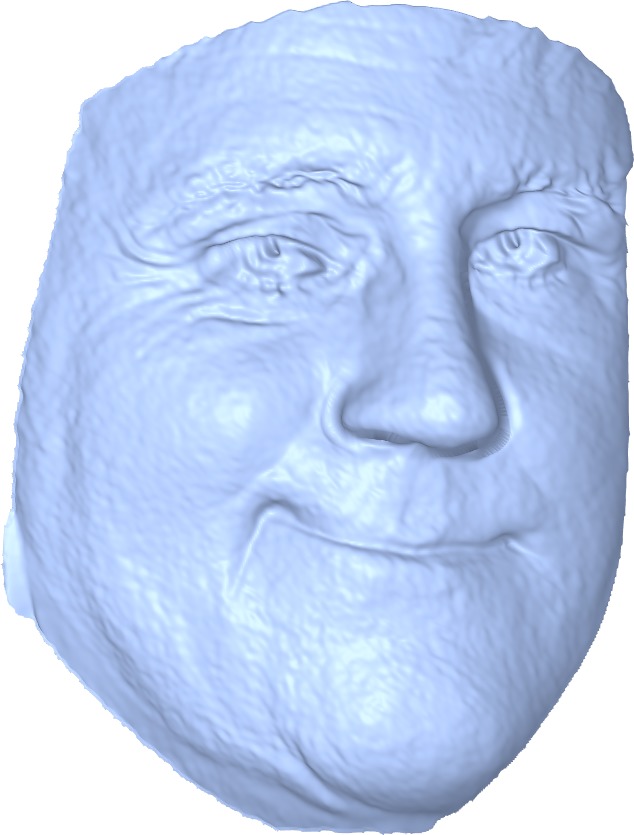}&
      \includegraphics[height=0.35\textwidth]{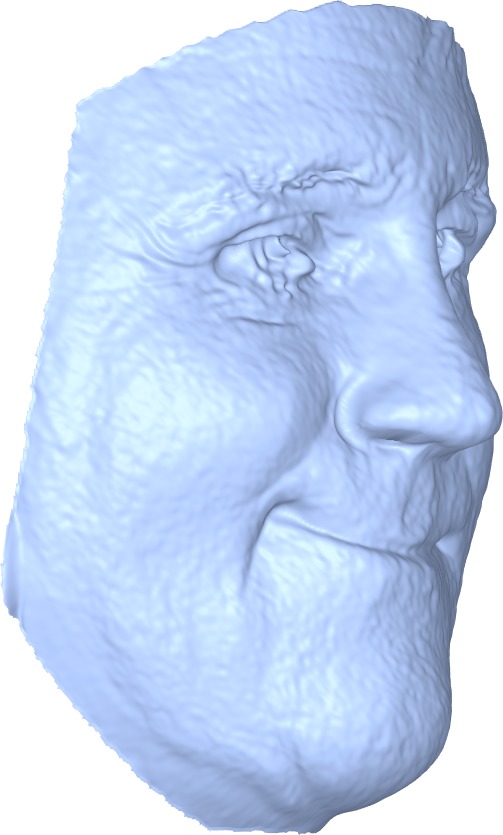}\tabularnewline
      \includegraphics[height=0.35\textwidth]{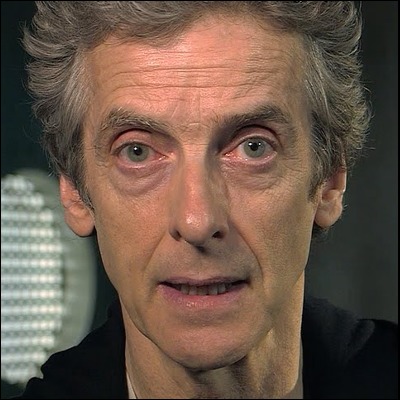}&
      \includegraphics[height=0.35\textwidth]{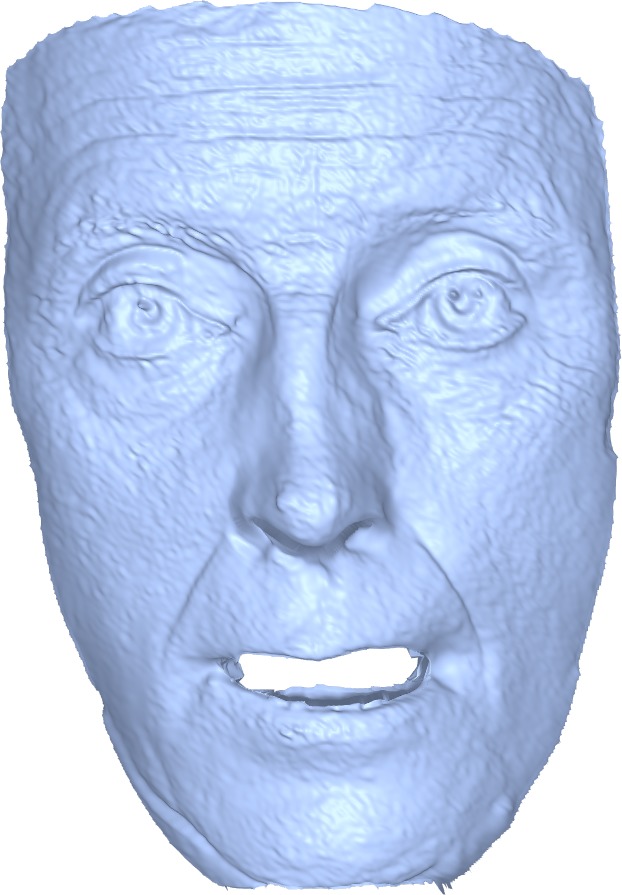}&
      \includegraphics[height=0.35\textwidth]{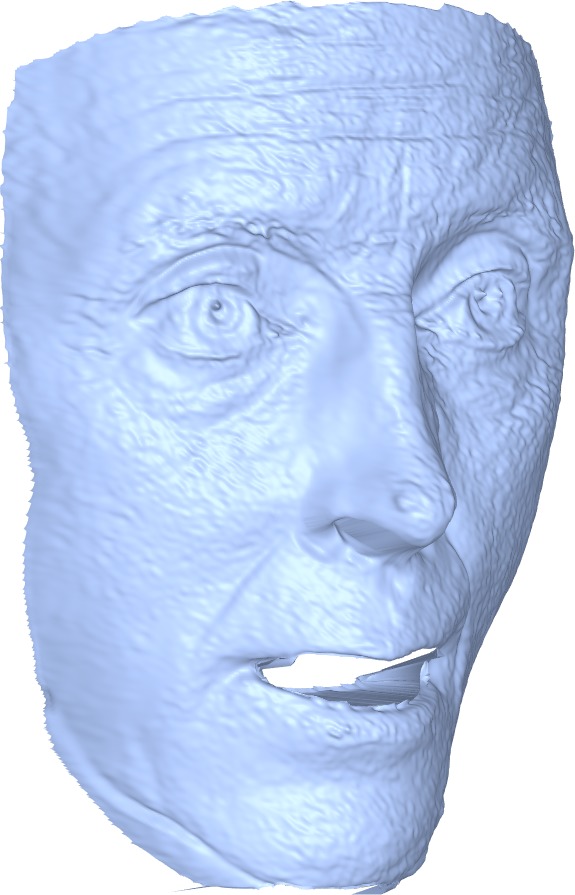}\tabularnewline
      \end{tabular}
    \caption{Additional reconstruction results.}
    \label{fig:more_qual}
\end{figure}

\begin{figure}
    \ContinuedFloat
	\setlength{\tabcolsep}{5pt}
    \centering
    \begin{tabular}{ccc}
      \includegraphics[height=0.33\textwidth]{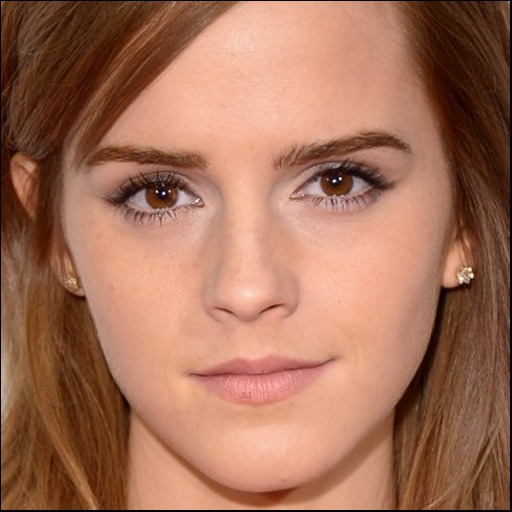}&
      \includegraphics[height=0.33\textwidth]{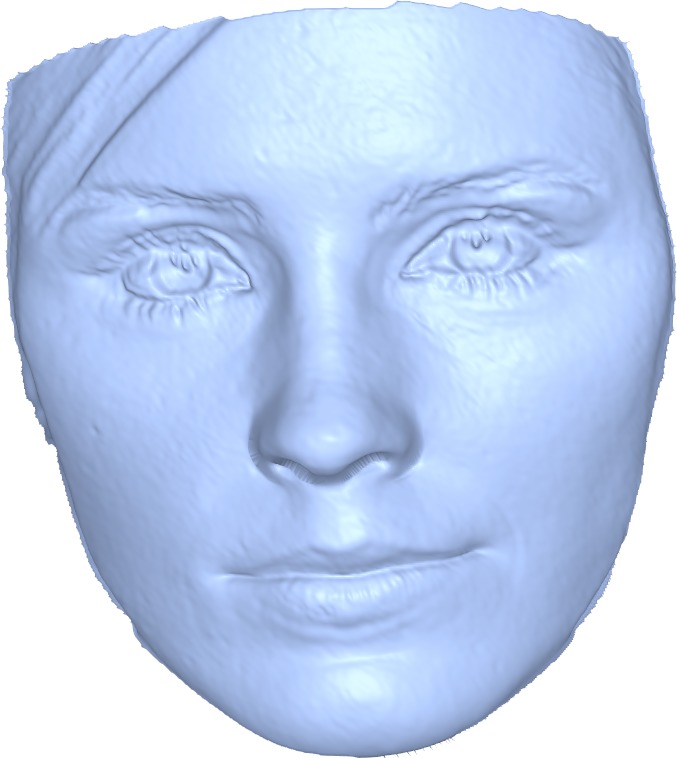}&
      \includegraphics[height=0.33\textwidth]{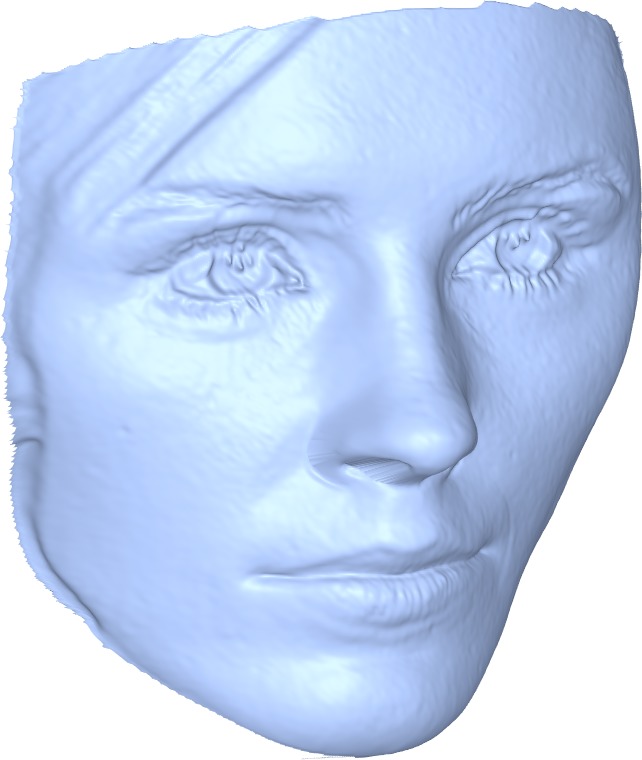}\tabularnewline
      \includegraphics[height=0.33\textwidth]{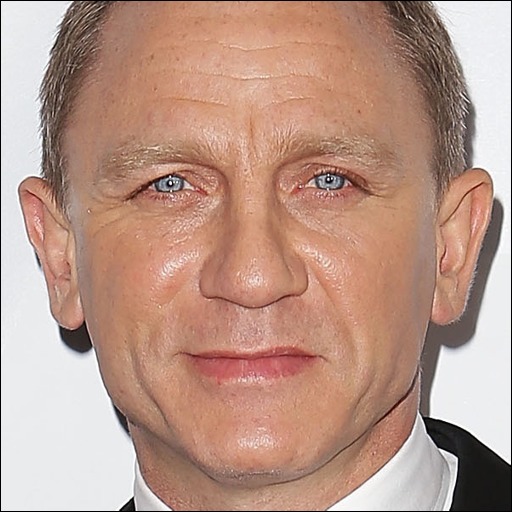}&
      \includegraphics[height=0.33\textwidth]{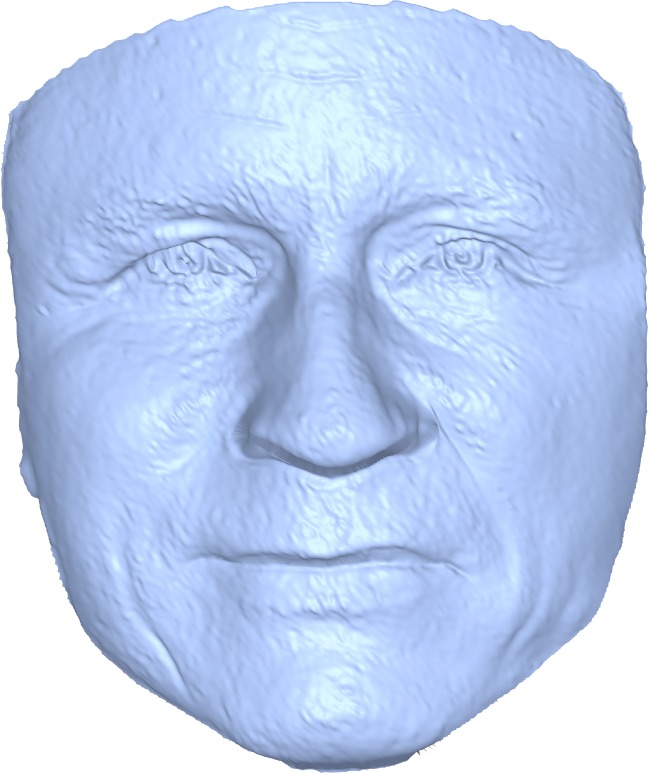}&
      \includegraphics[height=0.33\textwidth]{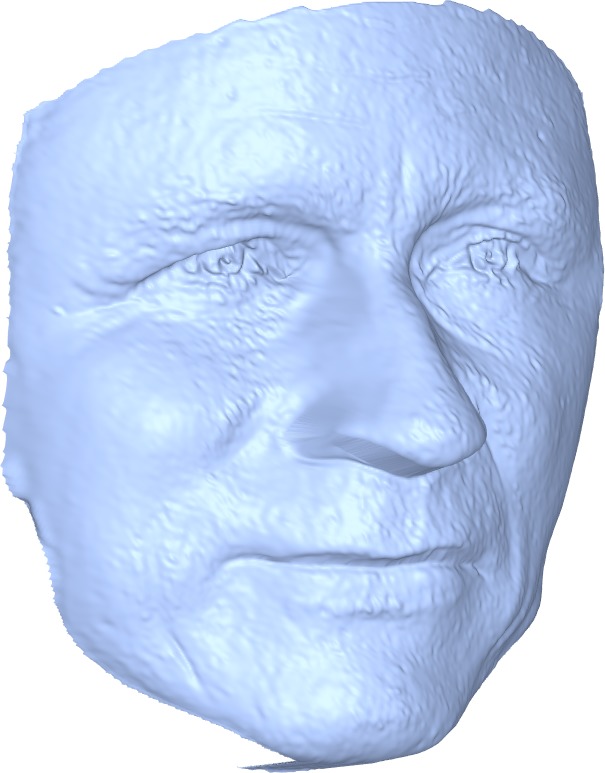}\tabularnewline

 \includegraphics[height=0.33\textwidth]{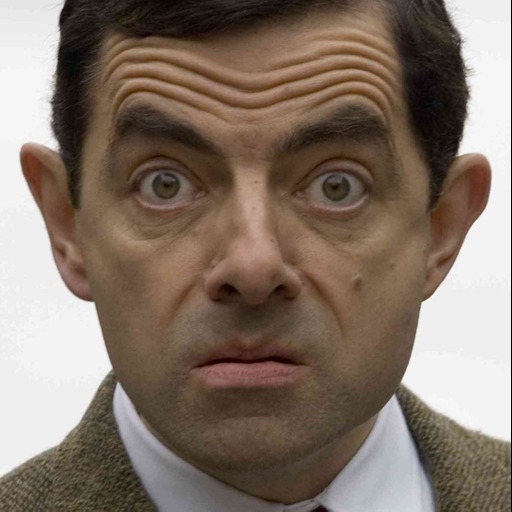}&
      \includegraphics[height=0.33\textwidth]{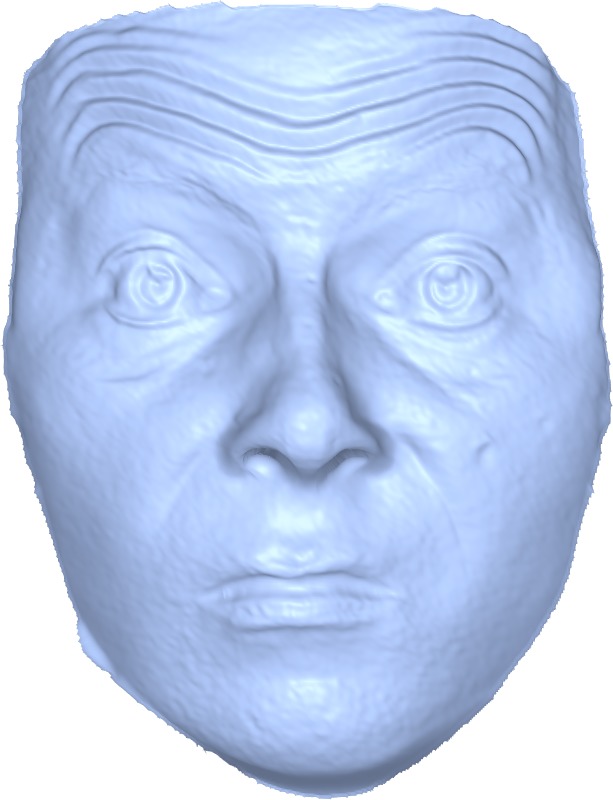}&
      \includegraphics[height=0.33\textwidth]{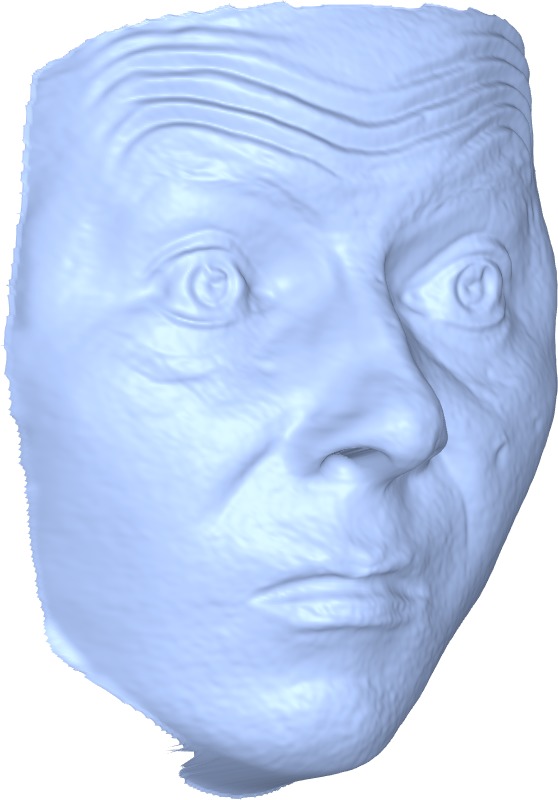}\tabularnewline
      \includegraphics[height=0.33\textwidth]{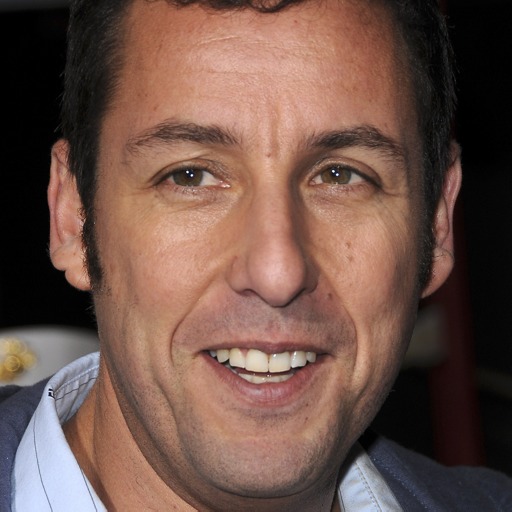}&
      \includegraphics[height=0.33\textwidth]{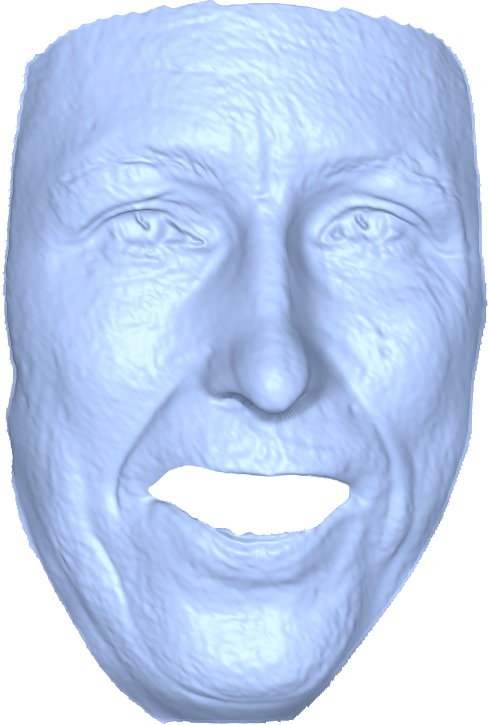}&
      \includegraphics[height=0.33\textwidth]{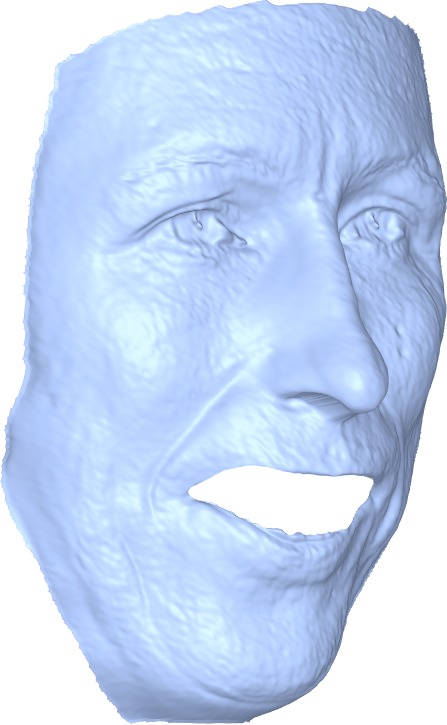}\tabularnewline
      \end{tabular}
    \caption{Additional reconstruction results.}
    \label{fig:more_qual_1}
\end{figure}

\begin{figure}
    \ContinuedFloat
	\setlength{\tabcolsep}{5pt}
    \centering
    \begin{tabular}{ccc}
      \includegraphics[height=0.33\textwidth]{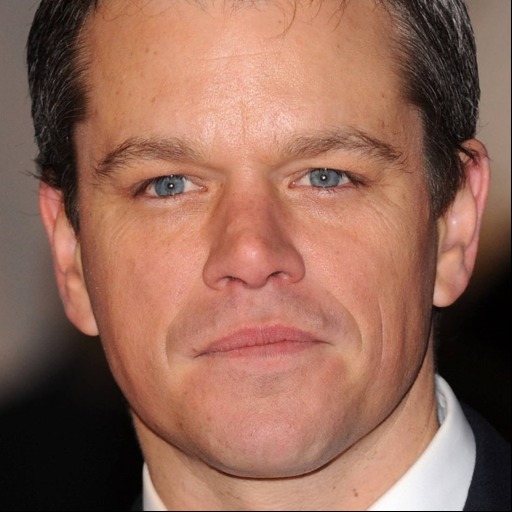}&
      \includegraphics[height=0.33\textwidth]{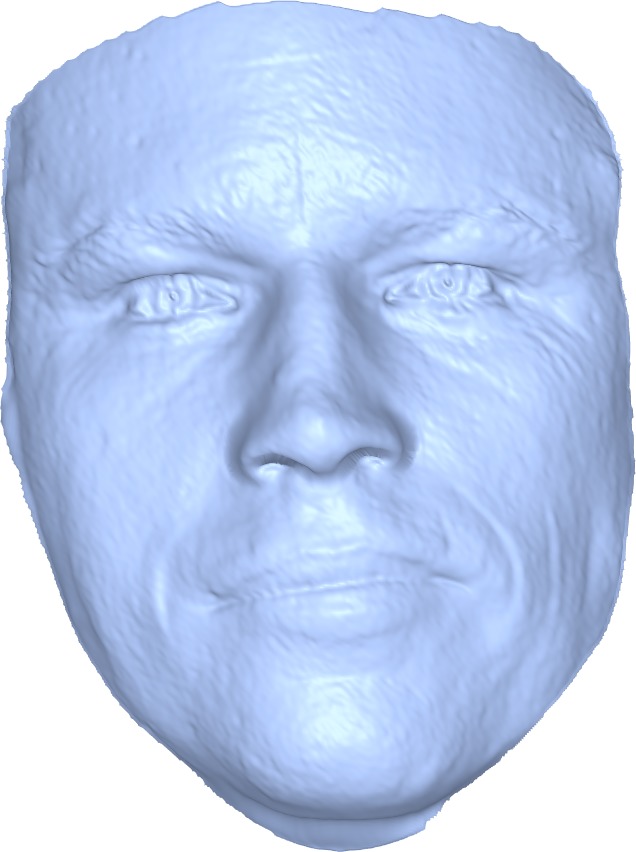}&
      \includegraphics[height=0.33\textwidth]{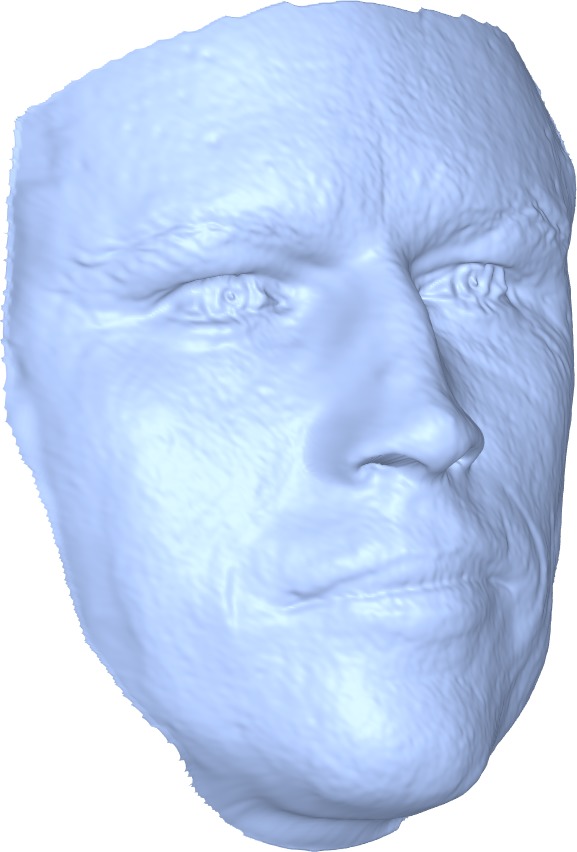}\tabularnewline
      \includegraphics[height=0.33\textwidth]{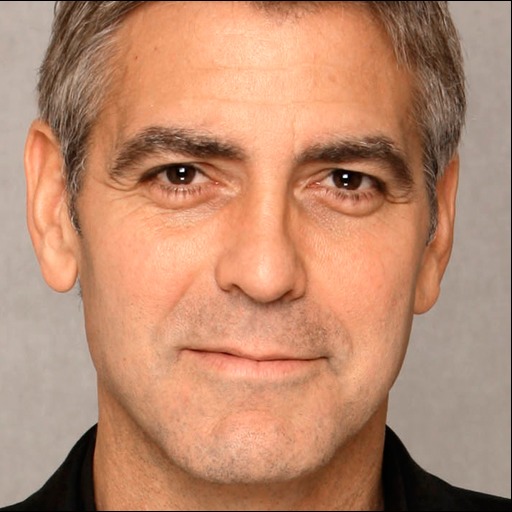}&
      \includegraphics[height=0.33\textwidth]{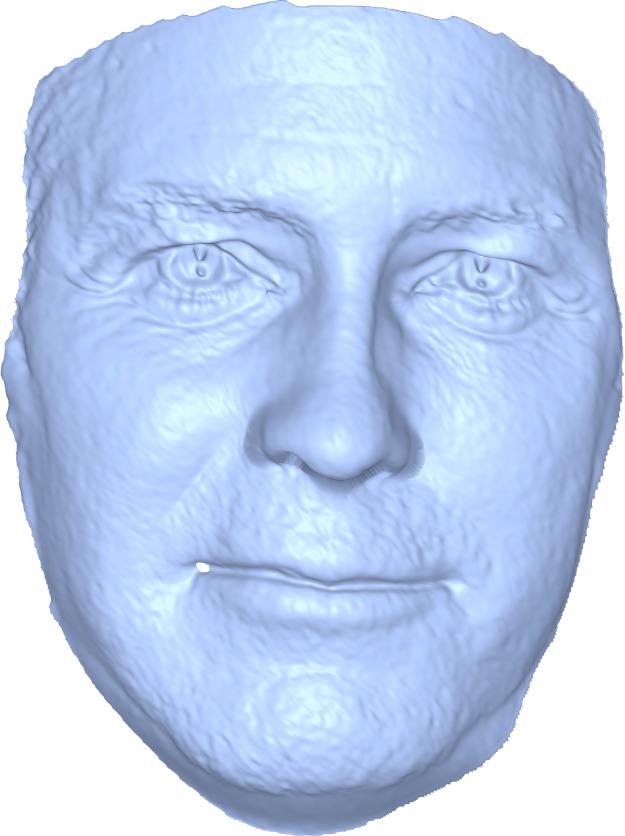}&
      \includegraphics[height=0.33\textwidth]{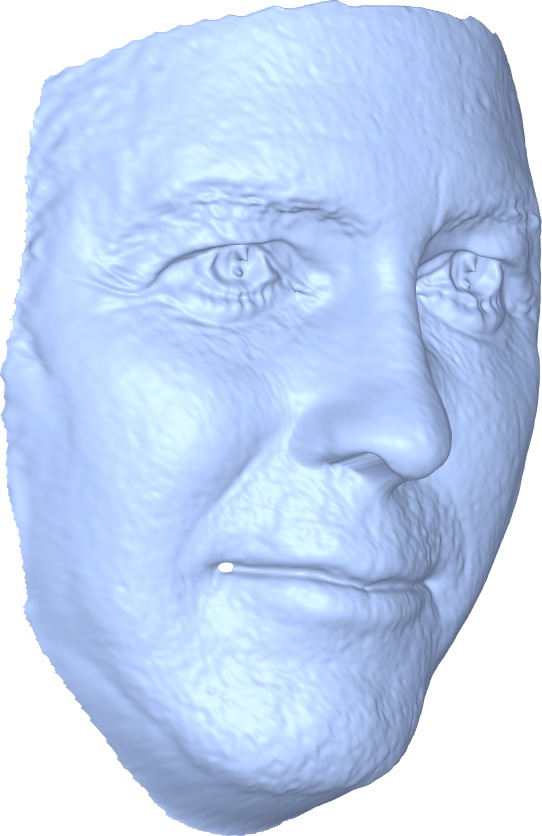}\tabularnewline

 \includegraphics[height=0.33\textwidth]{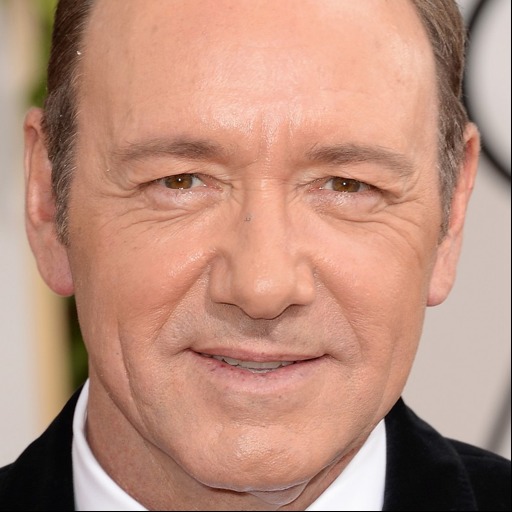}&
      \includegraphics[height=0.33\textwidth]{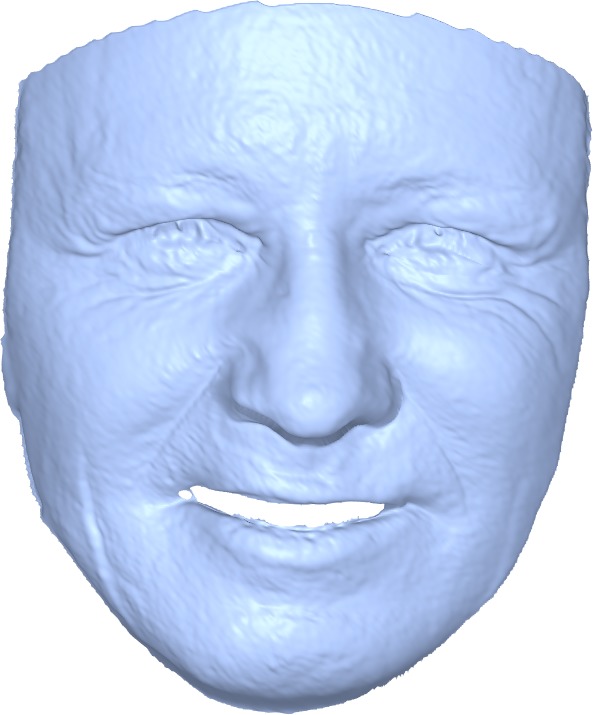}&
      \includegraphics[height=0.33\textwidth]{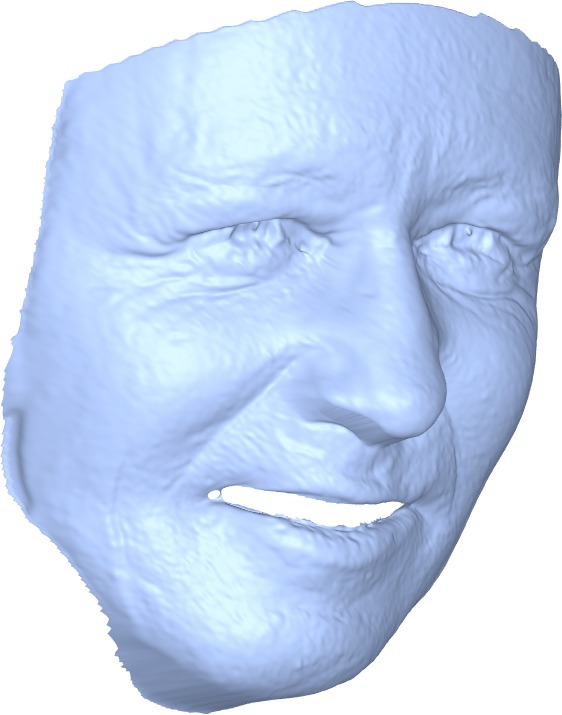}\tabularnewline
      \includegraphics[height=0.33\textwidth]{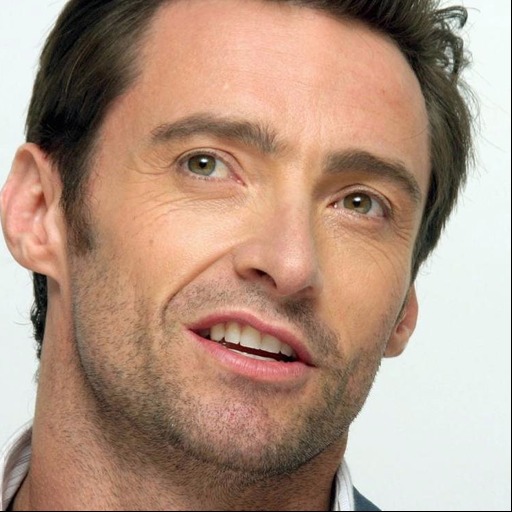}&
      \includegraphics[height=0.33\textwidth]{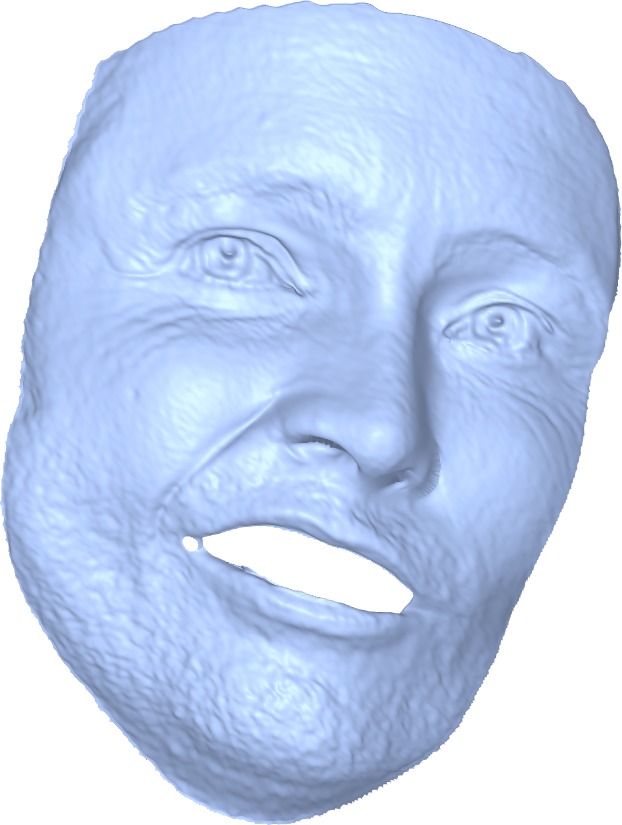}&
      \includegraphics[height=0.33\textwidth]{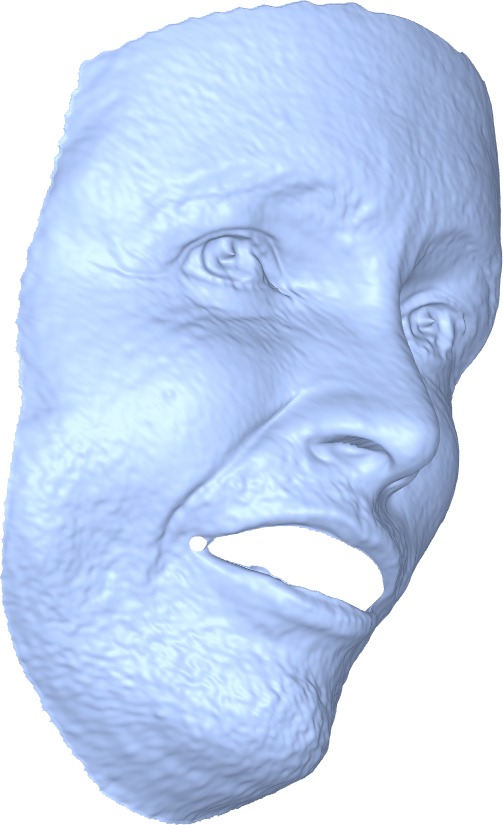}\tabularnewline
      \end{tabular}
    \caption{Additional reconstruction results.}
    \label{fig:more_qual_2}
\end{figure}
\begin{figure}
	\setlength{\tabcolsep}{0.8pt}
    \centering
    \begin{tabular}{cccccc}
      \includegraphics[height=0.28\textwidth,trim={30 5 30 5},clip]{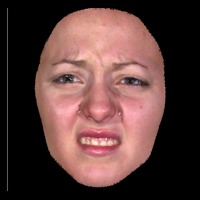}&
    \includegraphics[height=0.28\textwidth,trim={47.5 15 47.5 15},clip]{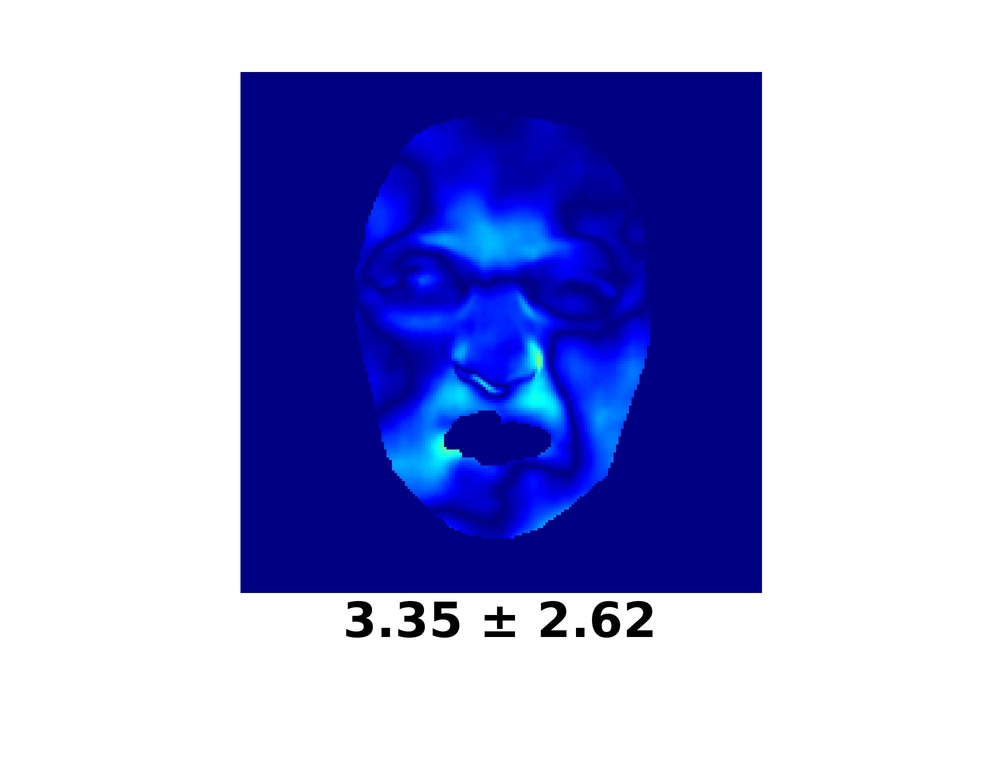}&
      \includegraphics[height=0.28\textwidth,trim={47.5 15 47.5 15},clip]{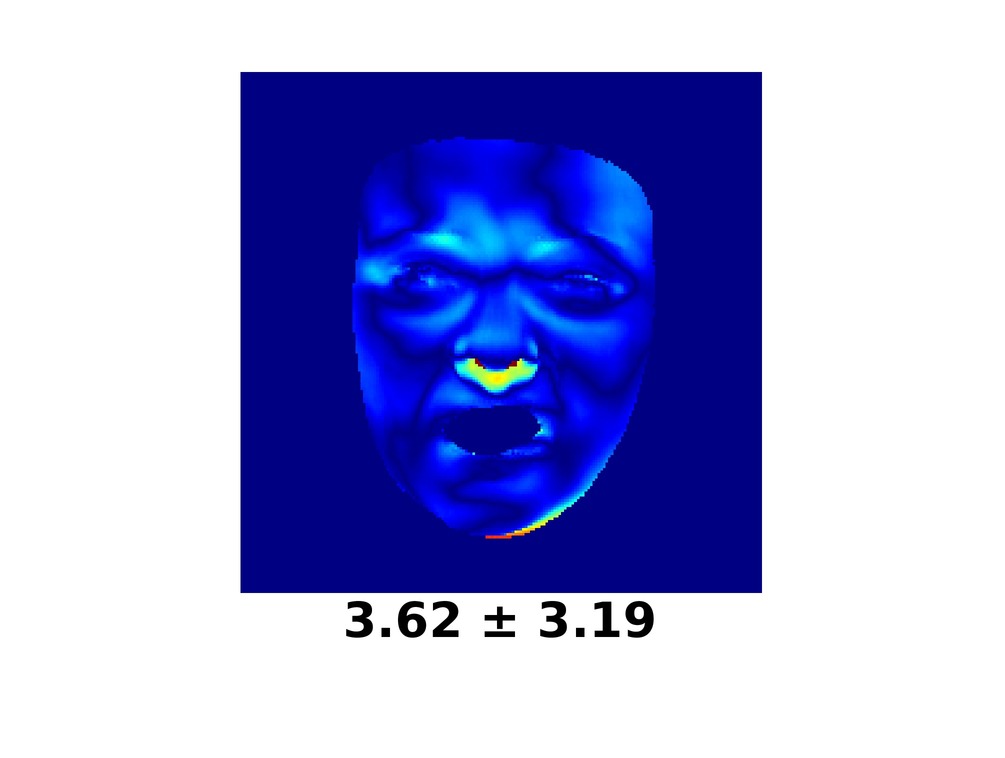}&
      \includegraphics[height=0.28\textwidth,trim={47.5 15 47.5 15},clip]{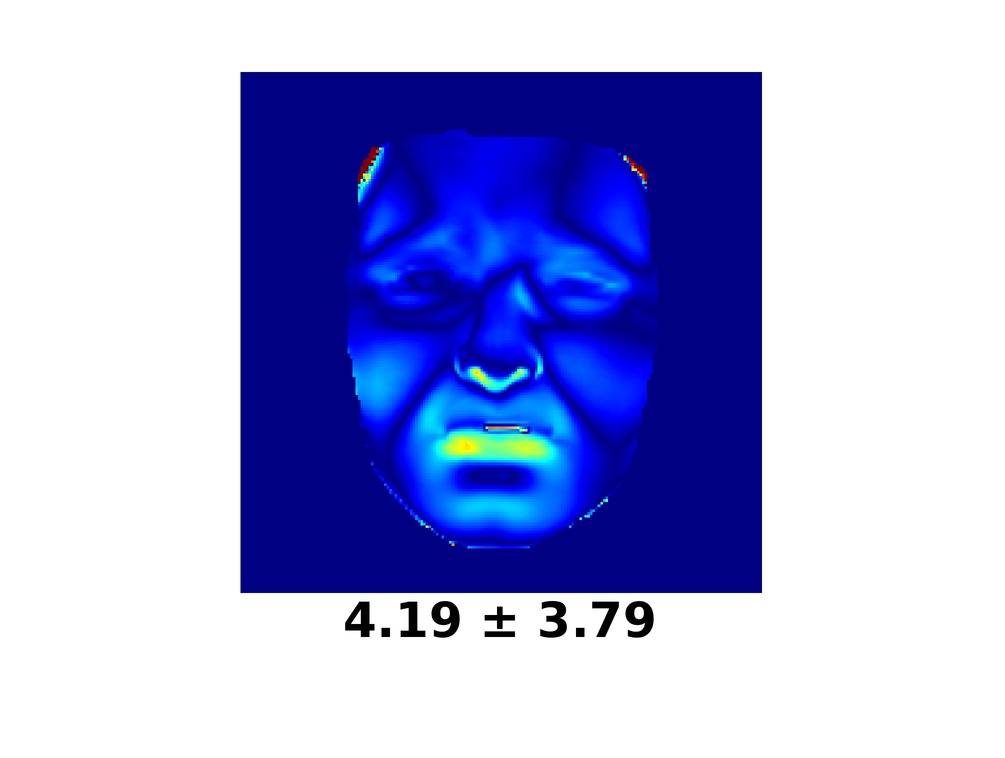} &
      \includegraphics[height=0.28\textwidth,trim={47.5 15 47.5 15},clip]{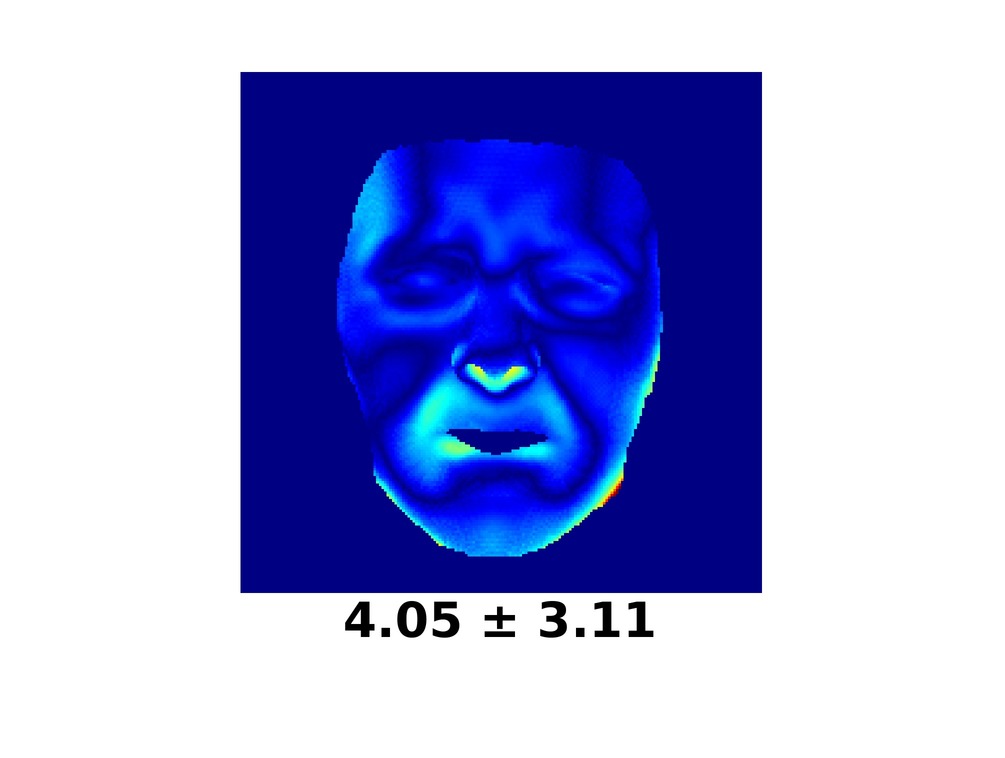} &
      \includegraphics[height=0.28\textwidth]{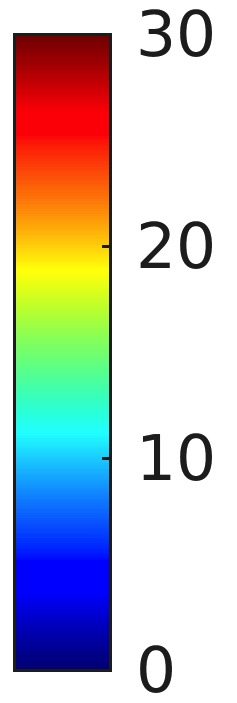}
      \tabularnewline

      \includegraphics[height=0.28\textwidth,trim={30 5 30 5},clip]{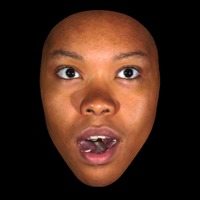}&
    \includegraphics[height=0.28\textwidth,trim={47.5 15 47.5 15},clip]{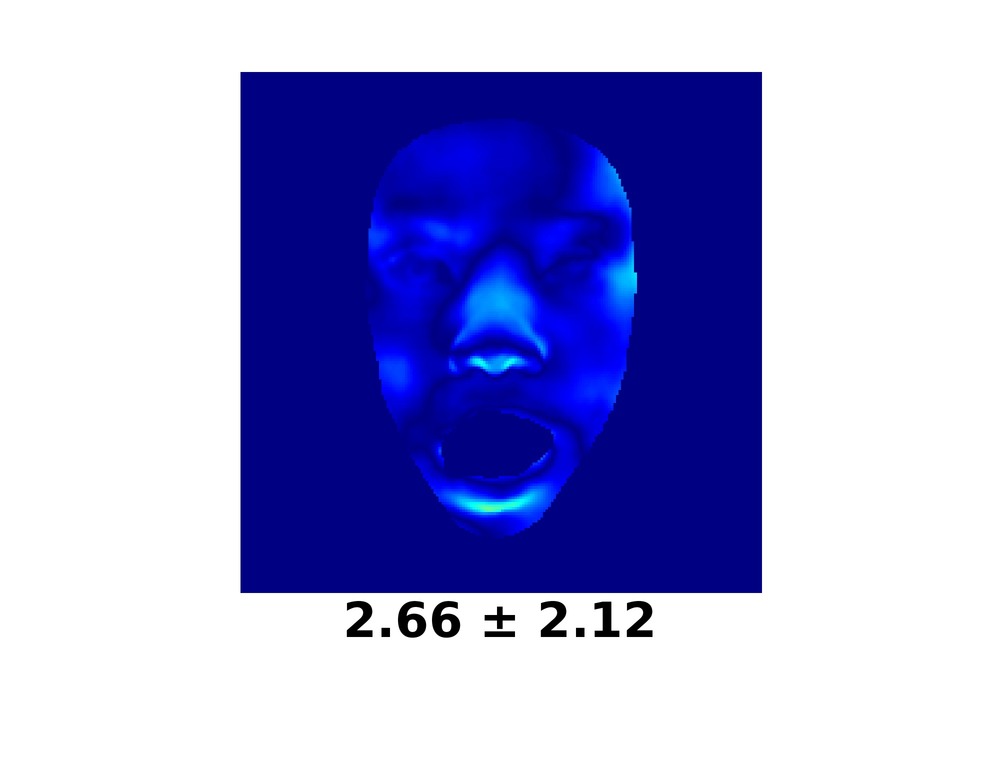}&
      \includegraphics[height=0.28\textwidth,trim={47.5 15 47.5 15},clip]{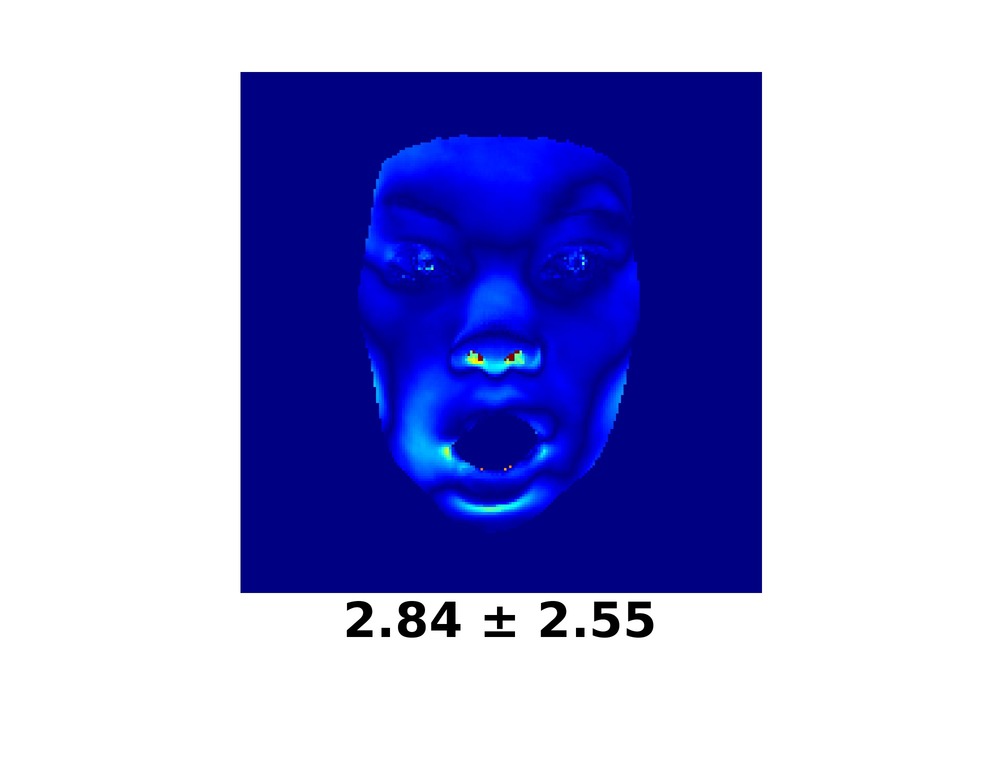}&
      \includegraphics[height=0.28\textwidth,trim={47.5 15 47.5 15},clip]{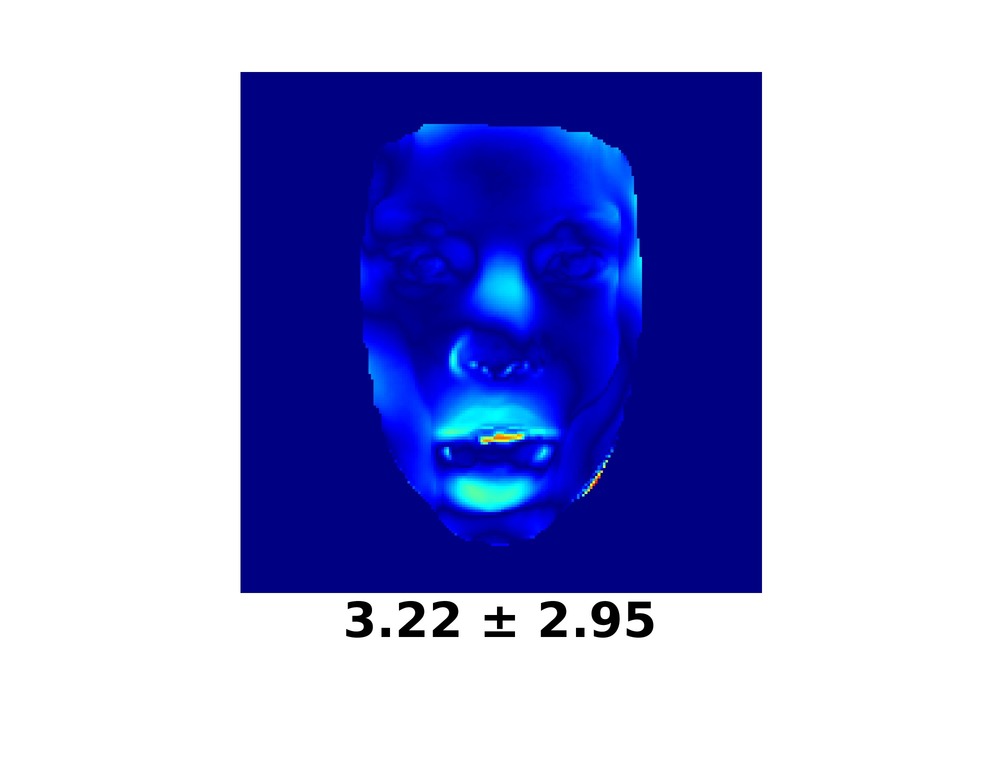} &
      \includegraphics[height=0.28\textwidth,trim={47.5 15 47.5 15},clip]{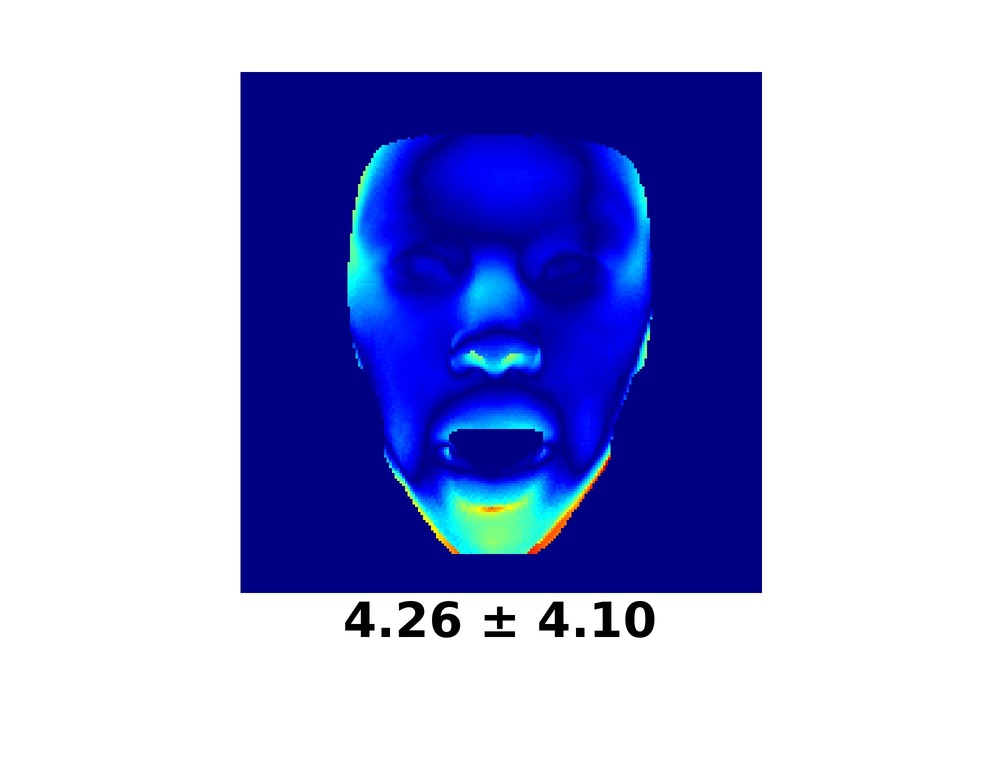} &
      \includegraphics[height=0.28\textwidth]{images/heat_maps/colorbar}
      \tabularnewline

      \includegraphics[height=0.28\textwidth,trim={30 5 30 5},clip]{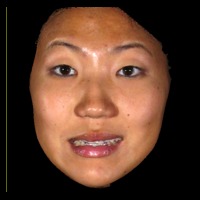}&
    \includegraphics[height=0.28\textwidth,trim={47.5 15 47.5 15},clip]{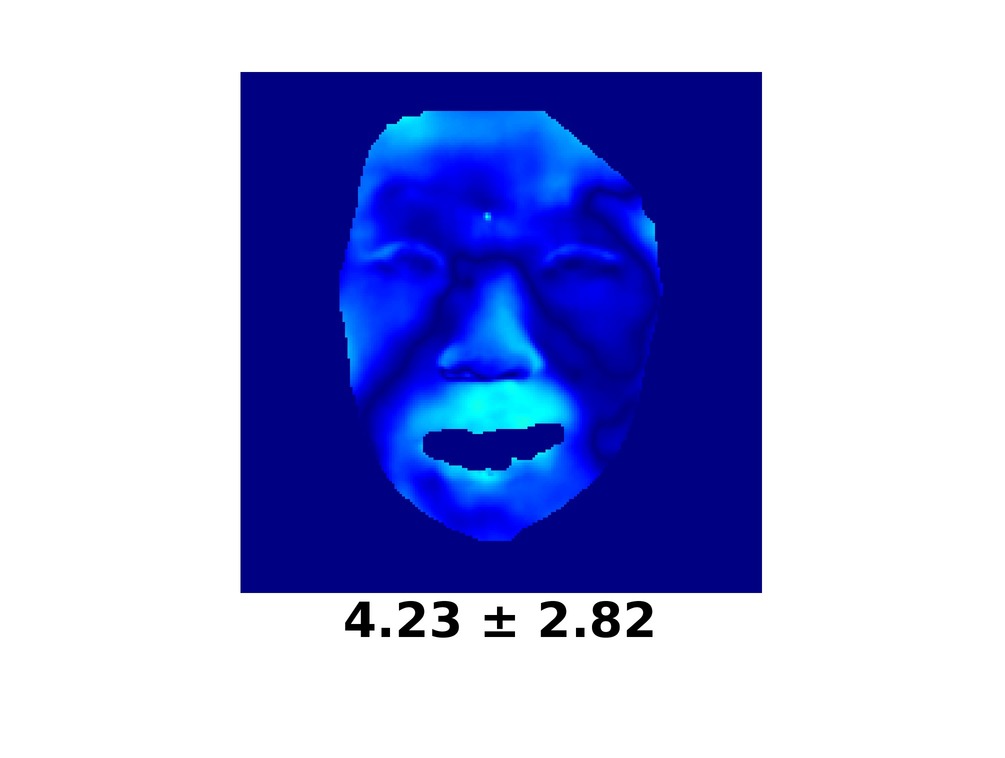}&
      \includegraphics[height=0.28\textwidth,trim={47.5 15 47.5 15},clip]{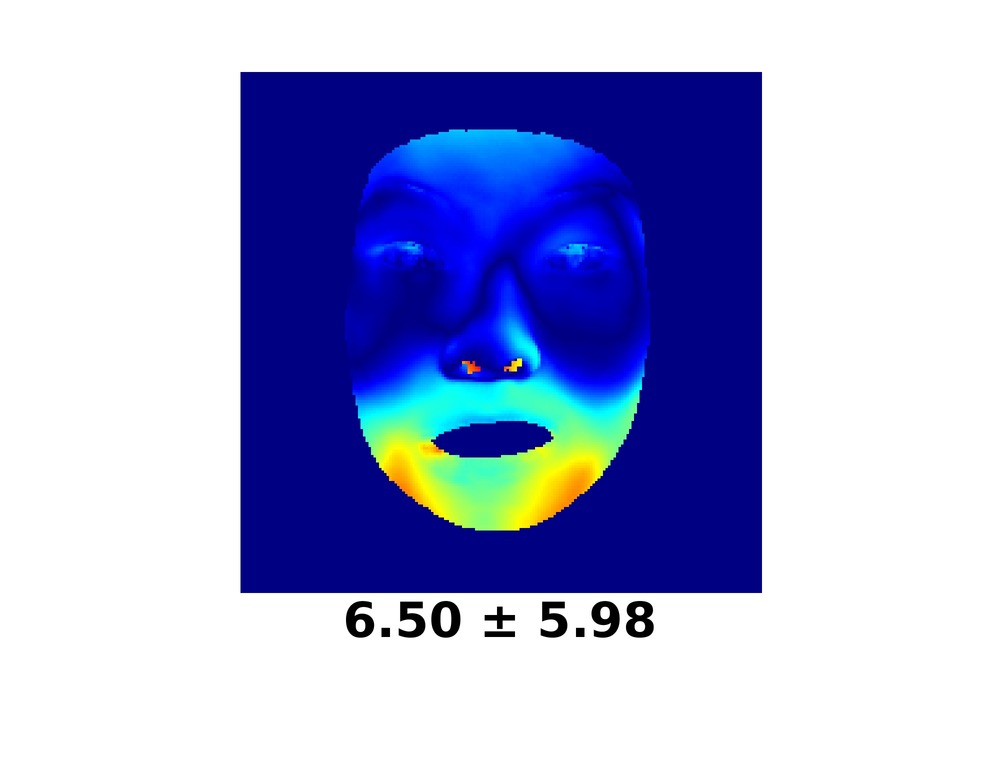}&
      \includegraphics[height=0.28\textwidth,trim={47.5 15 47.5 15},clip]{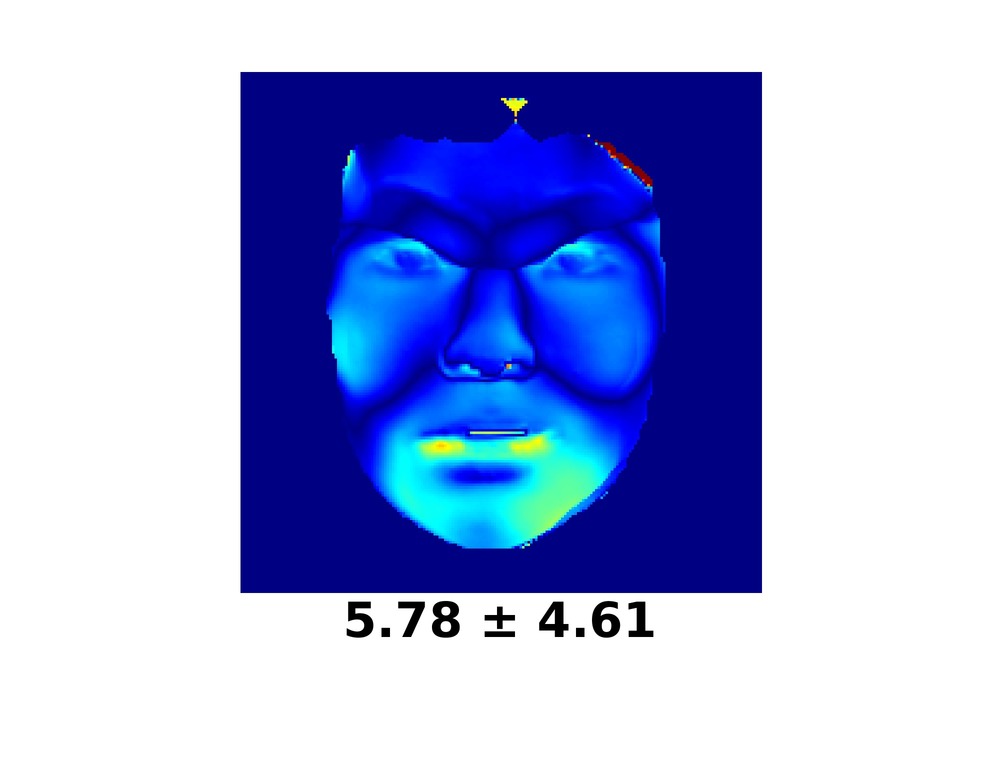} &
      \includegraphics[height=0.28\textwidth,trim={47.5 15 47.5 15},clip]{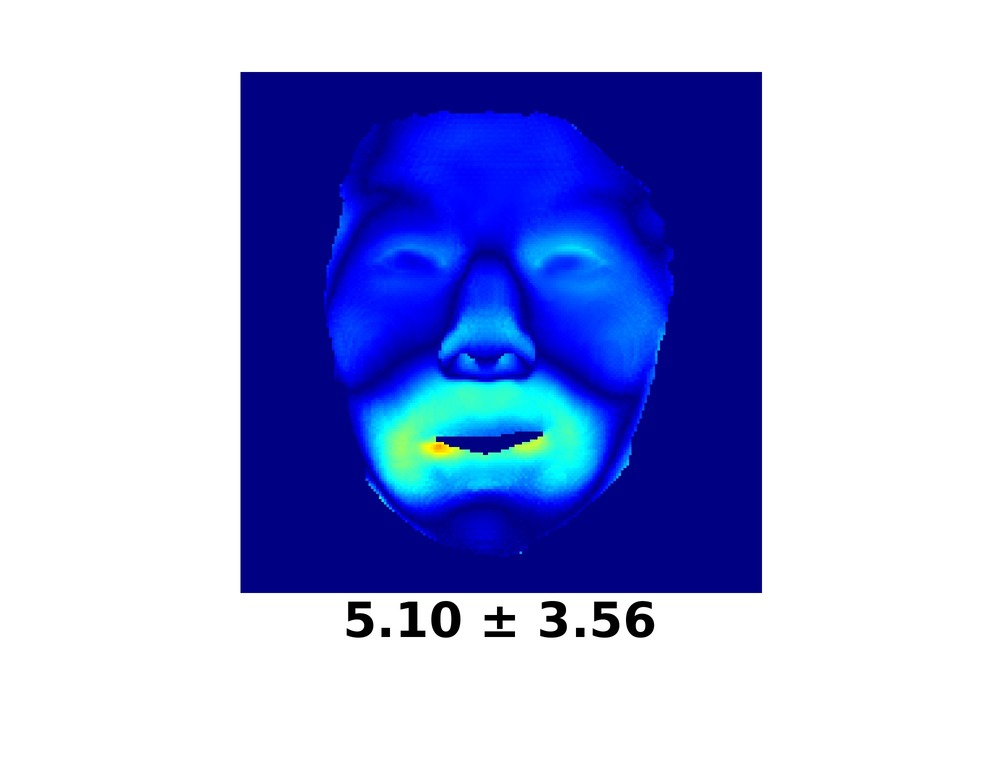} &
      \includegraphics[height=0.28\textwidth]{images/heat_maps/colorbar}
      \tabularnewline

      \includegraphics[height=0.28\textwidth,trim={30 5 30 5},clip]{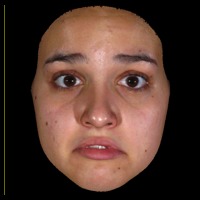}&
    \includegraphics[height=0.28\textwidth,trim={47.5 15 47.5 15},clip]{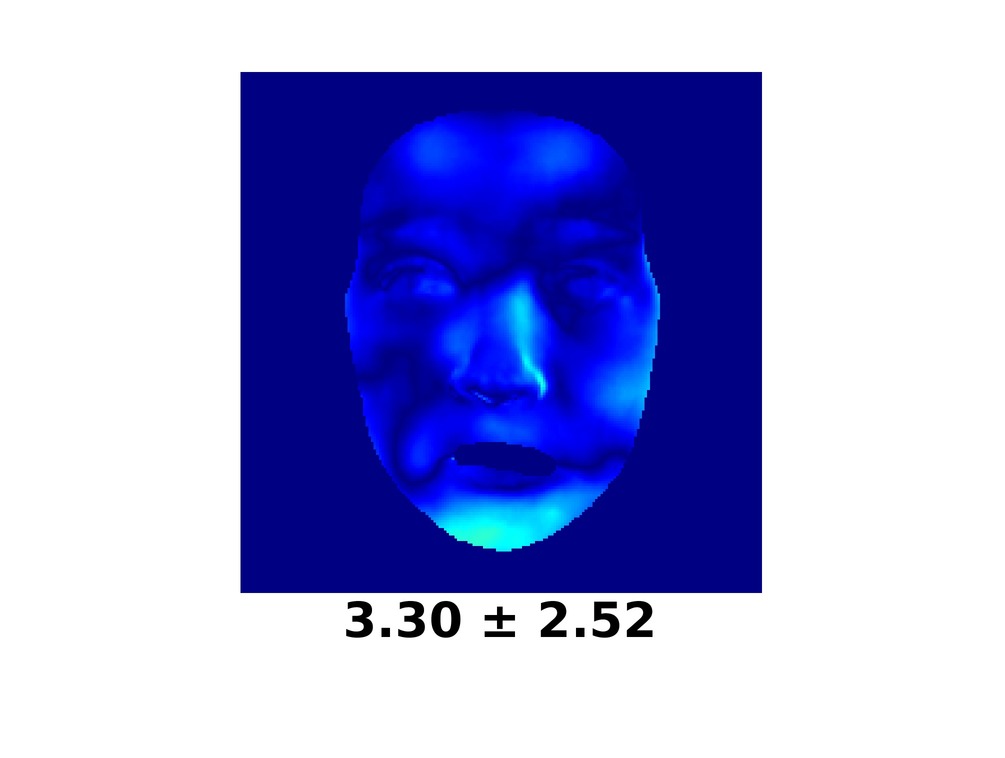}&
      \includegraphics[height=0.28\textwidth,trim={47.5 15 47.5 15},clip]{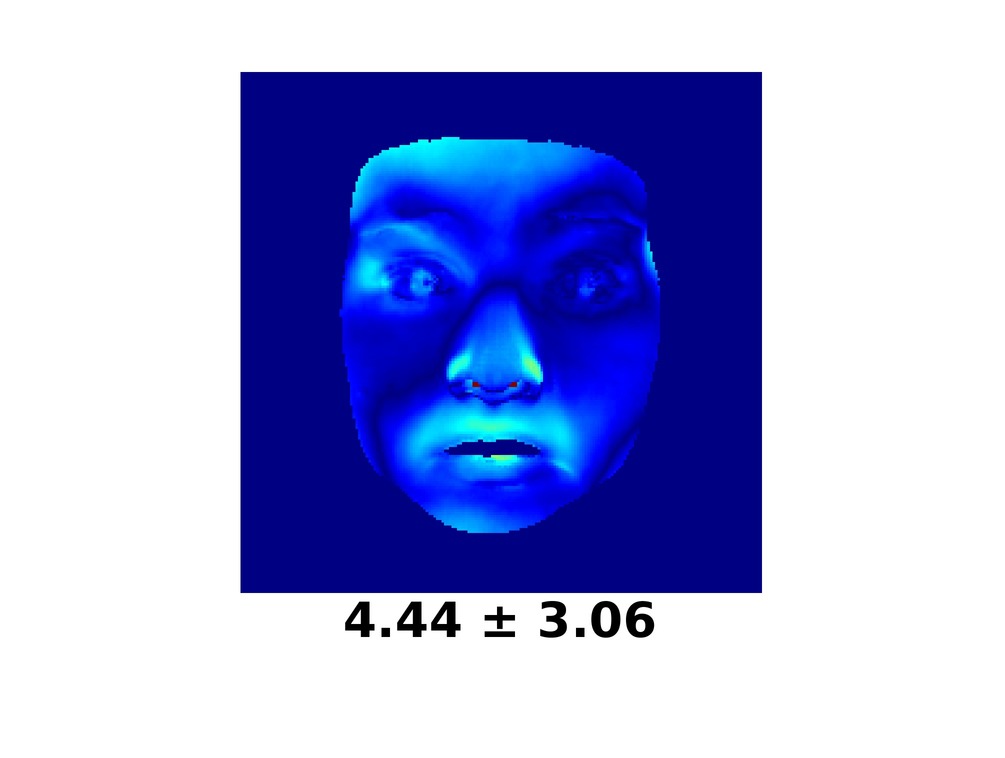}&
      \includegraphics[height=0.28\textwidth,trim={47.5 15 47.5 15},clip]{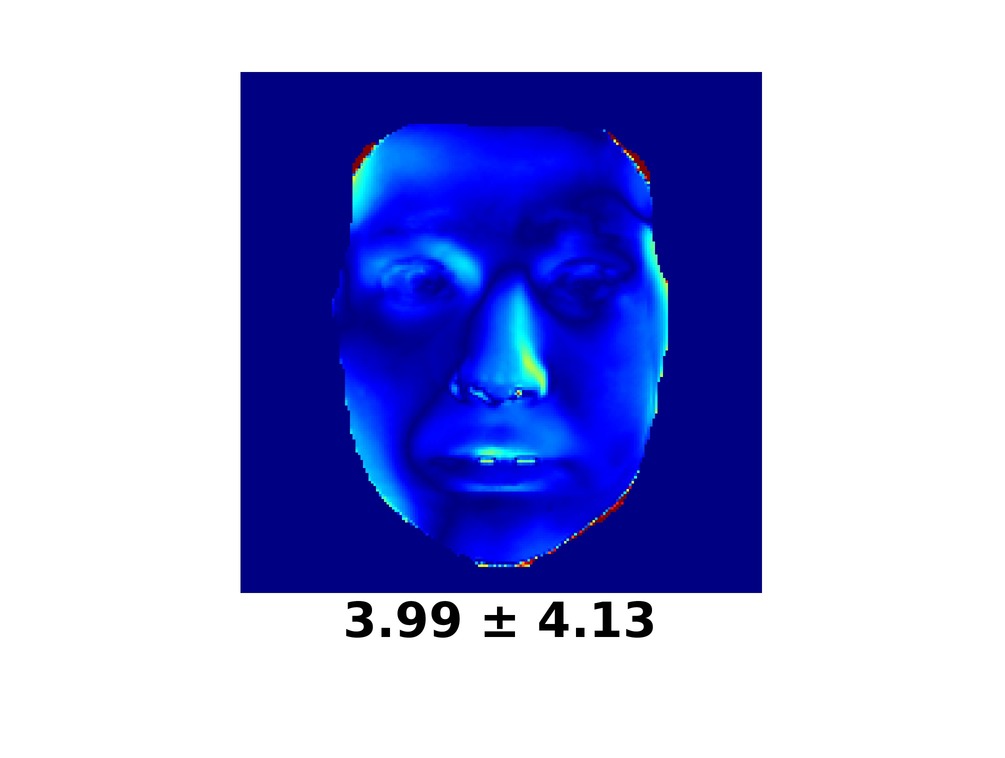} &
      \includegraphics[height=0.28\textwidth,trim={47.5 15 47.5 15},clip]{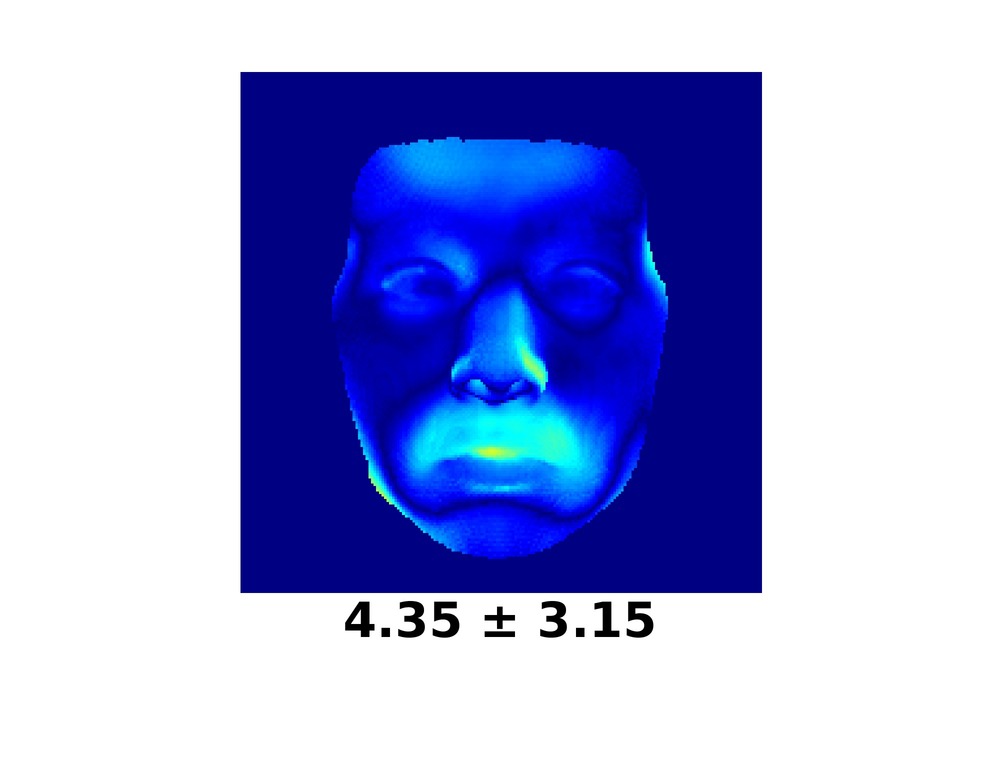} &
      \includegraphics[height=0.28\textwidth]{images/heat_maps/colorbar}
      \tabularnewline
      Input & Proposed & \cite{richardson2016learning} & \cite{kemelmacher20113d} & \cite{zhu2015high} & Err. \% Scale
      \end{tabular}
    \caption{Error heat maps in percentile of ground truth depth.}
    \label{fig:heat_maps}
\end{figure}

\ifshowmain
\else
\newpage
\clearpage

{\small
\bibliographystyle{ieee}
\bibliography{egbib}
}
\fi

\fi

\end{document}